\documentclass[10pt,twocolumn,letterpaper]{article}

\usepackage[pagenumbers]{wacv}              %
\usepackage[accsupp]{axessibility} 
\usepackage{multirow}
\usepackage{diagbox}

\newcommand{\borderimg}[2][]{%
  \fbox{\includegraphics[#1]{#2}}%
}

\newcommand{\rowlabel}[1]{%
  \makebox[0pt][r]{%
    \hspace{-8mm}%
    \raisebox{0.06\textwidth}[0pt][0pt]{%
      \rotatebox[origin=c]{90}{\strut #1}%
    }%
  }%
}

\newcommand{\pare}[1]{\left(#1\right)}

\definecolor{wacvblue}{rgb}{0.21,0.49,0.74}
\usepackage[pagebackref,breaklinks,colorlinks,allcolors=wacvblue]{hyperref}

\def\wacvPaperID{****} %
\def\confName{WACV}
\def\confYear{2027}

\title{UnSPAD: Uncertainty-aware Single Photon Burst Photography}

\author{Sachin Shah\footnotemark[1] \quad
Kalyan Bhetwal \quad
Levi Burner \quad
Christopher A. Metzler\\
University of Maryland, College Park\\
{\tt\small https://inventshah.github.io/UnSPAD}
}

\begin{document}

\twocolumn[{%
\renewcommand\twocolumn[1][]{#1}%
\maketitle
\begin{center}
    \centering
    \captionsetup{type=figure}
    \includegraphics[width=\textwidth]{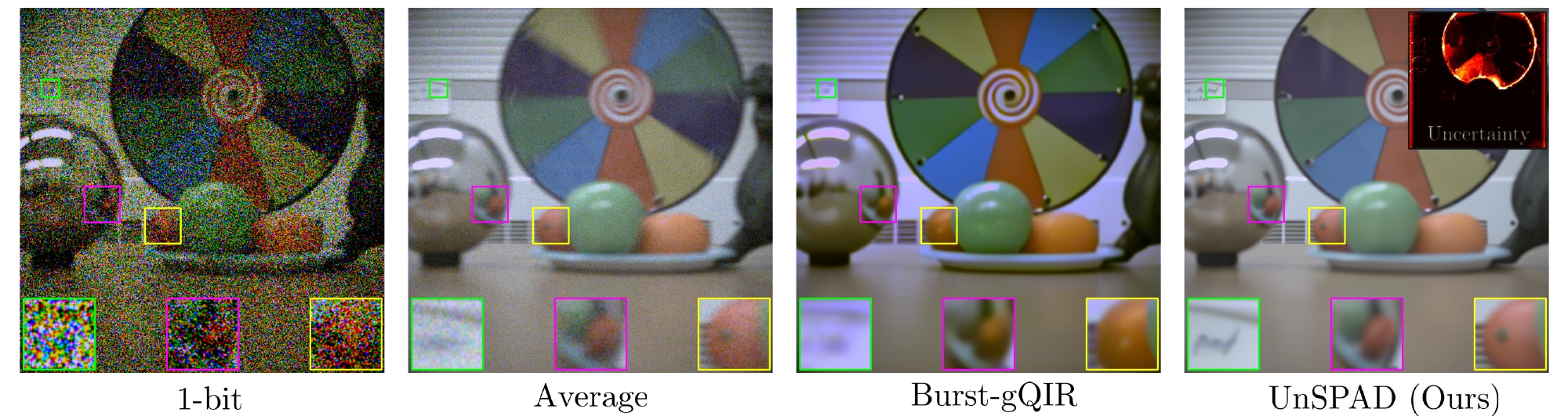}
    \captionof{figure}{ %
    \textbf{UnSPAD applied to experimental SPAD data.}
    By combining robust optical flow algorithms, maximum likelihood estimation, and uncertainty-guided image denoising, UnSPAD is able to reconstruct a high-quality hallucination-free image from a noisy sequence of binary SPAD measurements of a moving scene. Dataset and Burst-gQIR baseline from~\cite{garg_2026_gqir}.
    }\label{fig:teaser}
\end{center}%
 }]

\footnotetext[1]{Corresponding author: shah2022@umd.edu}

\begin{abstract}

\vspace{-5px}
We introduce UnSPAD, an uncertainty-aware burst photography framework capable of fusing and denoising arbitrarily long sequences of binary single photon avalanche diode (SPAD) measurements. 
Unlike prior methods that require fixed burst lengths or rely on powerful generative priors, UnSPAD combines learned motion estimation with uncertainty-informed measurement fusion and denoising to produce temporally consistent, hallucination-free reconstructions. 
UnSPAD first forms an initial scene estimate by averaging flow-aligned binary frames. 
UnSPAD then uses per-pixel uncertainty maps to perform uncertainty-aware denoising; the denoiser is informed about which regions accumulated the fewest observations (and thus contain the most relative shot noise) due to motion, occlusion, or alignment errors. 
This two-step reconstruction pipeline helps mitigate hallucinations---outside of the structure implicitly introduced by the flow estimates, the initial scenes estimates are entirely prior-free. 
Experiments on real and simulated SPAD and quantum image sensor (QIS) benchmarks demonstrate state-of-the-art reconstruction quality, improved temporal consistency, and superior color fidelity relative to the state of the art. %

\end{abstract}

\section{Introduction}
\label{sec:intro}

Single-photon imaging enables sensing under extreme low-light conditions by recording binary photon detection events at micro to nano-second timescales. Such sensors can operate in regimes where conventional cameras fail, making them attractive for applications including scientific imaging~\cite{Bruschini2019bio}, night vision~\cite{Goyal_2021_ICCV}, and high-speed sensing~\cite{Fabrizio2010high}. However, the resulting measurements are corrupted by severe shot noise, and the binary nature of detection during self or independent motion requires fusing multiple binary frames to estimate a high-quality image~\cite{Yang2012bits}. Further, time-varying occlusion patterns determine which and how many binary frames contain overlapping measurements. Our work addresses this challenge by combining learned denoising and flow models with classical geometric reasoning about occlusions and image formation. The result is high-quality frames and videos that are consistent across time.

Fundamentally, images encode the intensity of light or the expected photon arrival rate, where the precise arrival times follow a Poisson distribution. If the scene is stationary, the number of photons arriving at a single-pixel in an interval is a sufficient statistic for the maximum-likelihood estimator of that pixel's brightness, and its error decreases slowly, proportional to $\frac{1}{\sqrt{N}}$. By combining numerous binary frames, one can reconstruct high-quality imagery.

However, in the presence of motion, an individual point in the world projects to multiple pixels, and its projected image trajectory can intersect those of its neighbors. Further, occlusion changes frequently, causing new scene points to become visible and others invisible. In traditional image sensors, these effects present as irreversible motion blur or ghosting in the final images. SPAD sensors offer a path forward by recording individual photon arrival times, so that integration along a scene point's path in image space-time can result in a true, uncorrupted intensity value.
However, untangling these trajectories requires estimating the image motion, or optical flow, from noisy, binary frames, a problem unanticipated by existing optical flow estimation methods, which still rely on the idea of brightness constancy (albeit often abstracted as feature consistency). %

Early approaches attempted to bridge this gap through heavy per-temporal accumulation so that classical vision techniques could estimate pixel motion~\cite{Ma_SIGGRAPH20}.
These methods quickly fail under extreme or complex motion profiles. 
Recent works replace these components with neural networks~\cite{chennuri2024quanta, Chennuri2025QuDI, garg_2026_gqir}, with QuDi~\cite{Chennuri2025QuDI} and gQIR~\cite{garg_2026_gqir} achieving state-of-the-art (SOTA) by leveraging diffusion models which can apply strong priors.
These methods preclude infinite burst photography, since the network needs to be retrained depending on burst size, limiting flexibility.

Our method uses independent networks applied to single or pairs of frames to aggregate bursts of arbitrary length. In practice, this results in high-quality images with SOTA accuracy, even when restricted to the fixed burst lengths of prior works~\cite{chennuri2024quanta, Chennuri2025QuDI, garg_2026_gqir}. Further, by employing maximum likelihood estimators, we avoid incorporating strong priors that result in hallucinations and temporal inconsistency. Finally, our architecture theoretically allows for arbitrary accuracy as a moving burst duration approaches infinity.

Our contributions are summarized as follows:

\begin{itemize}
    \item A learned method for arbitrary length 1-bit burst fusion
    \item A hybrid architecture with geometric and learned components that is hallucination free and temporally consistent
    \item Denoising of aligned bursts via uncertainty quantification
    \item SOTA performance on real and synthetic benchmarks
\end{itemize}

\begin{figure}
\centering
\includegraphics[width=0.9\linewidth]{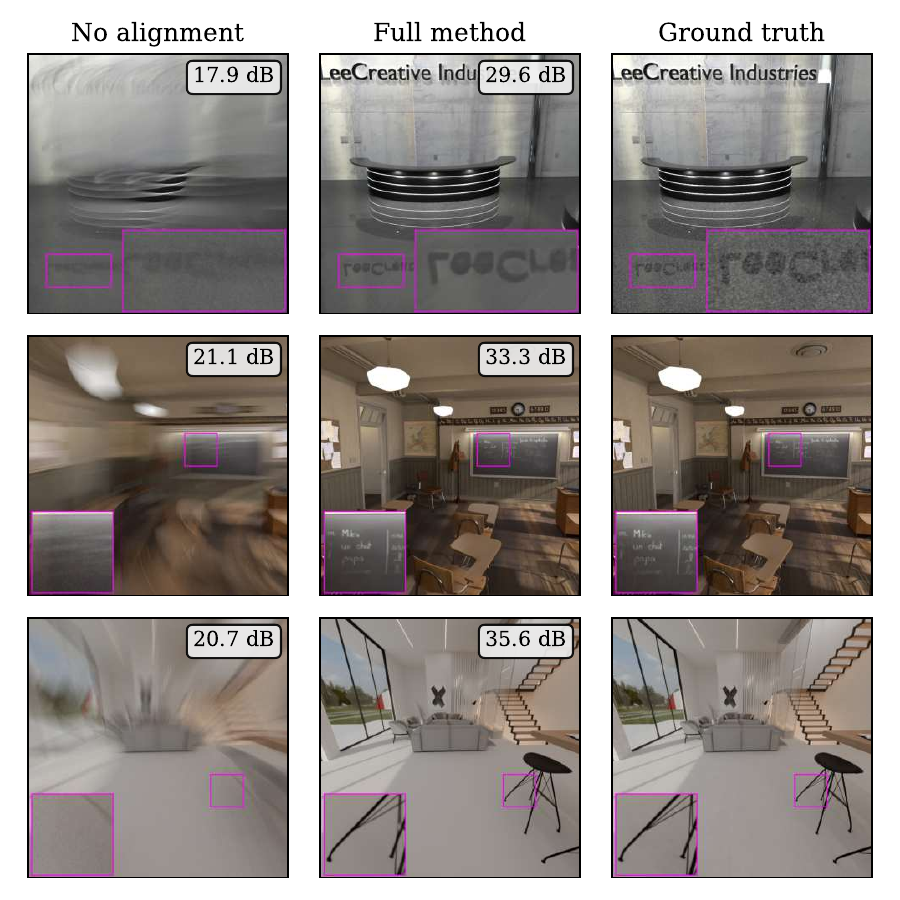}
\caption{\textbf{UnSPAD qualitative results} on simulated SPAD data.
UnSPAD's robust temporal aggregation with spatial denoising recovers low-blur and low-noise images.
}
\label{fig:grid-gt}
\end{figure}

\section{Related Work}
\label{sec:related}

\subsection{Spatial Denoising}

Spatial image denoising has evolved from classical hand-crafted algorithms to advanced deep-learning-based models. Early classical approaches relied on internal image statistics and structural self-similarity~\cite{Tomasi1998BilateralFF, bm3d, Lebrun2013, Condat2013tv}. One example is the Block-Matching and 3D Filtering (BM3D) algorithm~\cite{bm3d}, which groups similar 2D image patches into 3D arrays to perform denoising in the transform domain. 

With the advent of deep learning, state of the art methods moved to learning complex noise priors directly from data.
Convolutional Neural Networks (CNNs) have been a popular architecture for this task~\cite{zhang2017beyond,zuo2018convolutional,Mohan2020Robust}. Performance improved through architecture changes such as U-Nets~\cite{zhang2021plug,hurault2022gradient} or Transformers~\cite{zamir2022restormer,zhang2023practical}.
One important innovation came from FDD-Net~\cite{zhang2018ffdnet}, where the network received an extra weighting mask to assist with spatially variable denoising needs.
DRUNet~\cite{zhang2021plug} integrated this idea into a plug-and-play image restoration framework.
Restormer~\cite{zamir2022restormer} introduced an efficient Transformer architecture tailored for high-resolution images, capturing long-range pixel dependencies across channels to achieve state-of-the-art performance in spatial denoising.

\subsection{Burst Photography}

Burst photography is when multiple frames are captured in quick succession. Even small burst sizes have been shown to be useful in a variety of applications, particularly on mobile platforms where the sensor size is limited. Processing bursts can enable high dynamic range imaging~\cite{hdrplus}, super resolution~\cite{lafenetre23handheld}, low-light imaging~\cite{Libalowlight}, depth estimation~\cite{chugunov2023shakes}, and occlusion removal~\cite{chugunov2024neural}.

\paragraph{Bursts for Denoising.} Cameras have many sources of noise, such as random photon arrival. A simple approach to reduce noise is to average multiple frames of the same scene. Since noise is uncorrelated across frames, the signal-to-noise ratio improves. However, in practice, capturing the exact same scene is difficult due to camera shake and scene motion. Classical approaches bridge this gap by first aligning frames based on a flow model. Google HDR+~\cite{hdrplus} introduced a robust architecture for aligning and merging a burst of underexposed frames using block matching to align alternate frames to the selected reference.

\paragraph{Bursts for Quanta Sensors.} Burst processing has shown general purpose use for SPAD data~\cite{Ma2023Burst}.  
Quanta Burst Photography (QBP)~\cite{Ma_SIGGRAPH20} introduced a strategy for fusing a burst of binary frames, with \cite{Ma_SIGGRAPH23} extending to mosaicked color data. It follows the classical align-and-merge concept from traditional burst photography, using block-matching temporal alignment with robust Wiener filtering. However, block-matching breaks down in scenes captured by cameras undergoing long rich 6-DOF motion, particularly rotation.

Recent advances have replaced these steps with trainable neural networks.
QUIVER~\cite{chennuri2024quanta} estimates pixel flow with SpyNet~\cite{spynet2017}, and uses recurrent fusion for merging.
QuDI~\cite{Chennuri2025QuDI} builds on this idea and leverages Diffusion U-Nets for merging features. 
These works require a Quanta Image Sensor (QIS) with 3-bit measurements. Adapting for a SPAD sensor requires averaging $7$-frames, which introduces motion blur at the earliest stages of the pipeline. In addition, these works are trained over $30$ to $500$ GPU hours.

Recently, gQIR~\cite{garg_2026_gqir} demonstrated reconstruction from only $7$ $1$-bit frames by leveraging large generative priors. They extended this powerful idea to the burst domain using RAFT optical flow~\cite{teed2020raft} to align latents and using a vision transformer for fusion. Burst-gQIR is trained for $77$ binary frames, naturally limiting burst size.
Moreover, generative models are susceptible to hallucinations.

Bit2bit~\cite{bit2bit} proposes an interesting alternative based on self-supervision. Denoising becomes training a sequence-specific U-Net to predict per-pixel photon-arrival probabilities that satisfy a masked loss. While bit2bit supports arbitrary window sizes, it is a classical optimization, requires long sequences, and is computationally expensive.

\begin{figure*}
    \centering
    \includegraphics[width=\linewidth]{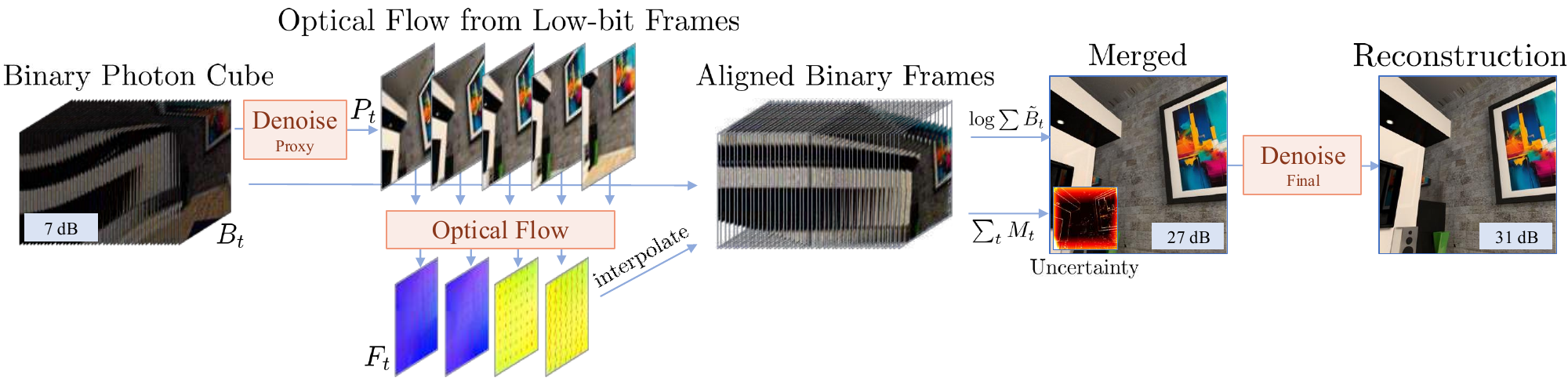}
    \caption{\textbf{Method overview.} UnSPAD aligns $1$-bit SPAD frames guided by optical flow computed on low-bit proxy frames generated by a light-weight denoiser. The aligned binary burst can be merged to produce a high-quality, temporally denoised result. However, due to scene/camera motion or optical flow errors, some regions may be under-observed. In the example, the camera is pan-tilting, causing the outer image edges to have higher uncertainty than the center. A final denoiser can leverage the uncertainty map to guide its strength.
    }\label{fig:method-overview}
\end{figure*}

\begin{figure}
    \centering
    \makebox[0.06\linewidth][c]{\raisebox{0.23\linewidth}{\rotatebox[origin=c]{90}{$1$-bit}}}%
    \borderimg[width=0.42\linewidth]{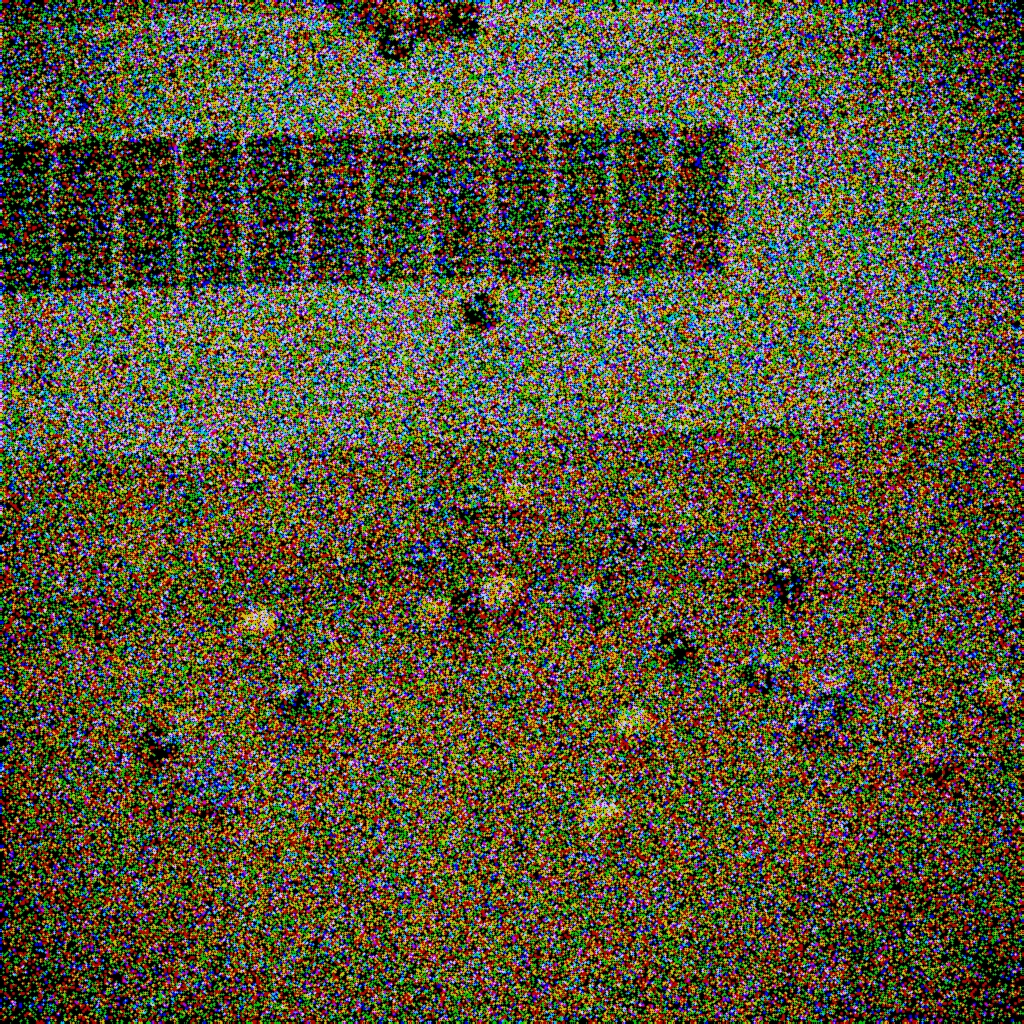}
    \borderimg[width=0.42\linewidth]{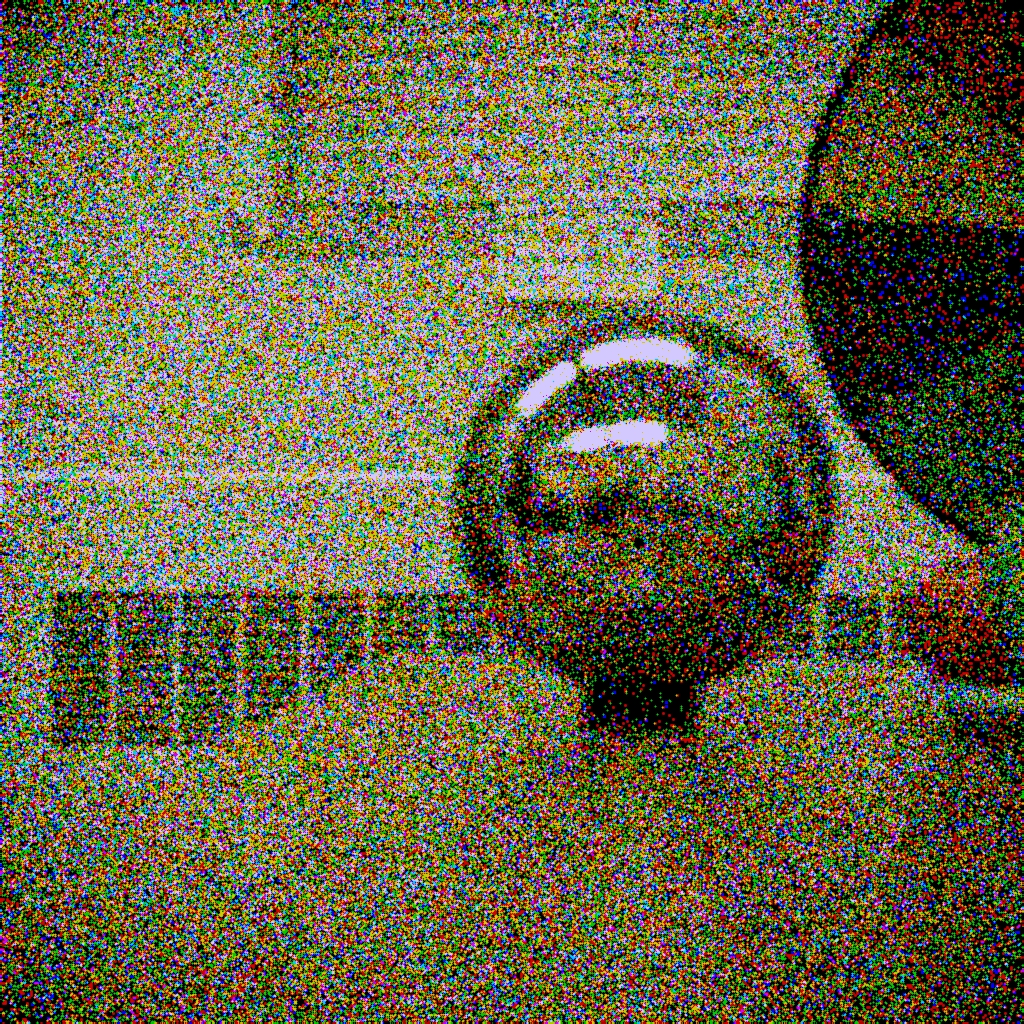}
    \makebox[0.06\linewidth][c]{\raisebox{0.23\linewidth}{\rotatebox[origin=c]{90}{Flow Field}}}%
    \borderimg[width=0.42\linewidth]{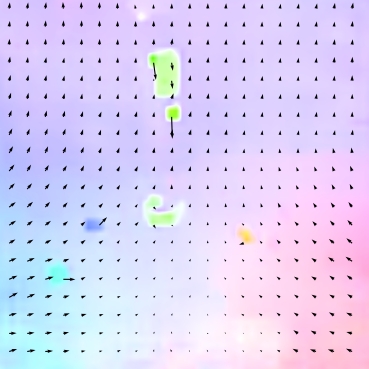}
    \borderimg[width=0.42\linewidth]{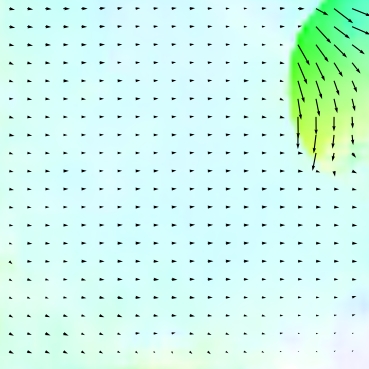}
    \makebox[0.06\linewidth][c]{\raisebox{0.23\linewidth}{\rotatebox[origin=c]{90}{Aligned}}}%
    \borderimg[width=0.42\linewidth]{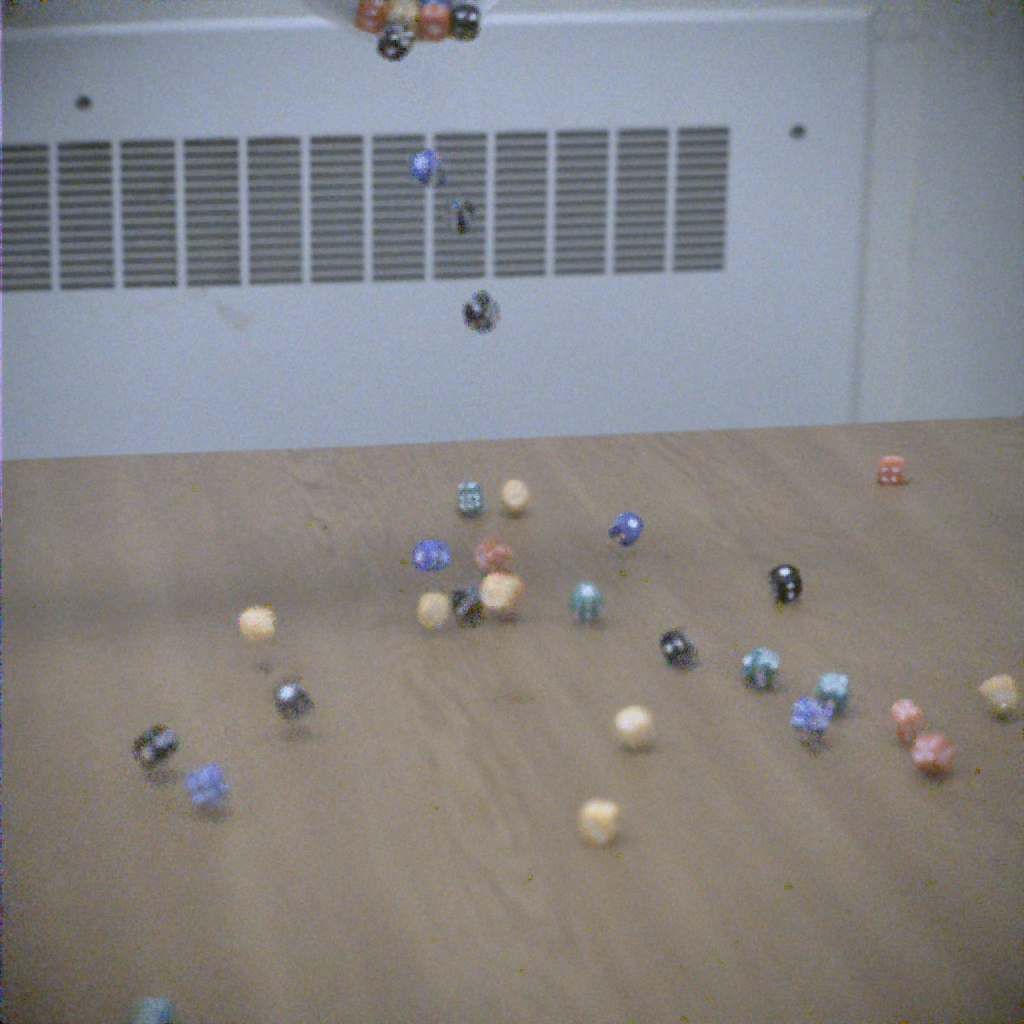}
    \borderimg[width=0.42\linewidth]{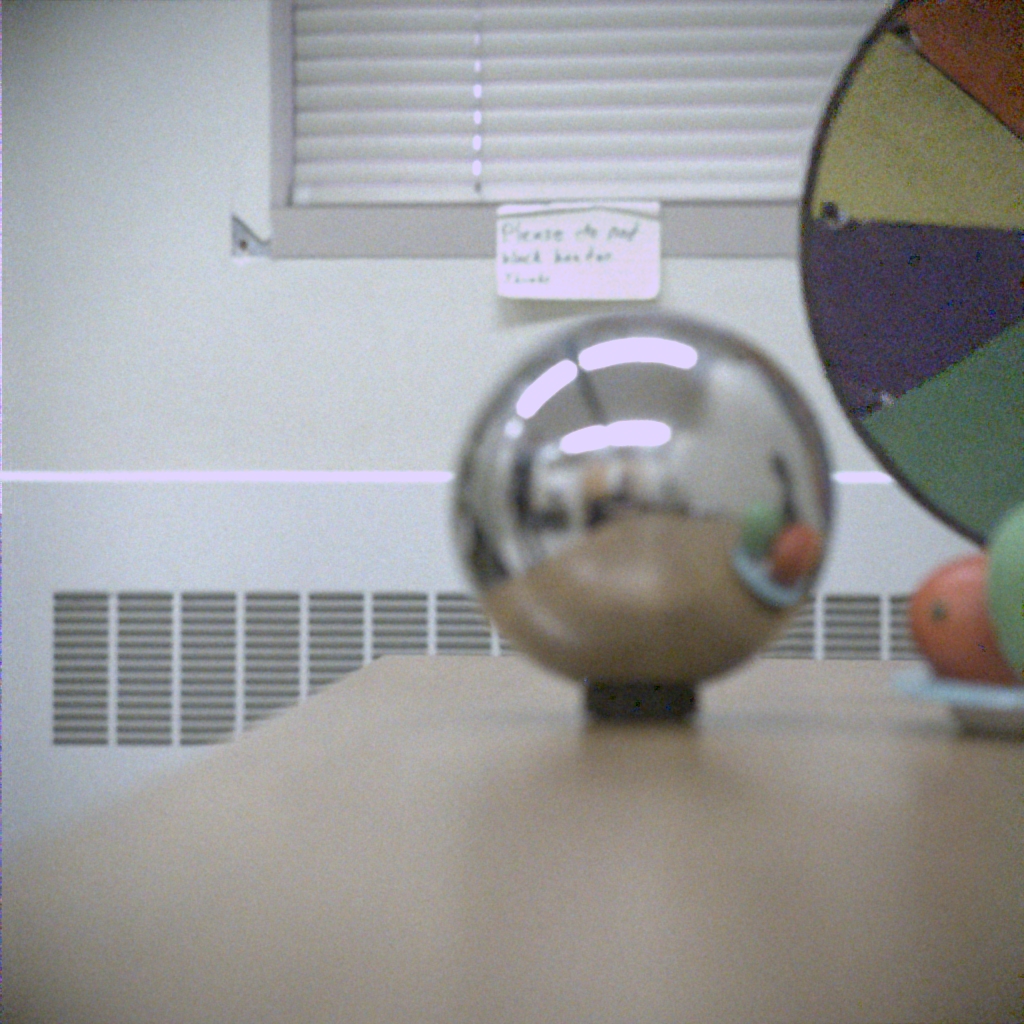}
    \makebox[0.06\linewidth][c]{\raisebox{0.23\linewidth}{\rotatebox[origin=c]{90}{Uncertainty}}}%
    \borderimg[width=0.42\linewidth]{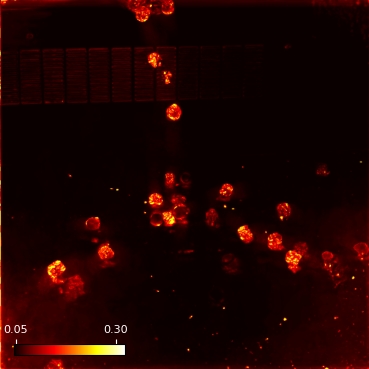}
    \borderimg[width=0.42\linewidth]{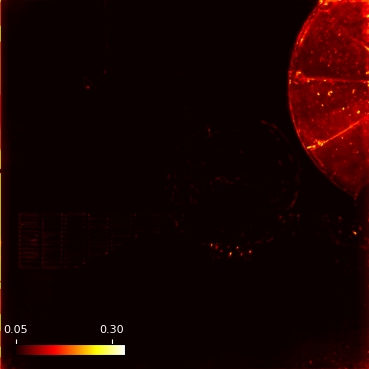}
    \makebox[0.06\linewidth][c]{\raisebox{0.23\linewidth}{\rotatebox[origin=c]{90}{Final}}}%
    \borderimg[width=0.42\linewidth]{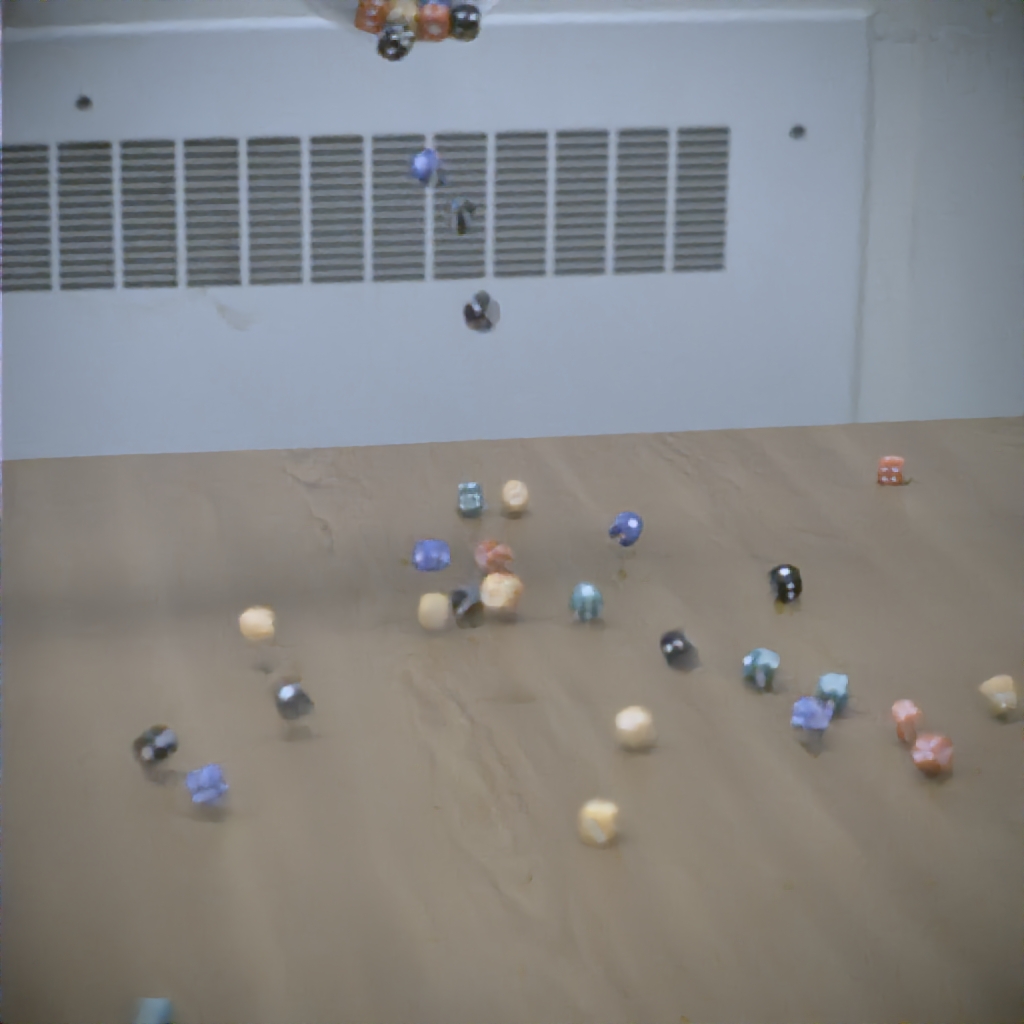}
    \borderimg[width=0.42\linewidth]{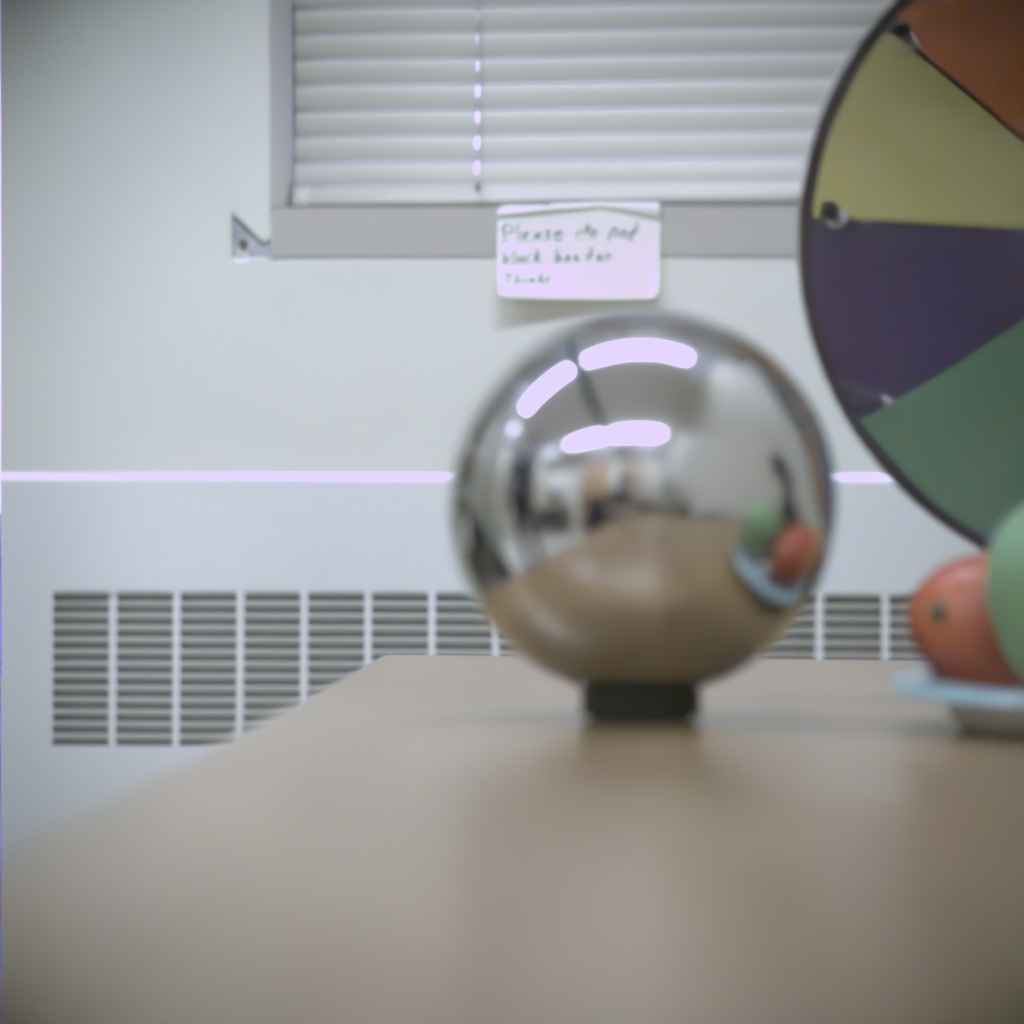}
    \caption{\textbf{Intermediate reconstructions.} Binary frames are warped with flow fields to produce temporally aligned bursts and uncertainty maps. Uncertainty is concentrated near moving objects since those pixels are sampled less often and have a higher likelihood of flow error, which guides the final denoiser. 
    }
    \label{fig:method-intermediates}
\end{figure}

\section{Method}

Our method consists of two main phases: (1) \textbf{Temporal Denoising} and (2) \textbf{Spatial Denoising}.
In the first phase, we attempt to maximize the information extracted from the photon burst by aligning measurements across time without extra hallucinations.
This already produces a sound result, but due to motion, may leave parts of the frame under observed and still noisy.
The second phase cleans up these parts to produce a more visually appealing image.

\subsection{Image Formation Model}

Passive single-photon sensors indicate if a photon arrived over a fixed exposure time interval. In an ideal passive imaging scenario, the sensor is exposed to a constant ambient light source with an optical flux of $\Phi$ (photons per second). The physical process of photon arrivals at the sensor surface is fundamentally stochastic. Assuming a stable light source, this arrival process is modeled as a Poisson process. The probability of $k$ photons arriving at a pixel during an exposure time $\tau$ depends on the expected number of incident photons, denoted by the rate parameter $\lambda$. This rate is proportional to the optical flux and the exposure duration $\lambda = \Phi\tau$.
The probability mass function dictating that exactly $k$ photons hit the pixel area within the window $\tau$ is:
\begin{equation}
P(K = k) = \frac{\lambda^k e^{-\lambda}}{k!}.
\end{equation}

In practice, a single-photon detector cannot register an infinite number of photons, instead the count is quantized to a single-bit where the pixel simply reports whether \textit{at least one} photon arrived during the exposure window $\tau$~\cite{Yang2012bits}. Using the Poisson distribution for $k\ge1$ with the effective detection rate, this probability is:
\begin{equation}
P(k \ge 1) = 1 - e^{-\lambda} = 1 - e^{-\Phi \tau}
\end{equation}
This binarization reduces the measurement down to a Bernoulli random variable, $B_t\sim \text{Bernoulli}\left(1 - e^{-\Phi \tau}\right)$.
The unbiased maximum likelihood estimator of $p$ is the sample mean, so averaging multiple $1$-bit frames forms the estimate of the photon detection probability. Using the invariance property of MLEs, the scene radiance $\Phi$ can be estimated with $-\frac{1}{\tau}\ln\left(1 - \hat{p}\right)$.

\subsection{Temporal Denoising}

We enhance the align-and-merge method of QBP to re-align pixels across time using optical flow such that each pixel stack corresponds to an independent Bernoulli distribution.

\subsubsection{Proxy Frames}

Computing optical flow directly on binary frames is highly sensitive to noise. To overcome this, prior work relies on proxy images that are more amenable to flow estimation. QBP uses temporal aggregation~\cite{Ma_SIGGRAPH20}, QUIVER and QuDi apply pre-denoising~\cite{chennuri2024quanta, Chennuri2025QuDI}, and gQIR uses generative reconstruction~\cite{garg_2026_gqir}. 

We take a similar approach, differing mainly in aggregation window size and denoising method. Specifically, we perform block-wise temporal averaging over non-overlapping windows of size $k=15$, similar to QBP but with shorter aggregation periods. We then apply average pooling downsampling, which effectively increases the pixel detection area by $4$x.
Next, we map photon detection probabilities to linear RGB and convert to sRGB values using VisionSIM~\cite{visionsim}.
Then, we apply a pretrained convolutional denoiser, DRUNet~\cite{zhang2021plug}, with a noise level of $0.15$ to remove residual noise and produce the final proxy frames.

Note that because the image motion is smooth, up to changes in occlusion, each low bit-depth proxy frame represents the image at the center of the aggregated time-window. %

\subsubsection{Flow Estimation}

Optical flow $F_{r \to i}$ is estimated from the target view's proxy frame $P_r$ to every other proxy frame $P_i$.
\begin{equation}
F_{r \to i}
= \operatorname{OF}(P_r, P_i;\, W_i),
\end{equation}
where $W_i$ is the warm start initialization used by the optical-flow solver. Naturally, $F_{r\to r} = 0$. 
We choose the small-RAFT optical flow network~\cite{teed2020raft} due to its strong performance on large motion. Since RAFT is recurrent, flow is progressively computed outwards from the reference index 
such that the previous iteration flow is the warm start for the current iteration. That is, $W_i = F_{r\to i-1}$ for $i=r+1,\ldots,N$, and $W_i = F_{r\to i+1}$ for $i=N,\ldots,r-1$.

To improve robustness, we run the above procedure forward and backward in time, allowing estimation of photometric consistency in both directions. For a given frame $i$, the validity of the flow field is estimated with
\begin{align}
    M_i = \lVert F_{r\xrightarrow{}i}\pare{x} + F_{i\xrightarrow{}r}\pare{x + F_{r\xrightarrow{}i}\pare{x}} \rVert_2 < t.
\end{align}
where $t=5$ is the consistency threshold. From this self-supervised forward-backward consistency check, photon detections that are mismatched due to either occlusion or erroneous flow estimates can be detected and discarded.

\subsubsection{Aligned Aggregation}

Once motion has been estimated from low-bit depth frames, we transfer it to the binary frames. In congruence with the earlier observation that each proxy frame corresponds to the center of its aggregation window, each proxy flow is interpreted as the motion from the reference time to the center binary frame of its temporal window. Consequently, we linearly interpolate between consecutive proxy flows to approximate the flow field corresponding at a specific binary frames timestamp and warp it to the reference viewpoint.

After warping, each raw binary frame contributes to the final image only where it is valid. Let $\tilde{B}_t$ denote the warped frame and $M_t$ its validity mask. Our temporal fusion is a weighted average:
\begin{equation}
\hat{I} = -\frac{1}{\tau} \ln\pare{1- \frac{\sum_t M_t \odot \tilde{B}_t}{\sum_t M_t}},
\end{equation}
where $\odot$ denotes elementwise multiplication, and the argument of the logarithm is clipped to be at least $10^{-7}$.

\subsection{Spatial Denoising}

We adopt Restormer~\cite{zamir2022restormer}, an efficient transformer-based restoration network that uses multi-head transposed self-attention operating along the channel dimension rather than the spatial dimension. This design enables a global receptive field without the quadratic spatial complexity of standard self-attention, making it well-suited for high-resolution SPAD images.

Inspired by FFDNet~\cite{zhang2018ffdnet}, we provide the model with additional context about which regions of the image may require more restoration via a per-pixel weight map. The previous stage tracks the number of effective frames contributing to each pixel, denoted by~$\sum_t M_t$. We convert this into an uncertainty channel with the inverse-square-root: $1/{\sqrt{\sum_t M_t}}$.
Values near $1$ indicate unreliable pixels (few contributing frames) and values near $0$ indicate well-sampled regions. The uncertainty map is concatenated with the RGB input $\hat{I}_t$ as the 4\textsuperscript{th} channel, expanding the patch embedding from $\mathbb{R}^{48 \times 3 \times 3 \times 3}$ to $\mathbb{R}^{48 \times 4 \times 3 \times 3}$.
Restormer maps the 4 channel input to a denoised image $\hat{I}_s$.

 \subsubsection{Training Details}

 \paragraph{Initialization.}
Restormer is initialized from weights pretrained on millions of image patches from the DFWB dataset~\cite{singh2019recognizing} for the Gaussian color denoising task. This provides a strong natural image prior for edge preservation, texture recovery, and smooth region denoising.

\paragraph{Datasets.}
We use the 1{,}850 paired training samples across $50$ scenes from the 2026 Single Photon Challenge~\cite{visionsim}. 10\% of the pairs are set aside for validation, with one scene held out entirely. Each burst is aligned and merged ahead of time, recording its uncertainty map simultaneously. These aligned images and uncertainties with corresponding ground truth images form the training dataset. We bias the training data sampler towards challenging regions which were under observed in a burst (either due to large motion or alignment error). 80\% of crops are centered on low-weight pixels selected from a candidate pool of $16$ positions ($8$ random, $4$ corners, and $4$ edge midpoints), choosing the candidate with the lowest mean weight. The remaining 20\% are uniformly random.

 \paragraph{Optimization.}
We finetune for 10{,}000 steps with batch size 4 and $512 \times 512$ random crops using Charbonnier loss $\ell_1\pare{\hat{I}_s, I}$, with $\epsilon = 10^{-6}$.
We use AdamW with initial learning rate $10^{-4}$ and cosine annealing to $10^{-7}$. Mixed-precision training (FP16) is used for efficiency, with gradient clipping at norm $1.0$. All experiments were performed on an NVIDIA RTX A6000 and H200 SXM.

\section{Experiments}

\begin{figure*}[htbp]
    \centering
    \makebox[0.03\linewidth][c]{\raisebox{0.095\textwidth}{\rotatebox[origin=c]{90}{$1$-bit}}}%
    \includegraphics[width=0.19\textwidth,height=0.19\textwidth]{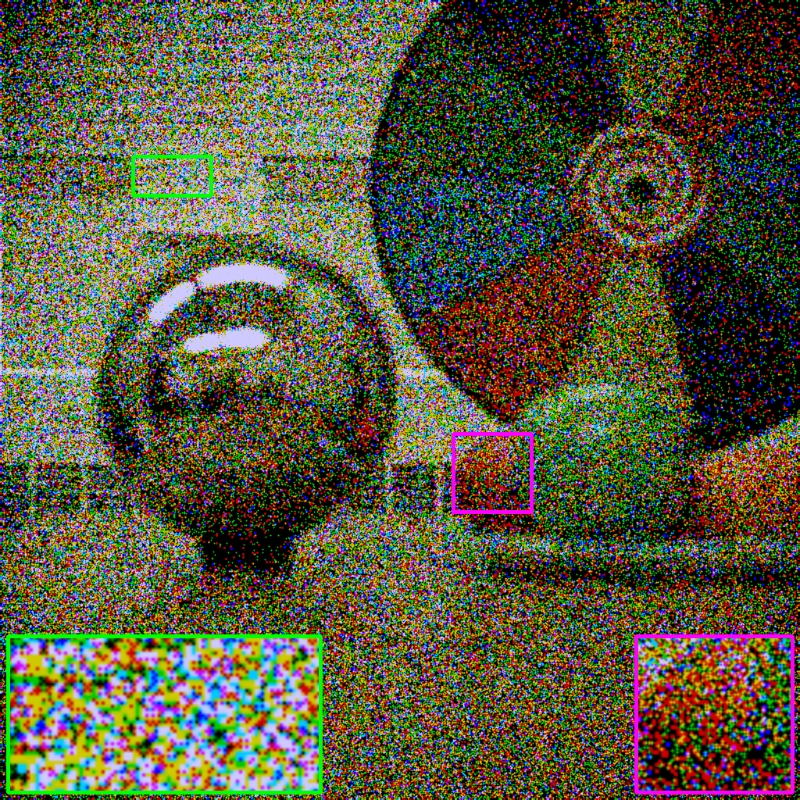}
    \includegraphics[width=0.19\textwidth,height=0.19\textwidth]{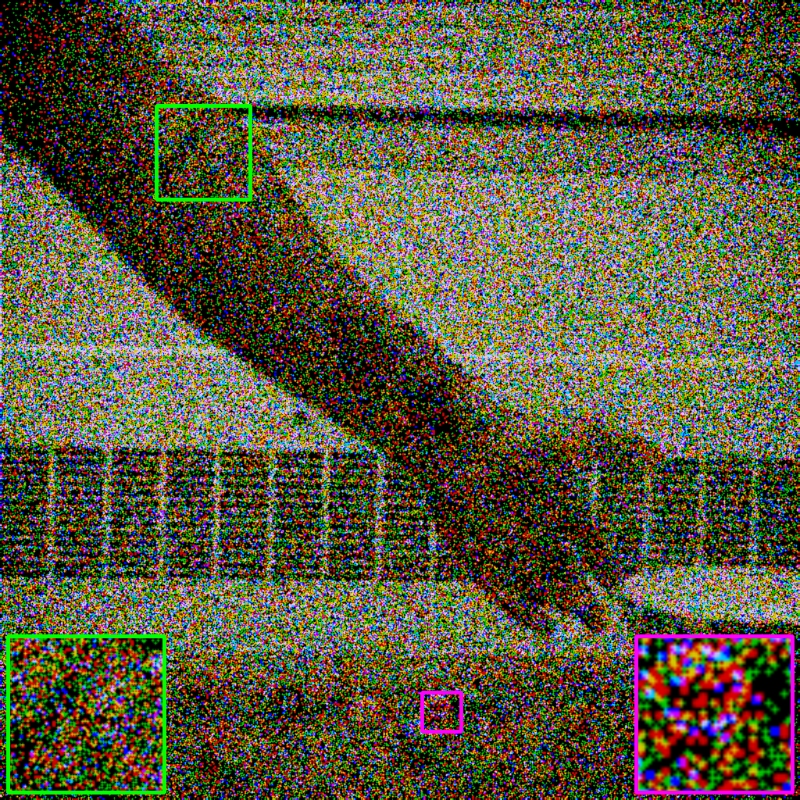}
    \includegraphics[width=0.19\textwidth,height=0.19\textwidth]{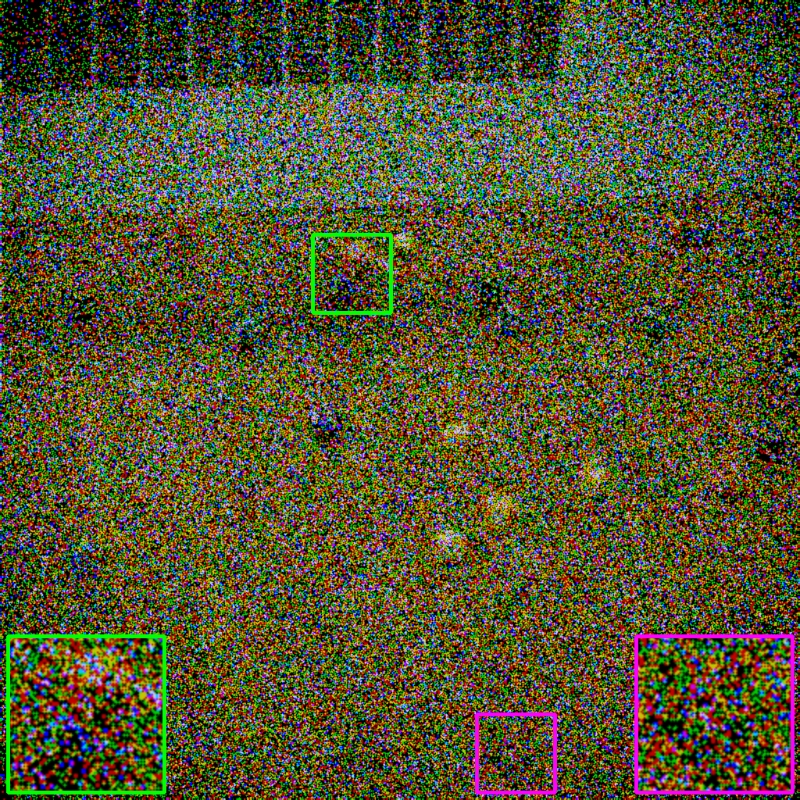}
    \includegraphics[width=0.19\textwidth,height=0.19\textwidth]{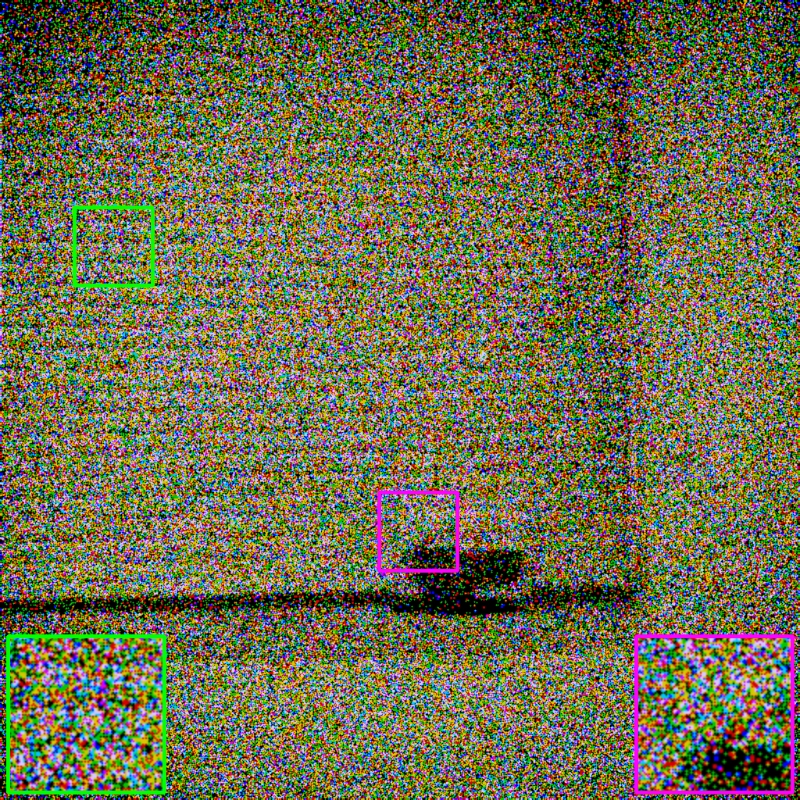}
    \includegraphics[width=0.19\textwidth,height=0.19\textwidth]{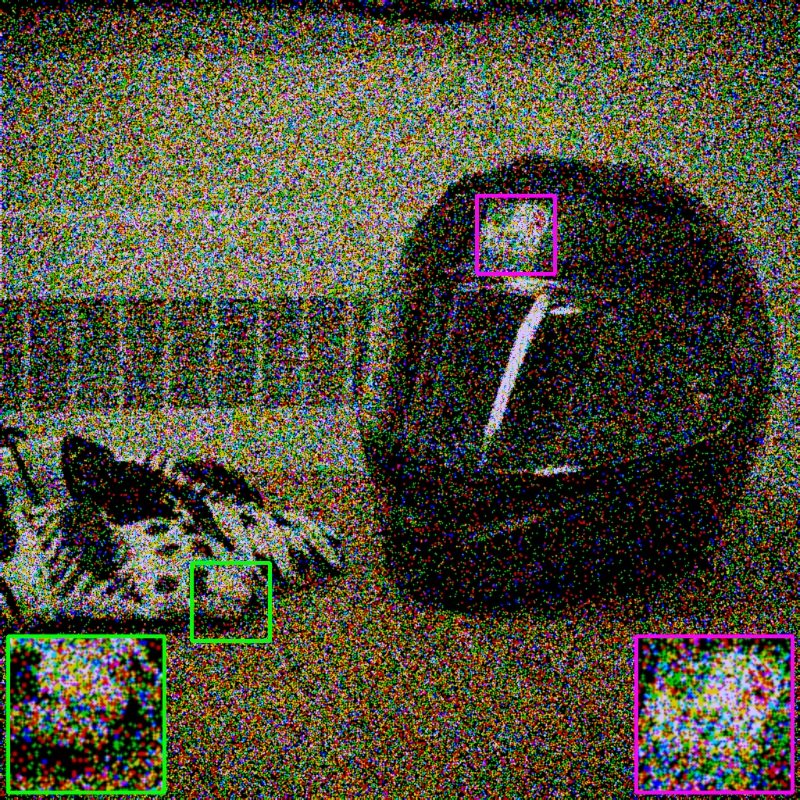}
    \makebox[0.03\linewidth][c]{\raisebox{0.095\textwidth}{\rotatebox[origin=c]{90}{Average of $77$}}}%
    \includegraphics[width=0.19\textwidth,height=0.19\textwidth]{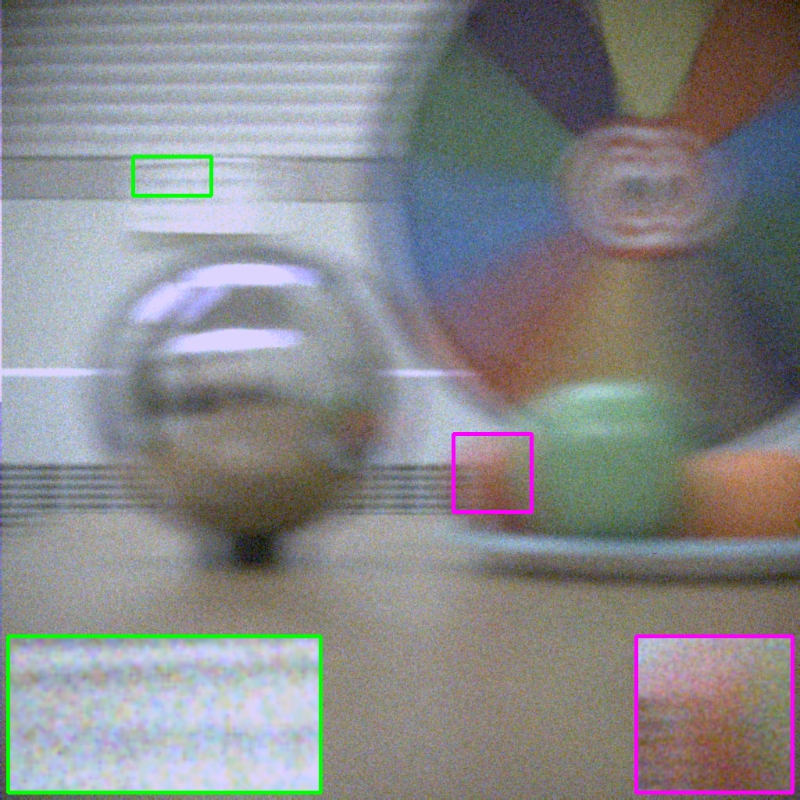}
    \includegraphics[width=0.19\textwidth,height=0.19\textwidth]{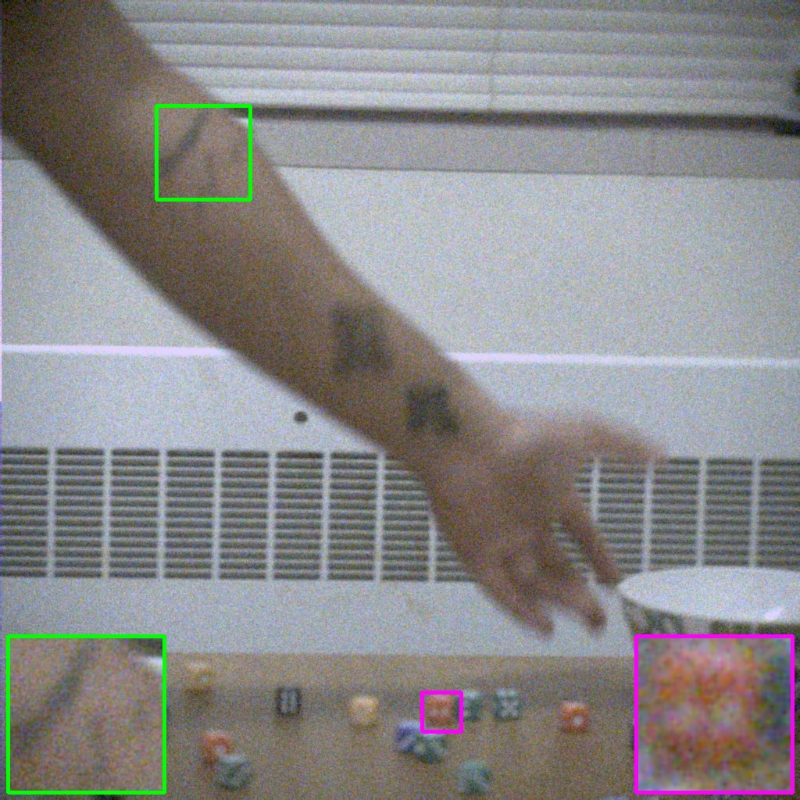}
    \includegraphics[width=0.19\textwidth,height=0.19\textwidth]{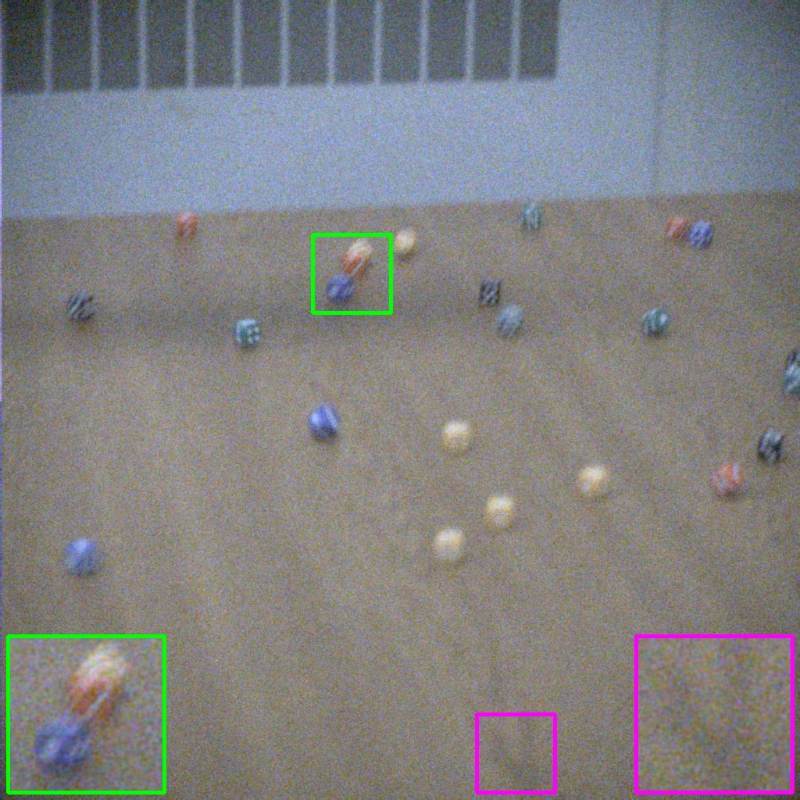}
    \includegraphics[width=0.19\textwidth,height=0.19\textwidth]{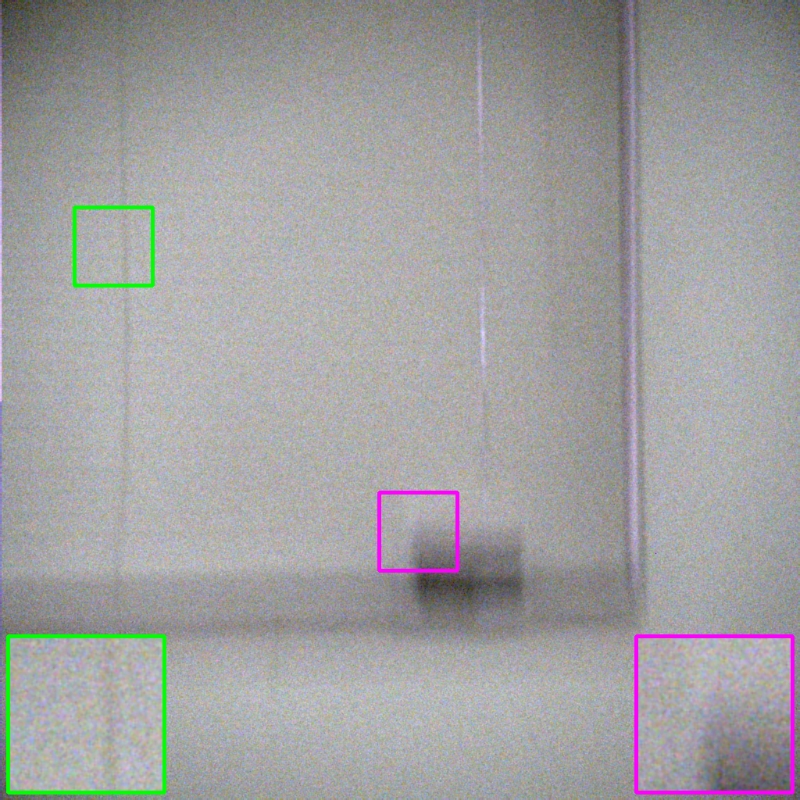}
    \includegraphics[width=0.19\textwidth,height=0.19\textwidth]{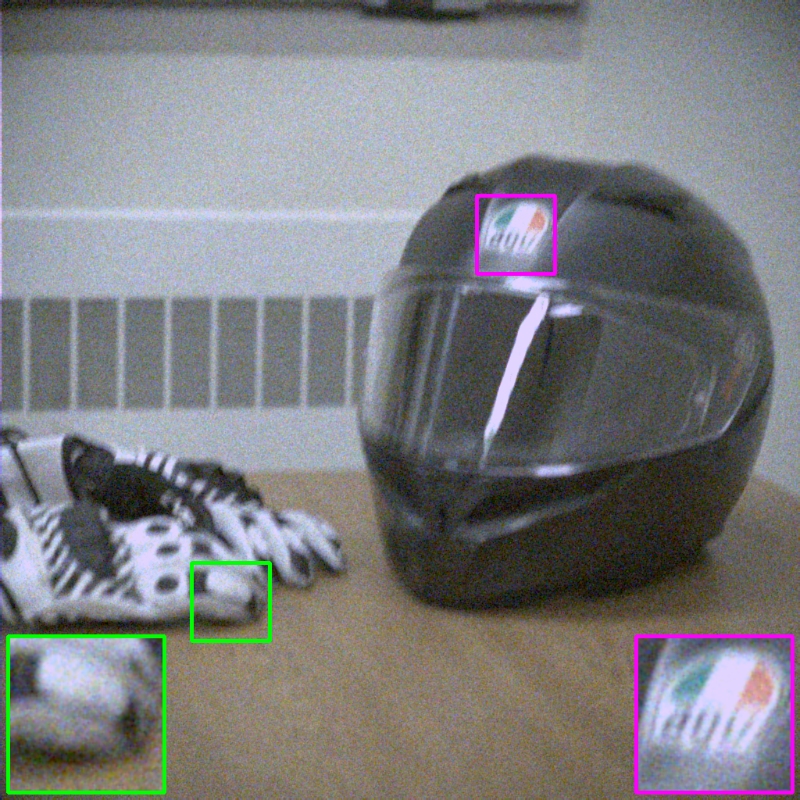}
    \makebox[0.03\linewidth][c]{\raisebox{0.095\textwidth}{\rotatebox[origin=c]{90}{Burst-gQIR~\cite{garg_2026_gqir}}}}%
    \includegraphics[width=0.19\textwidth,height=0.19\textwidth]{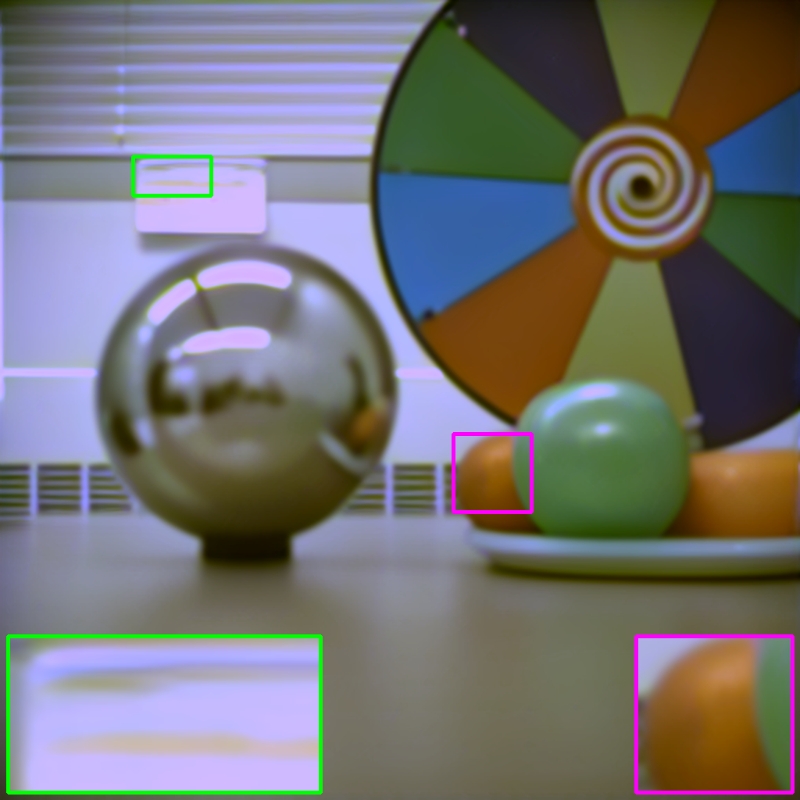}
    \includegraphics[width=0.19\textwidth,height=0.19\textwidth]{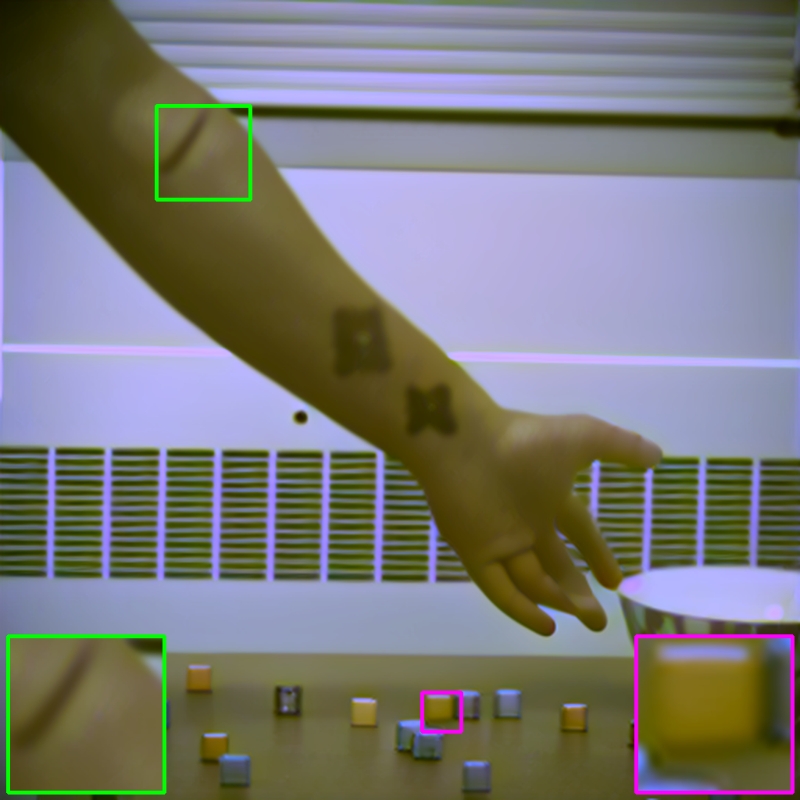}
    \includegraphics[width=0.19\textwidth,height=0.19\textwidth]{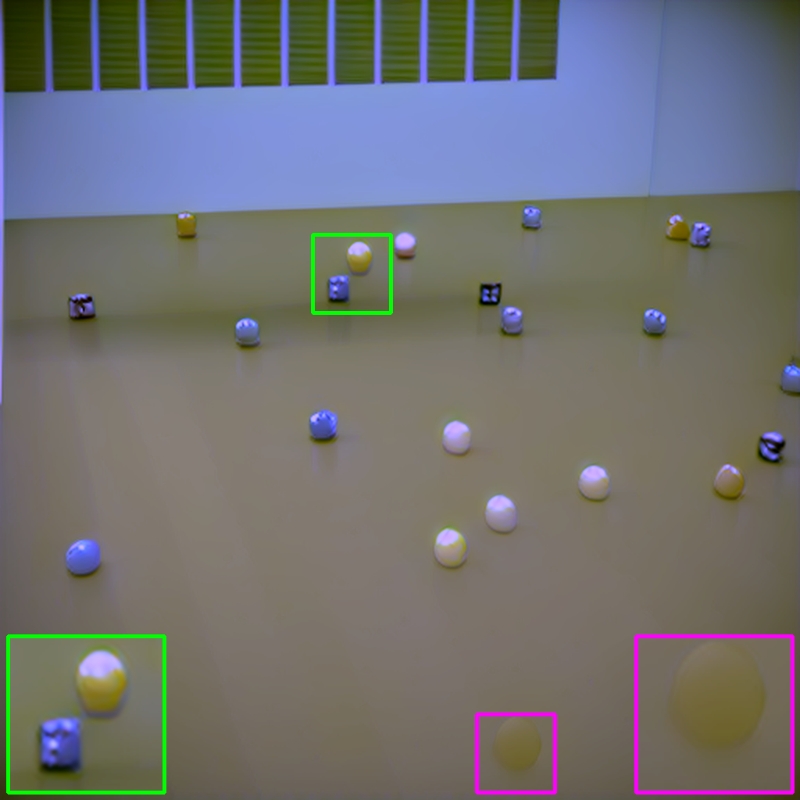}
    \includegraphics[width=0.19\textwidth,height=0.19\textwidth]{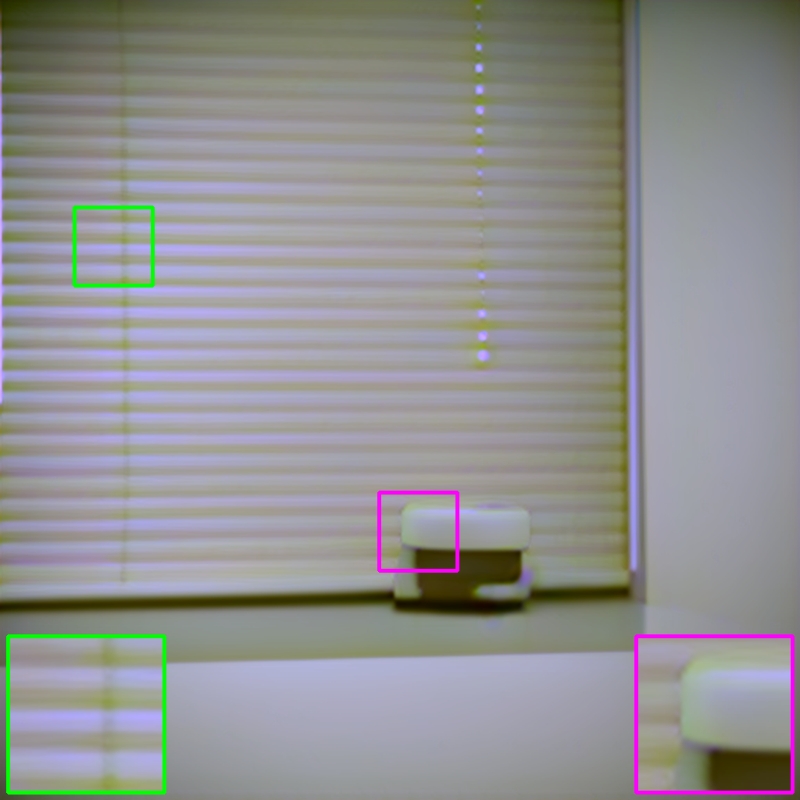}
    \includegraphics[width=0.19\textwidth,height=0.19\textwidth]{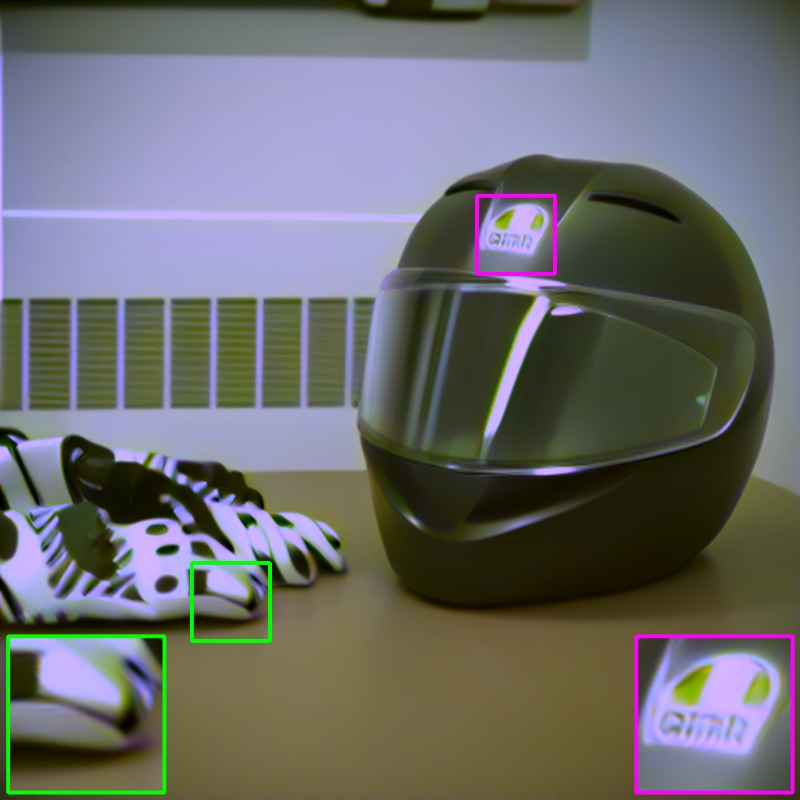}
    \makebox[0.03\linewidth][c]{\raisebox{0.095\textwidth}{\rotatebox[origin=c]{90}{UnSPAD (Ours)}}}%
    \includegraphics[width=0.19\textwidth,height=0.19\textwidth]{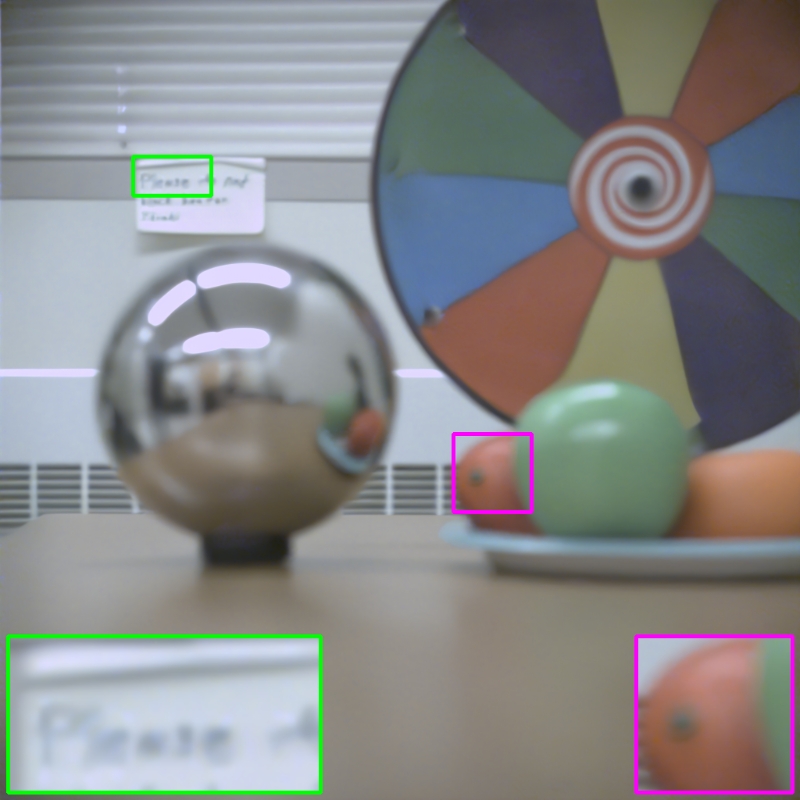}
    \includegraphics[width=0.19\textwidth,height=0.19\textwidth]{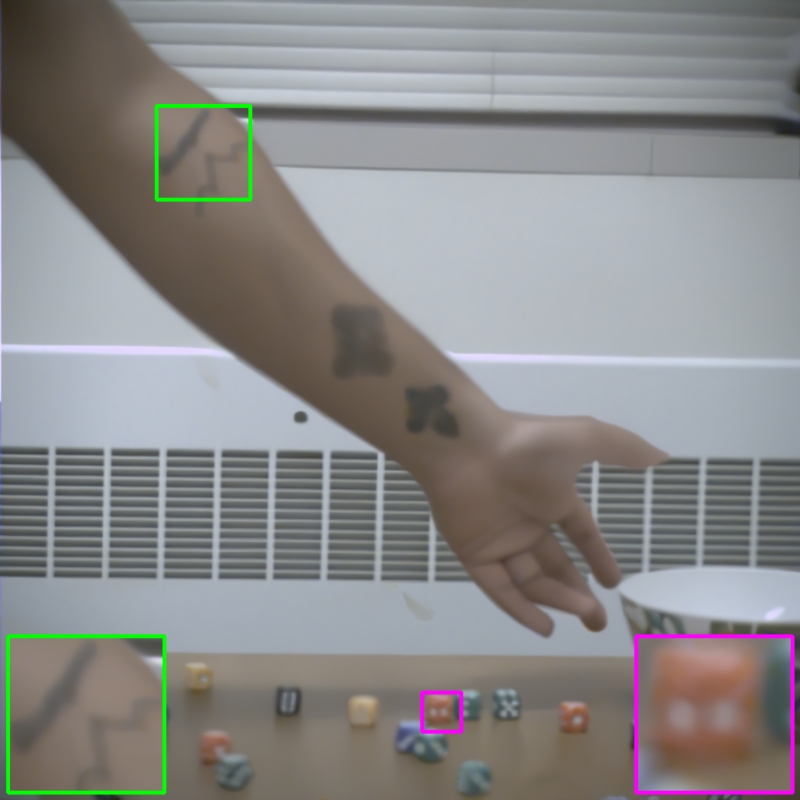}
    \includegraphics[width=0.19\textwidth,height=0.19\textwidth]{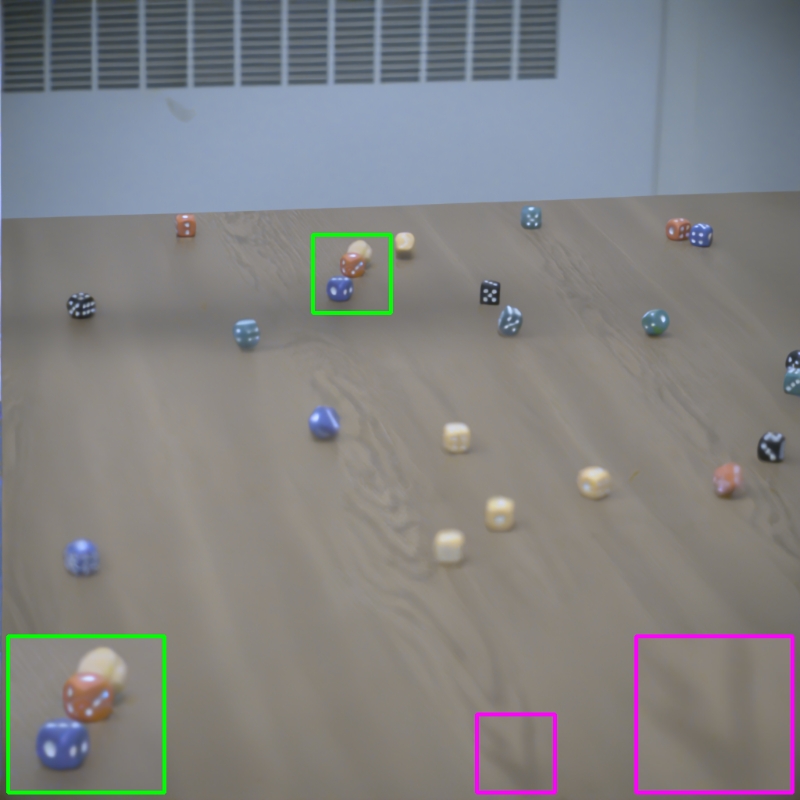}
    \includegraphics[width=0.19\textwidth,height=0.19\textwidth]{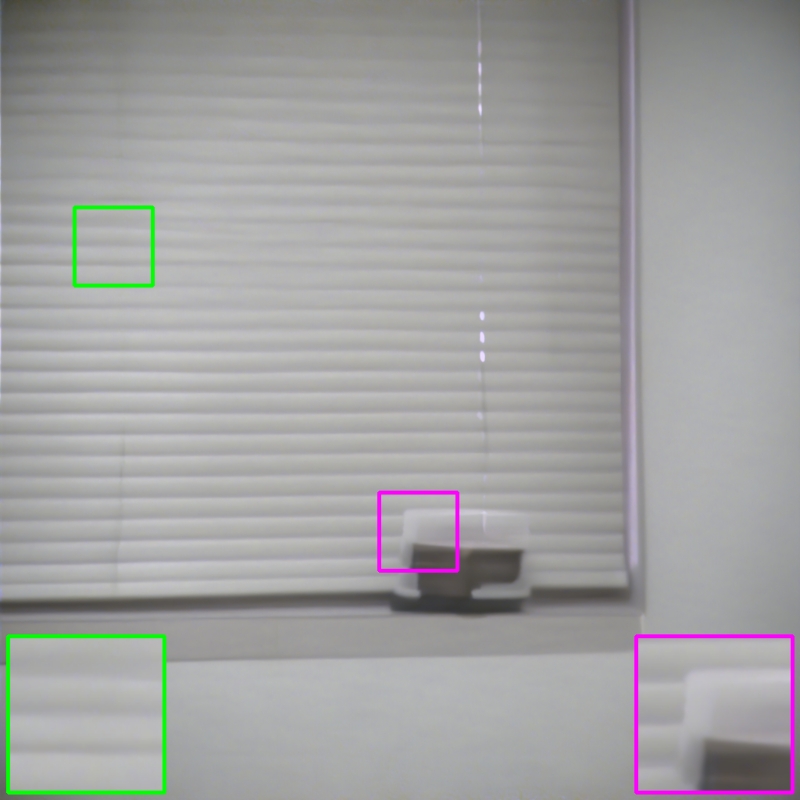}
    \includegraphics[width=0.19\textwidth,height=0.19\textwidth]{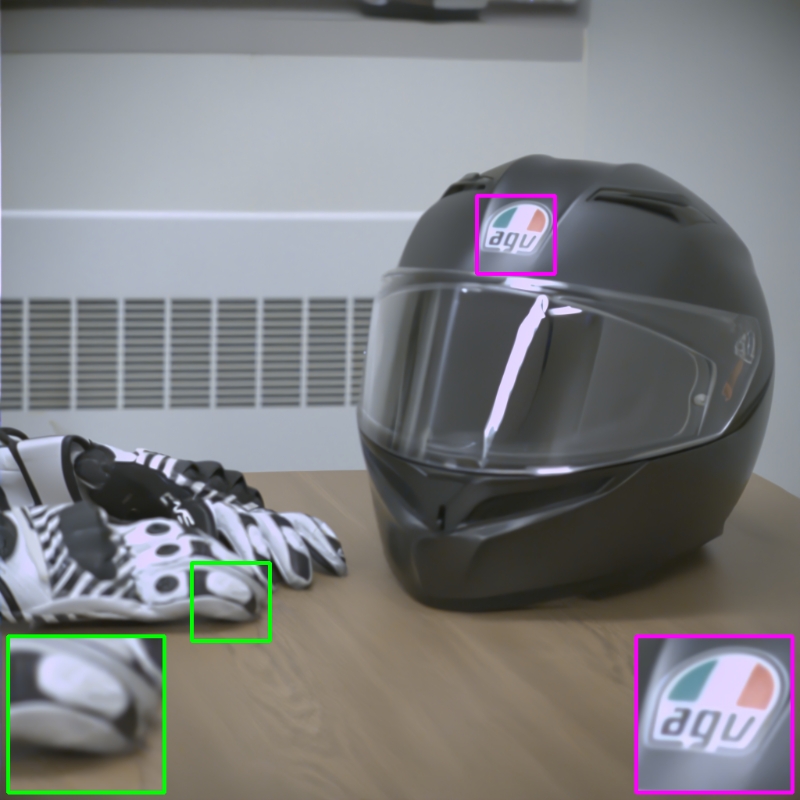}
    \caption{\textbf{Real color SPAD samples.} 
    Binary bursts captured at 6000 fps with a megapixel passive color SPAD prototype. Notice Burst-gQIR hallucinates semantic details. In column 2, the dice change color and are missing dots. In column 3, a marble not present is shown, and a dice is removed. In column 5, the sticker letters change. UnSPAD does not have similar failures.}
    
    \label{fig:grid-real}
\end{figure*}

\subsection{Real Color SPAD}

\begin{figure}
    \centering
    \includegraphics[width=\linewidth]{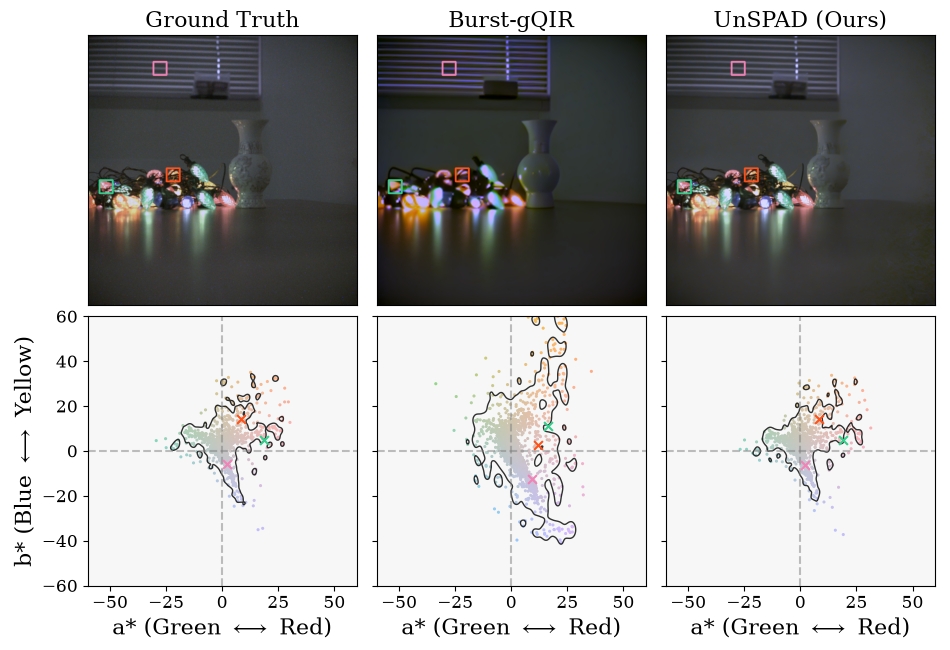}
    \caption{\textbf{Color preservation.} We project pixel values onto a 2D chromaticity plot to mitigate the effect of lightness. Specifically, we plot the $a^*$ and $b^*$ components of the CIELAB colorspace as its meant to preserve perceptual distance. We select 3 patches to highlight the difference in color. Observe UnSPAD's KDE shape matches more closely with a long average ground truth ($300$ ms), whereas gQIR over saturates as seen by the larger KDE spread.}
    \label{fig:color}

\end{figure}

We evaluated our method on real color SPAD data captured with a 1-megapixel passive color SPAD prototype at 6000 fps from gQIR~\cite{garg_2026_gqir}.
Bursts are linearly demosaicked before UnSPAD runs, and gray-world white balancing is applied as post-processing.
\autoref{fig:grid-real} shows sample reconstructions.
UnSPAD has no difficulty bridging the synthetic-to-real gap, having never seen real scenes nor real SPAD measurements while training.
Moreover, unlike Burst-gQIR, UnSPAD preserves scene color balance and minimizes hallucinations.
To analyze the color distortion caused by gQIR's generative method, we plot chromaticity in \autoref{fig:color}.

\begin{figure*}[htbp]
    \centering

    \makebox[0.49\textwidth][c]{Dynamic Scene}
    \makebox[0.49\textwidth][c]{Static Scene}

    \makebox[0.12\textwidth][c]{$t=180$ ms}
    \makebox[0.12\textwidth][c]{$193$}
    \makebox[0.12\textwidth][c]{$206$}
    \makebox[0.12\textwidth][c]{$219$}
    \makebox[0.12\textwidth][c]{$1220$}
    \makebox[0.12\textwidth][c]{$1233$}
    \makebox[0.12\textwidth][c]{$1246$}
    \makebox[0.12\textwidth][c]{$1259$}

    \rowlabel{Average}
\includegraphics[width=0.12\textwidth,height=0.12\textwidth]{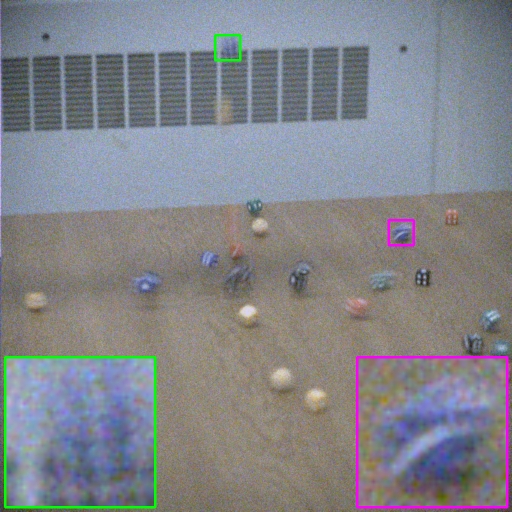}
    \includegraphics[width=0.12\textwidth,height=0.12\textwidth]{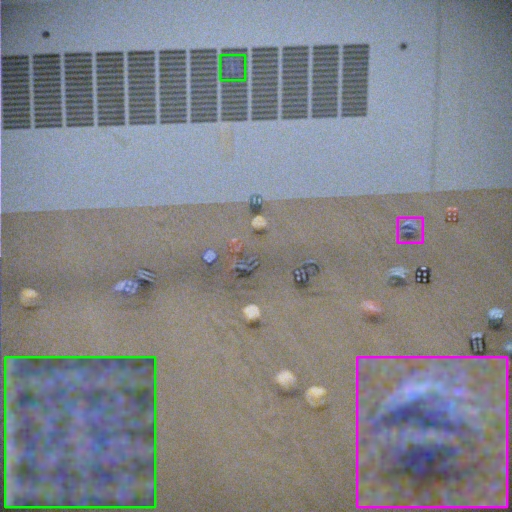}
    \includegraphics[width=0.12\textwidth,height=0.12\textwidth]{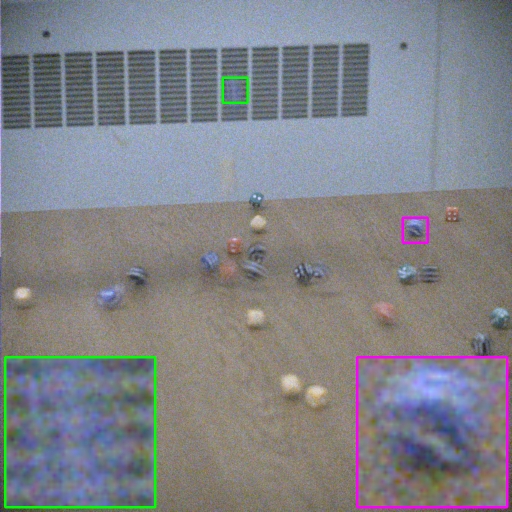}
    \includegraphics[width=0.12\textwidth,height=0.12\textwidth]{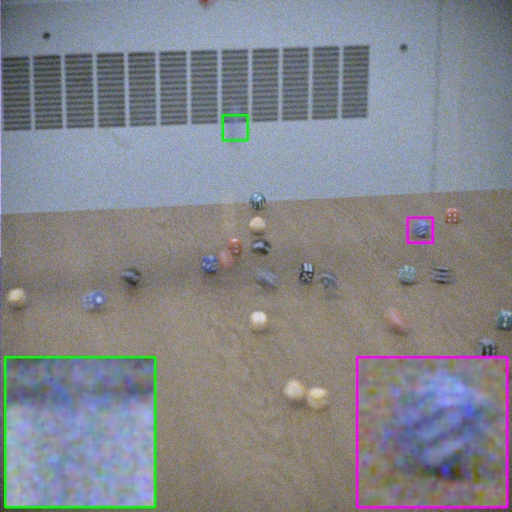}
    \includegraphics[width=0.12\textwidth,height=0.12\textwidth]{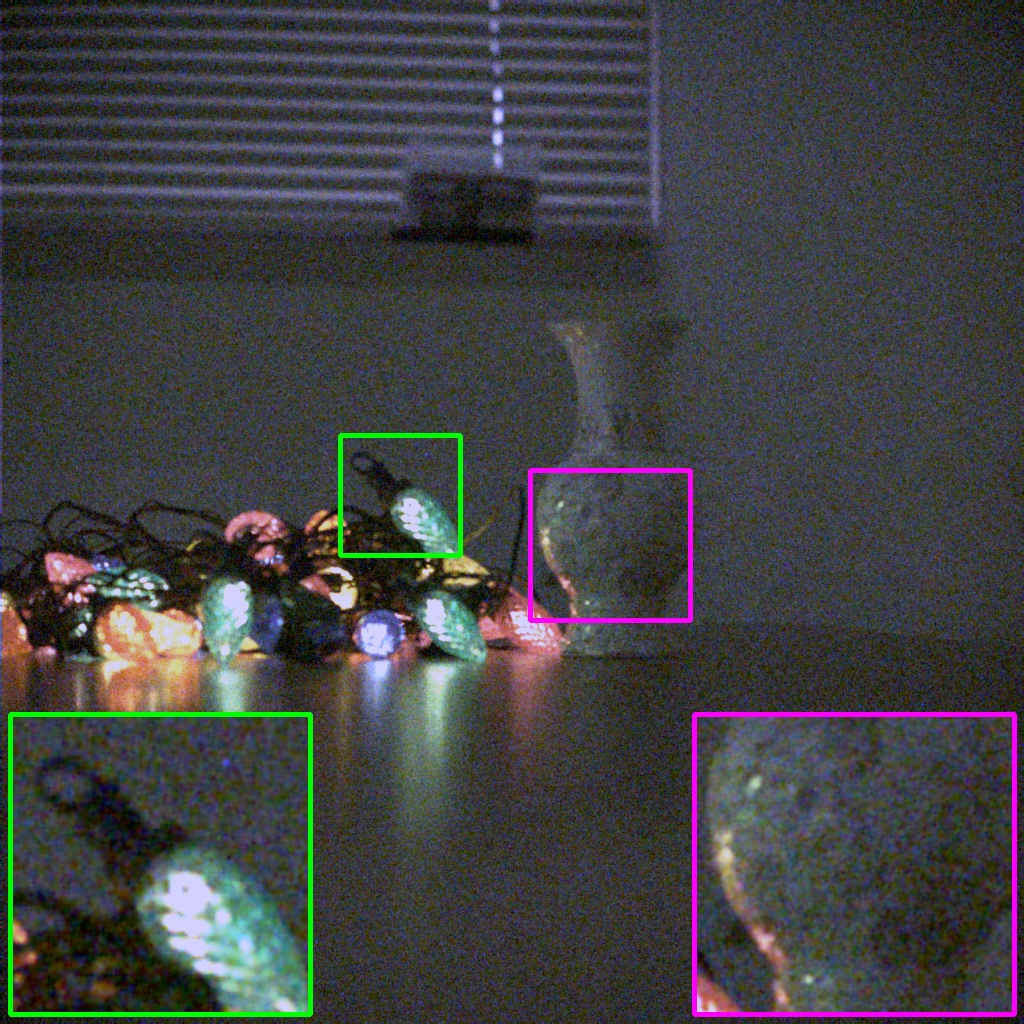}
    \includegraphics[width=0.12\textwidth,height=0.12\textwidth]{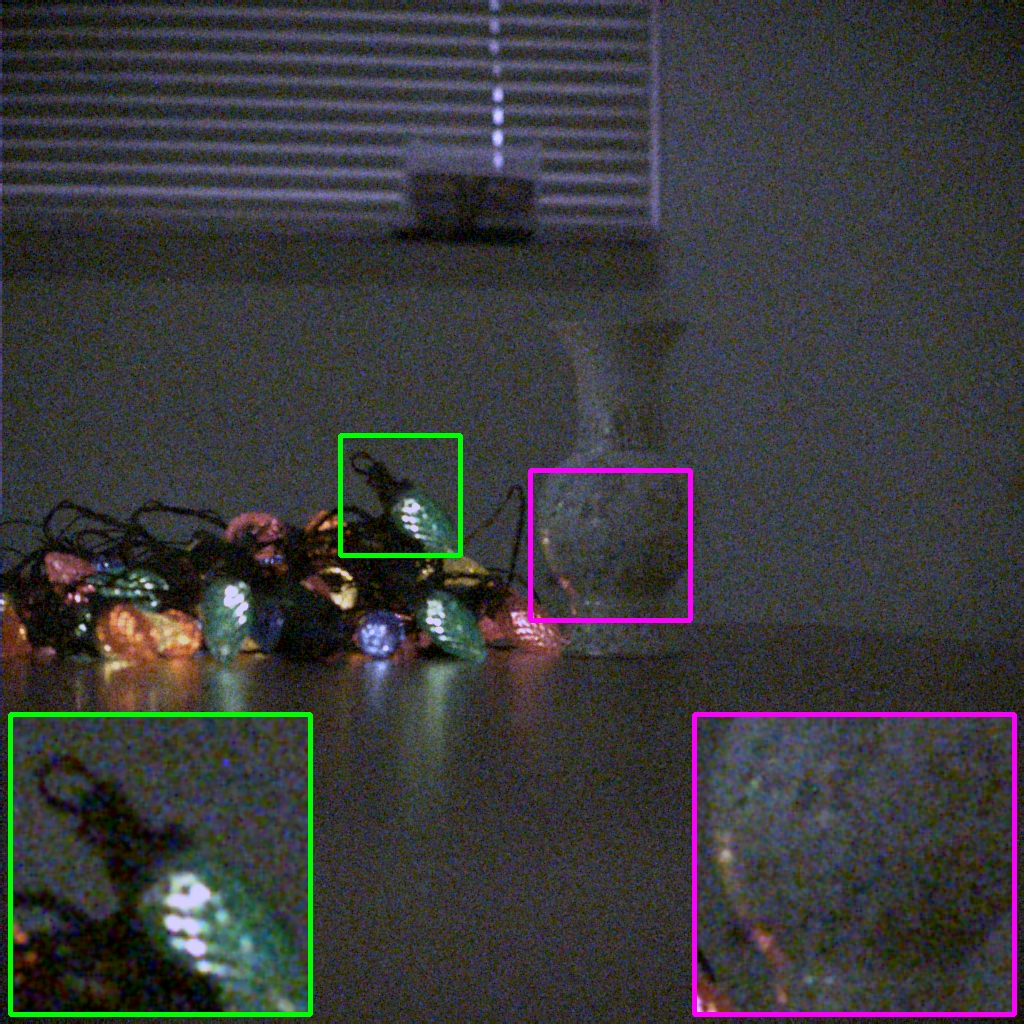}
    \includegraphics[width=0.12\textwidth,height=0.12\textwidth]{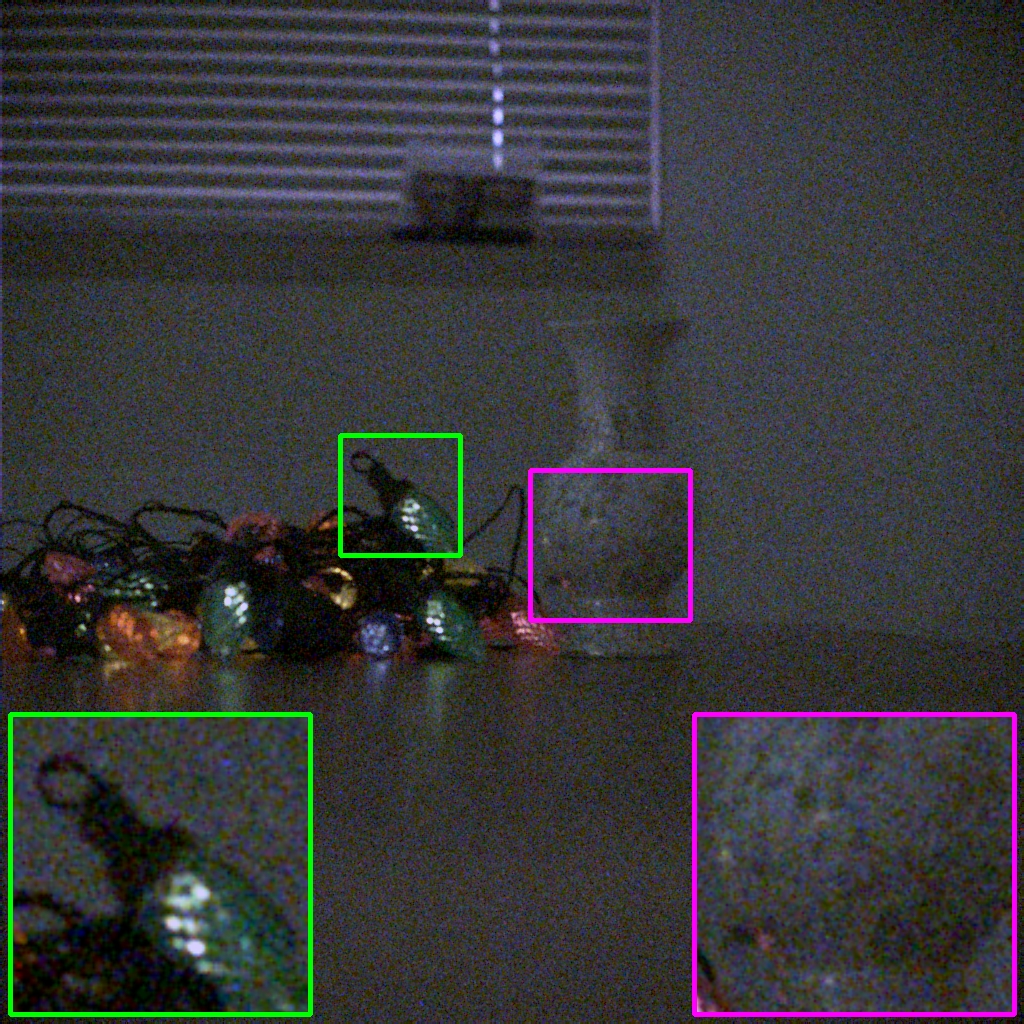}
    \includegraphics[width=0.12\textwidth,height=0.12\textwidth]{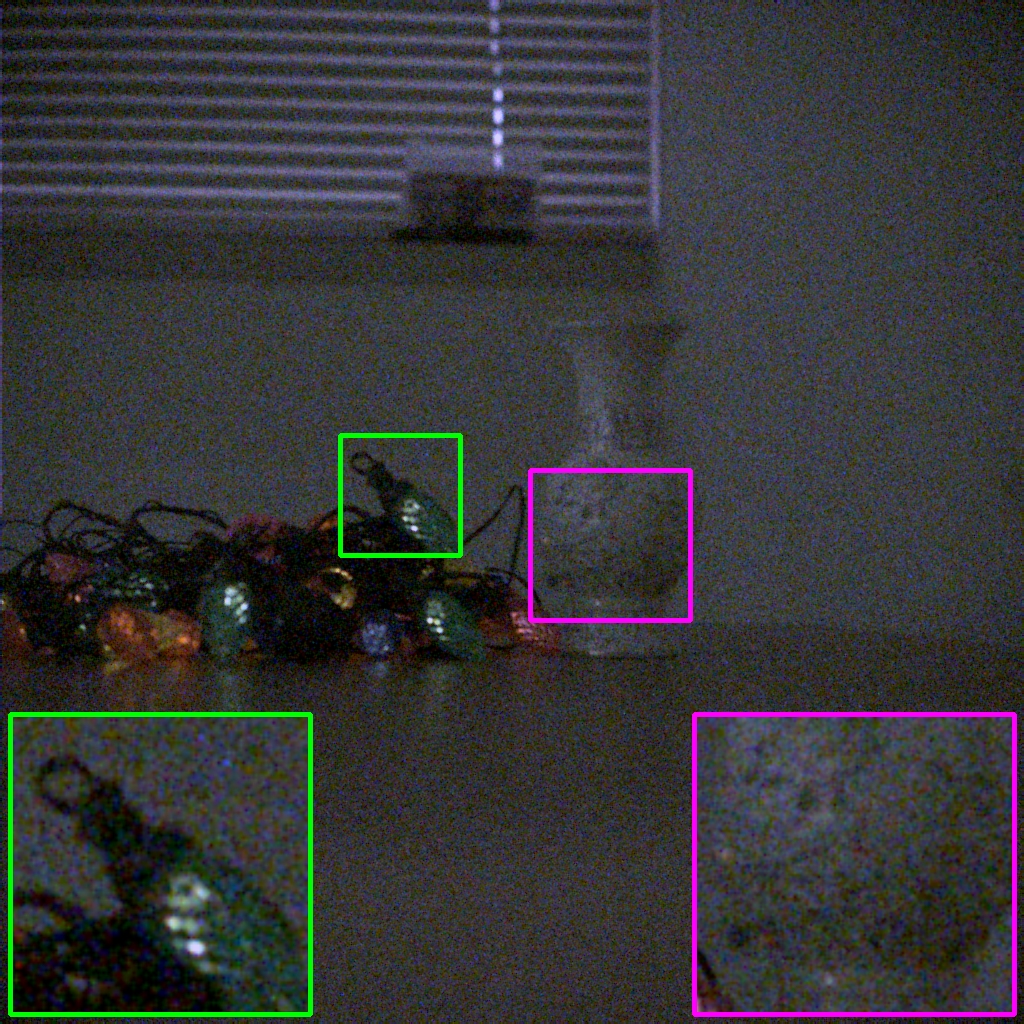}

    \smallskip

    \rowlabel{Burst-gQIR}\includegraphics[width=0.12\textwidth,height=0.12\textwidth]{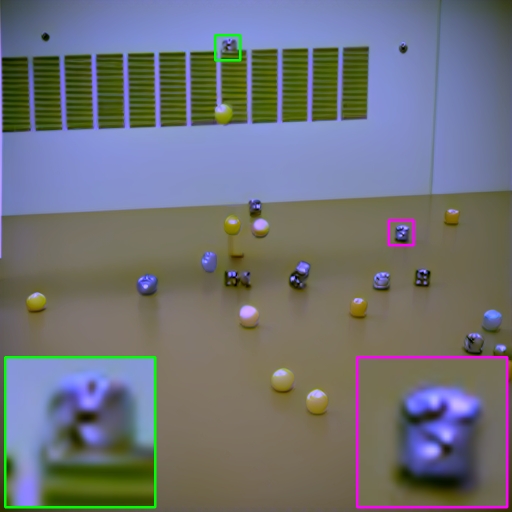}
    \includegraphics[width=0.12\textwidth,height=0.12\textwidth]{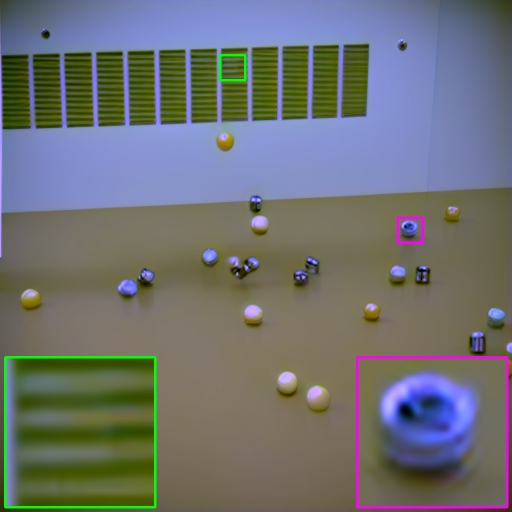}
    \includegraphics[width=0.12\textwidth,height=0.12\textwidth]{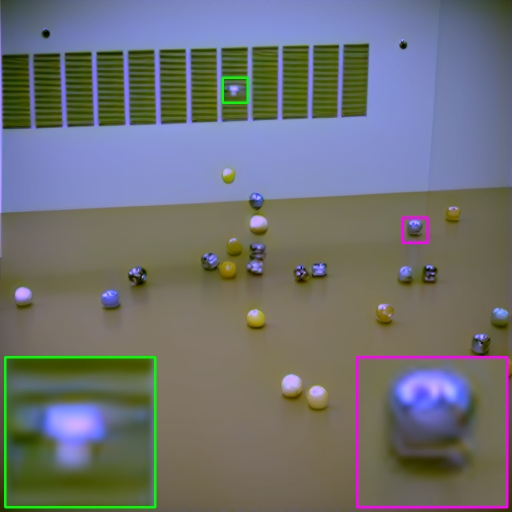}
    \includegraphics[width=0.12\textwidth,height=0.12\textwidth]{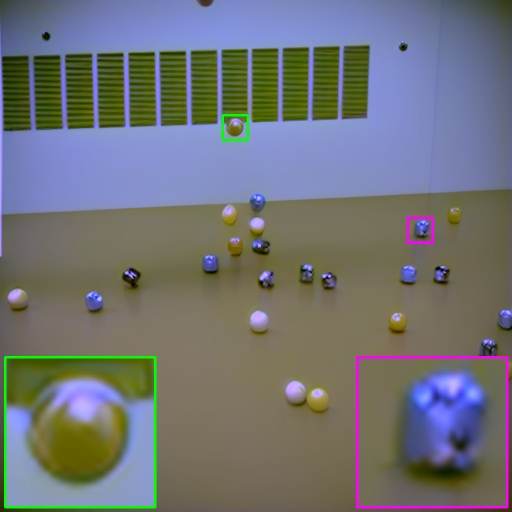}
    \includegraphics[width=0.12\textwidth,height=0.12\textwidth]{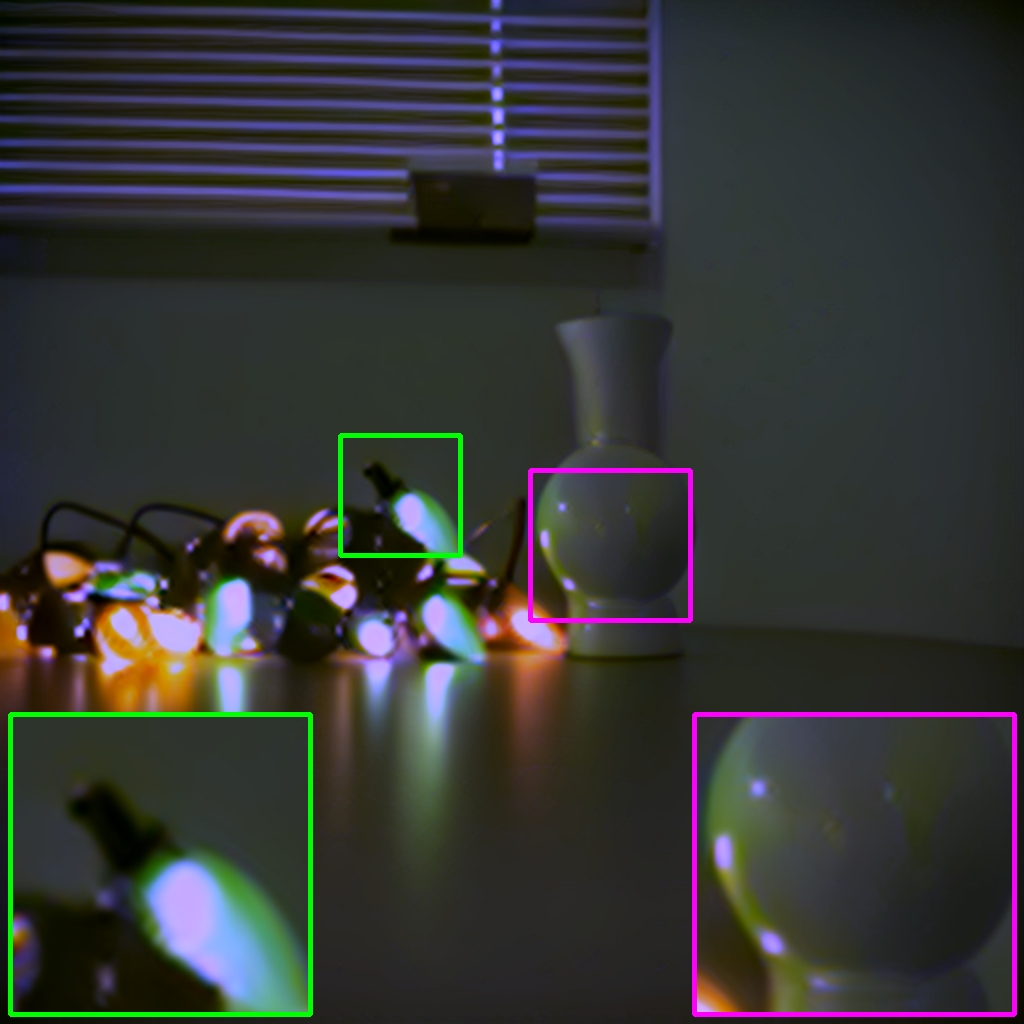}
    \includegraphics[width=0.12\textwidth,height=0.12\textwidth]{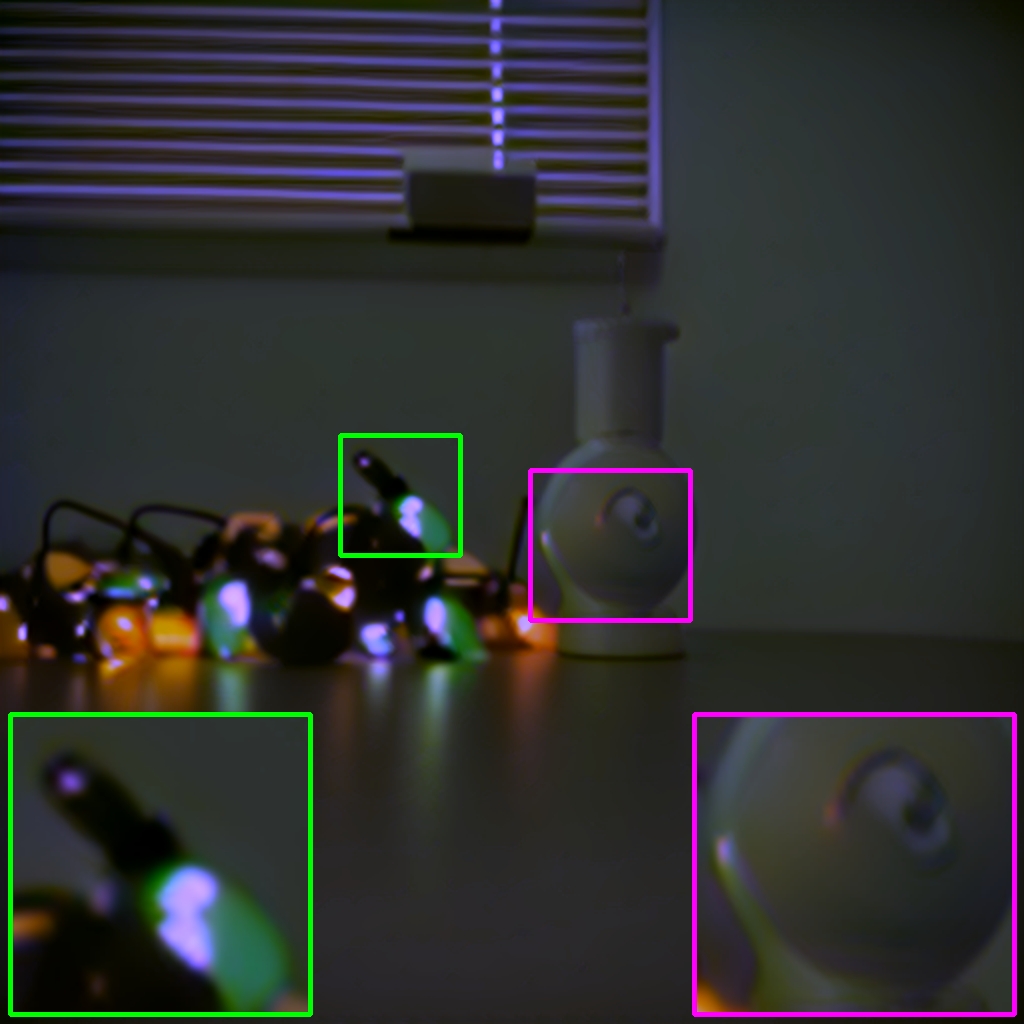}
    \includegraphics[width=0.12\textwidth,height=0.12\textwidth]{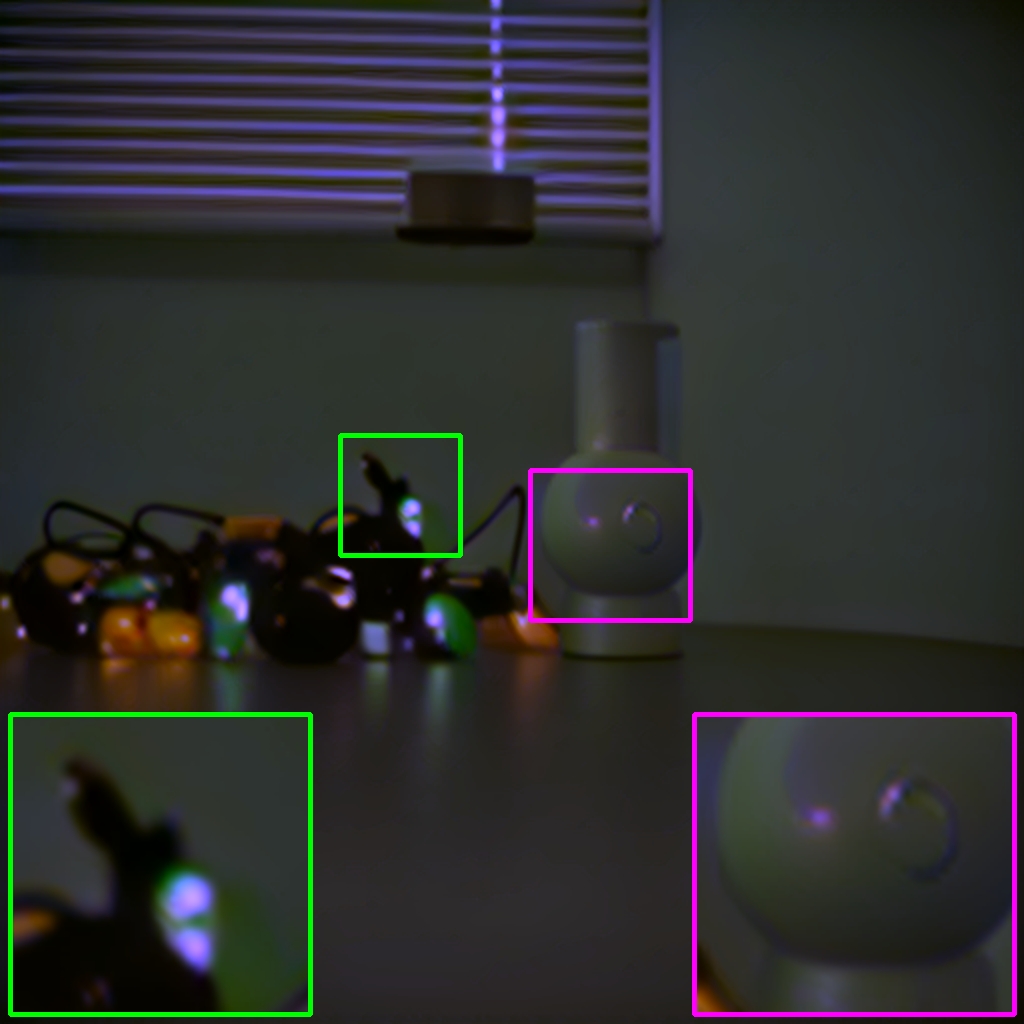}
    \includegraphics[width=0.12\textwidth,height=0.12\textwidth]{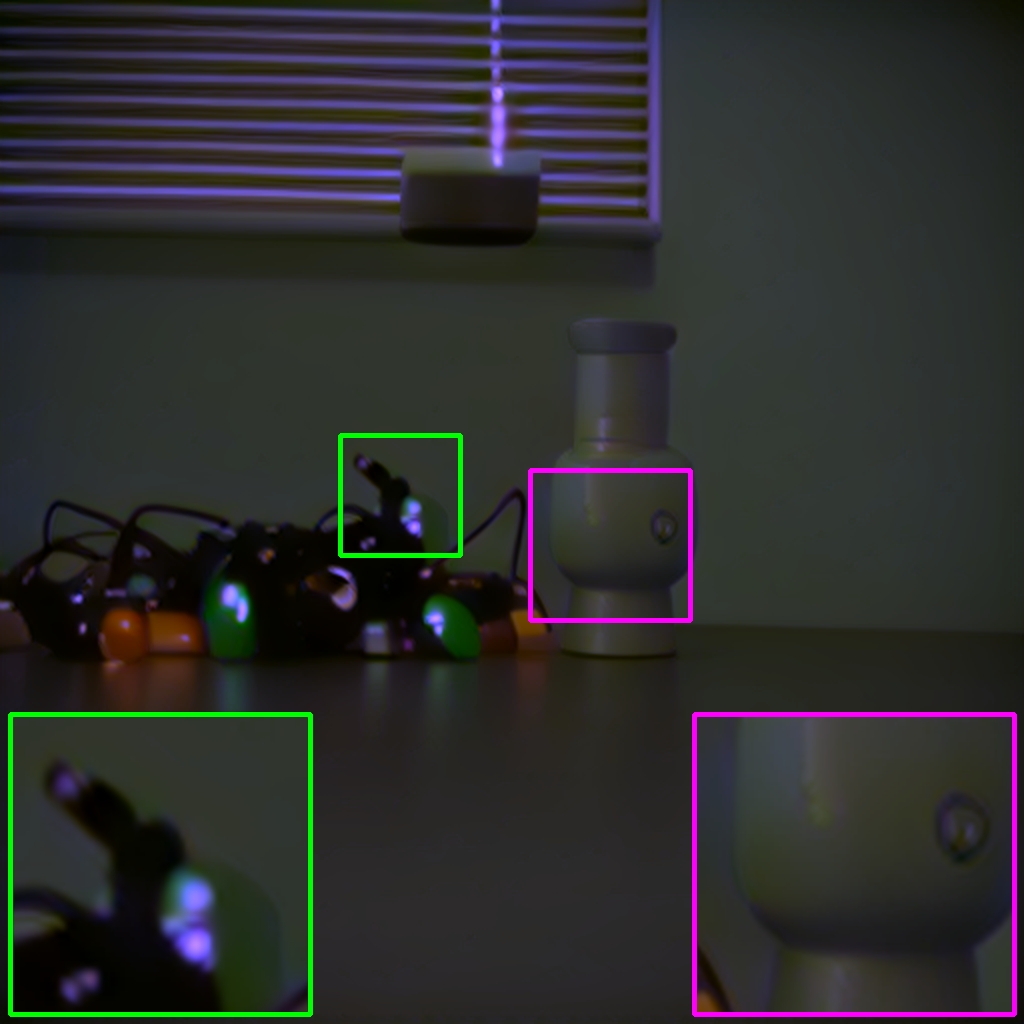}

    \smallskip
    
    \rowlabel{UnSPAD}\includegraphics[width=0.12\textwidth,height=0.12\textwidth]{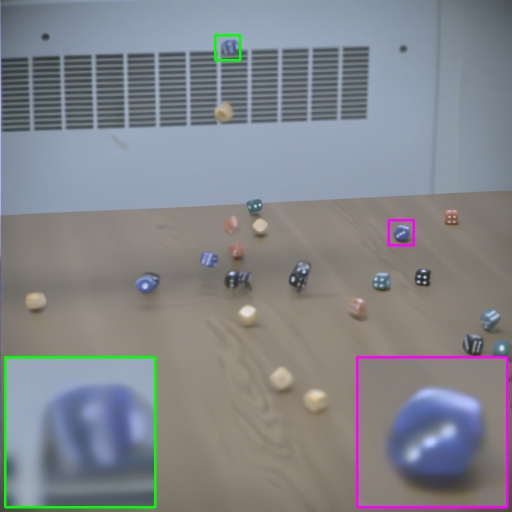}
    \includegraphics[width=0.12\textwidth,height=0.12\textwidth]{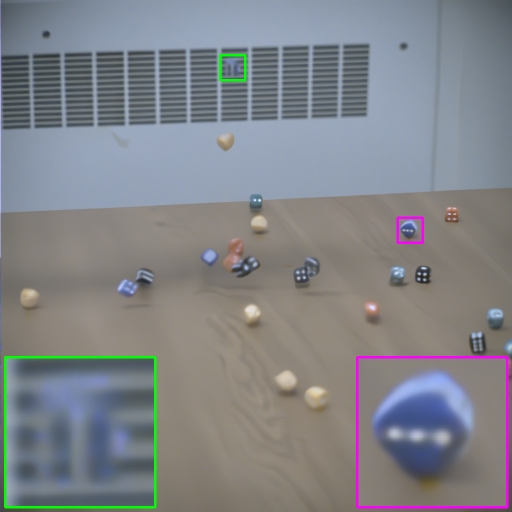}
    \includegraphics[width=0.12\textwidth,height=0.12\textwidth]{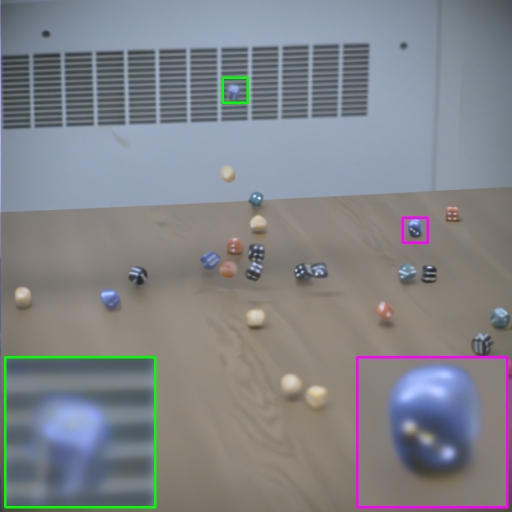}
    \includegraphics[width=0.12\textwidth,height=0.12\textwidth]{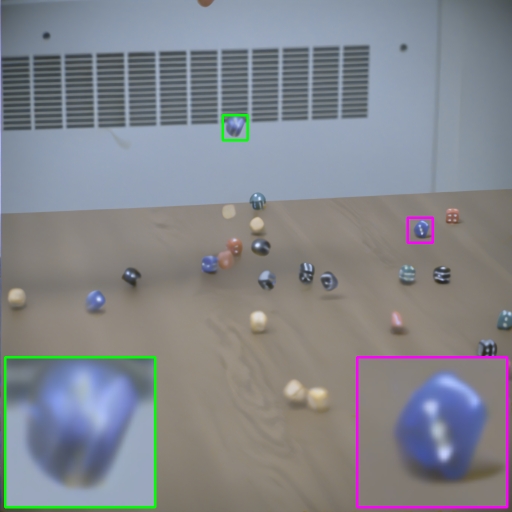}
    \includegraphics[width=0.12\textwidth,height=0.12\textwidth]{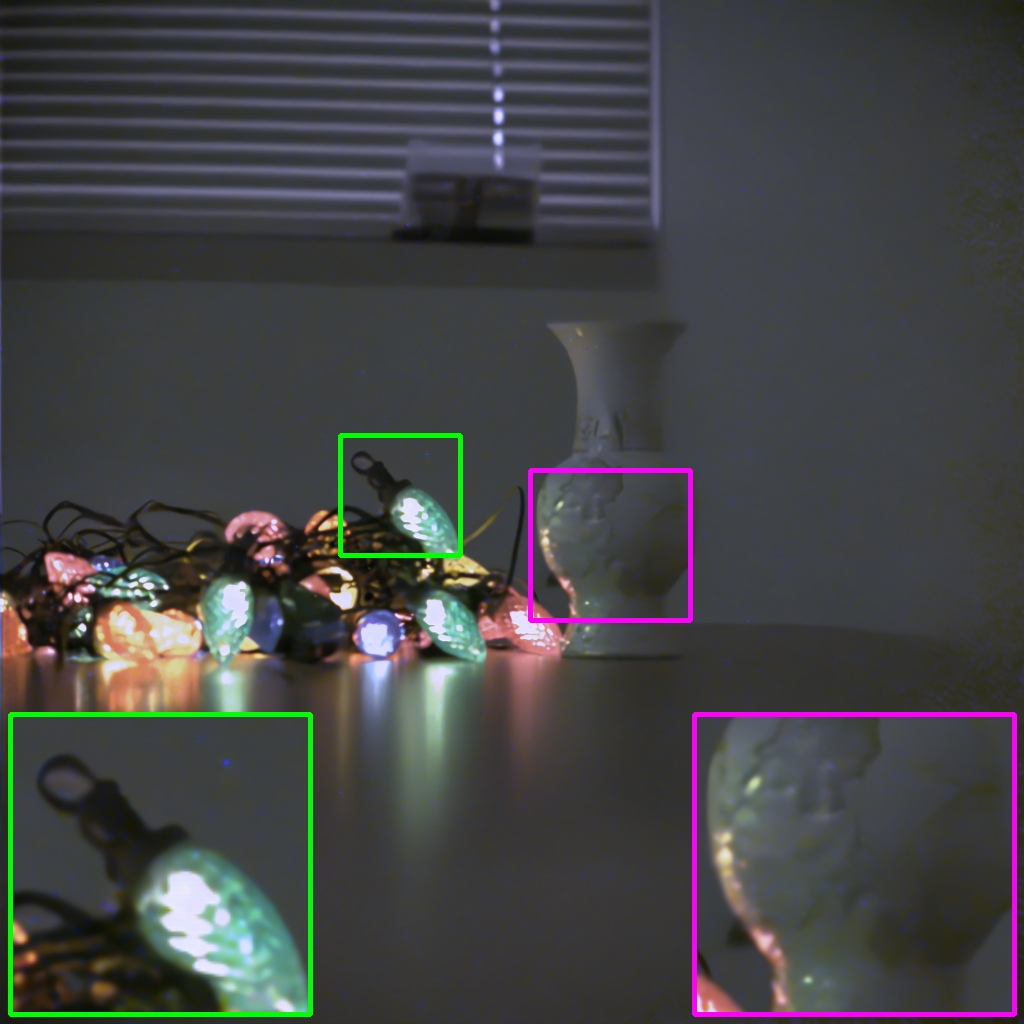}
    \includegraphics[width=0.12\textwidth,height=0.12\textwidth]{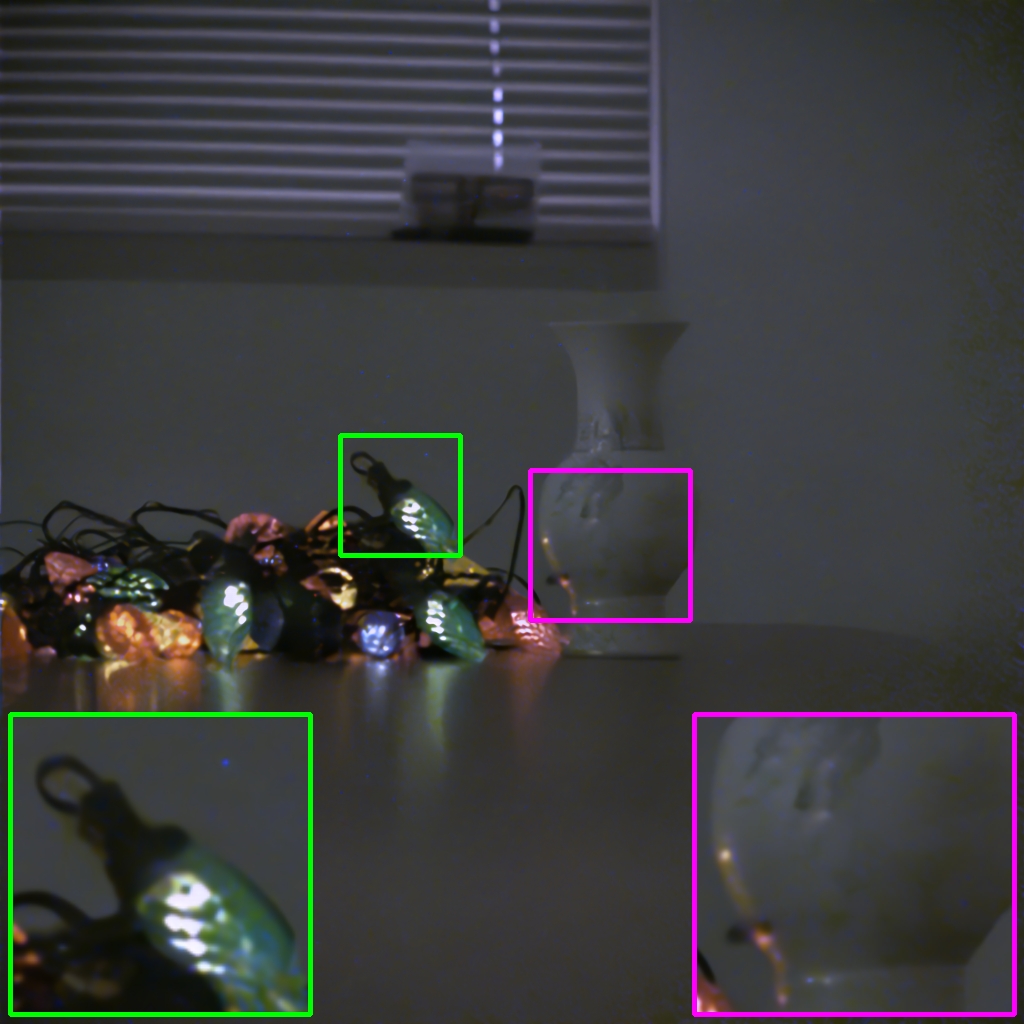}
    \includegraphics[width=0.12\textwidth,height=0.12\textwidth]{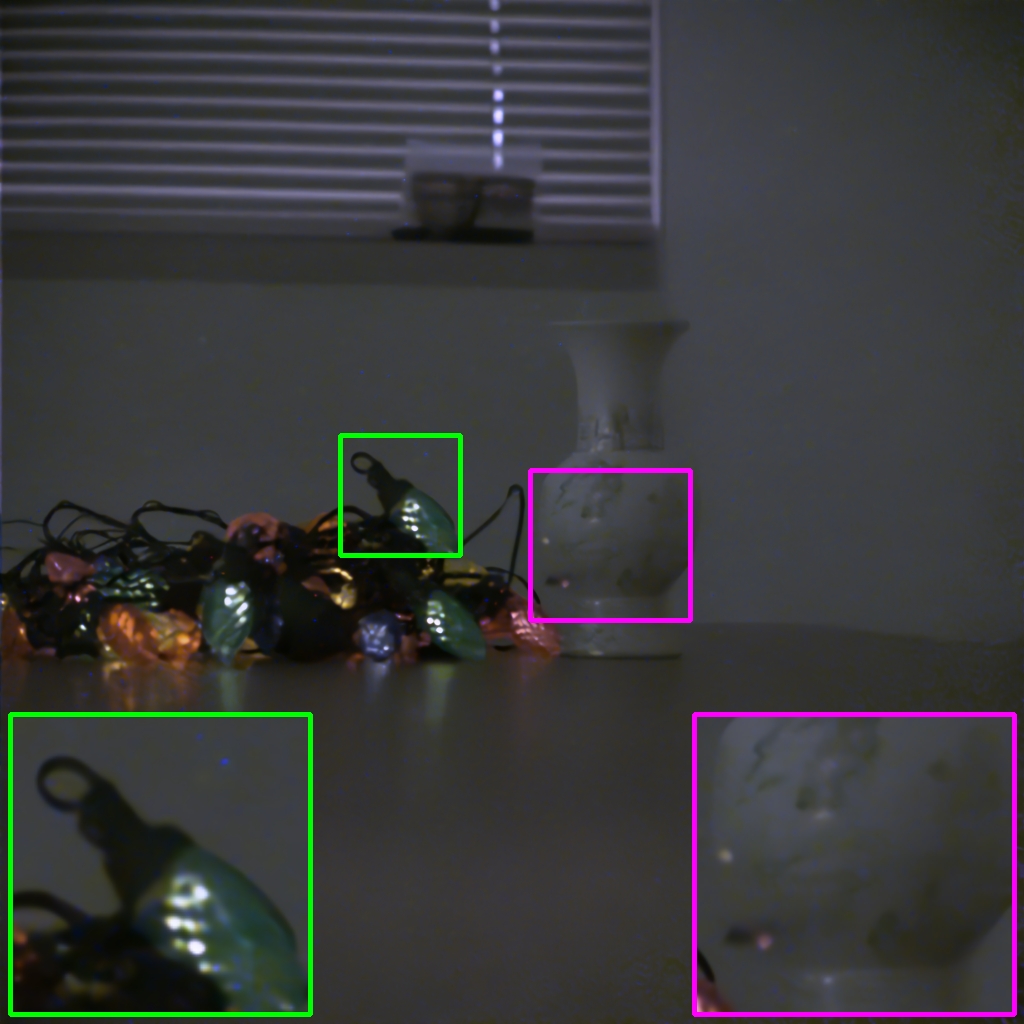}
    \includegraphics[width=0.12\textwidth,height=0.12\textwidth]{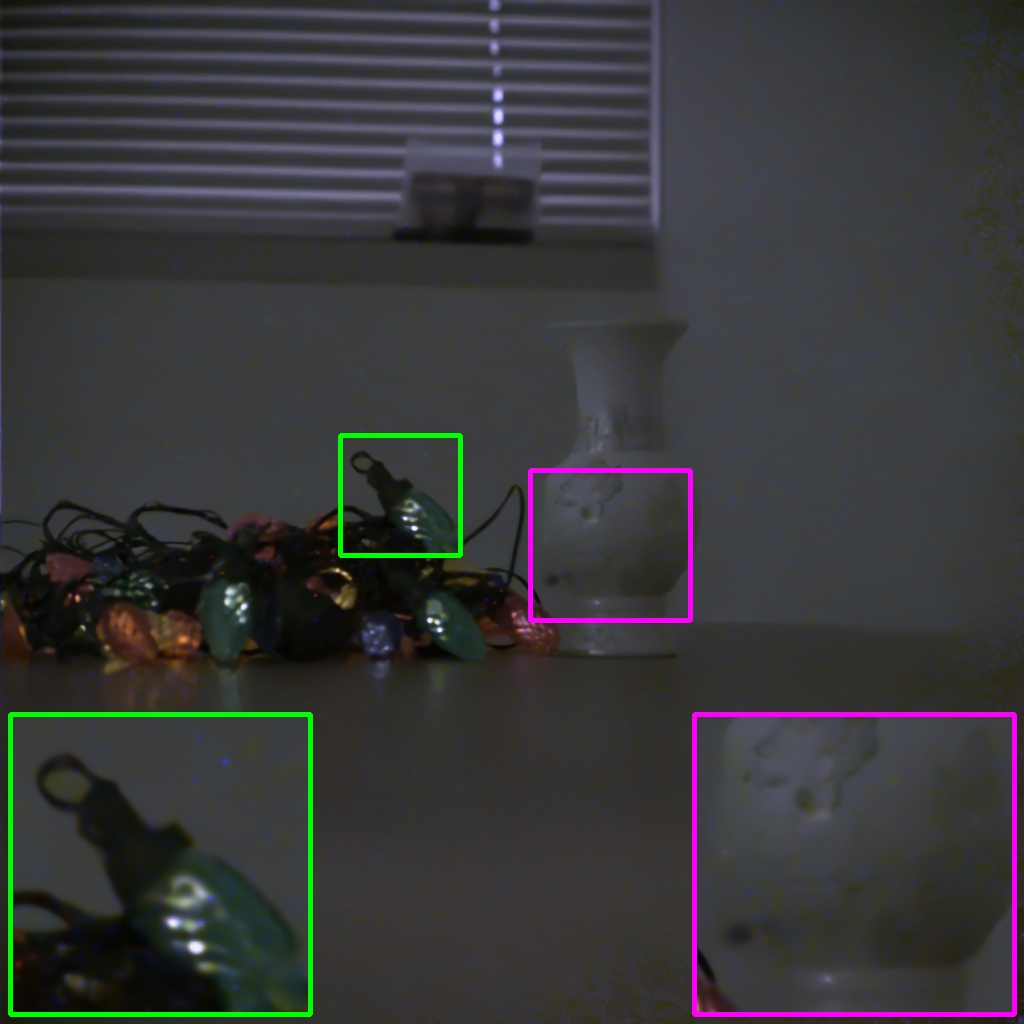}

    \caption{\textbf{Temporal consistency.}
    Two real scenes are visualized, one with motion (left) and another without (right).
    Each frame is $13$ms after the previous. 
    Since UnSPAD is not stochastic, features are better preserved over time in both cases.
    Although the diffusion model in gQIR~\cite{garg_2026_gqir} preserves some semantic information, object appearance is subject to change.
    }
    \label{fig:temporal}
\end{figure*}

To evaluate temporal consistency, we compute inter-frame similarity (ITF)~\cite{itf} and warping error ($E_{warp}^*=10^3E_{warp}$)~\cite{Lai-ECCV-2018-ewarp} on reconstructed $850$ fps video. We also use a reference-less perceptual image quality metric MANIQA~\cite{yang2022maniqa}. As shown in \autoref{tab:res-temporal}, our proposed method significantly increases temporal consistency, while maintaining image quality.
Since UnSPAD maximizes PSNR, perceptual scores like MANIQA are lower~\cite{Blau_2018_CVPR}.
Sample sequences are in \autoref{fig:temporal}.

\begin{table}[h]
    \centering
    \begin{tabular}{lccc}
        \toprule
        Method & ITF$\uparrow$ & $E^*_{warp}$$\downarrow$ & MANIQA$\uparrow$ \\
        \hline
        $1$-bit       & $7.24$          & $228.4$ & ${0.233}$ \\
        Average of $77$         & $\underline{32.03}$   & ${18.3}$    & $0.169$ \\
        Burst-gQIR~\cite{garg_2026_gqir}      & ${30.84}$          & $\underline{15.0}$ & $\mathbf{0.341}$ \\
        UnSPAD (Ours)   & $\mathbf{39.56}$ & $\mathbf{3.5}$ & $\underline{0.302}$ \\
        \bottomrule
    \end{tabular}
    \caption{\textbf{Temporal consistency and reference-less image quality} on real SPAD data. UnSPAD remains competitive in perceptual image quality, while significantly improving temporal consistency.}
    \label{tab:res-temporal}
\end{table}

\subsection{Simulated Benchmarks}

We compare against quanta burst photography methods on simulated data under two photon-counting sensors.

\paragraph{SPAD.} We evaluate on the test subset proposed by gQIR~\cite{garg_2026_gqir} with samples from XVFI~\cite{sim2021xvfi}, I2-2000fps (I2-2k)~\cite{chennuri2024quanta}, and eXtreme motion + Deforming (XD)~\cite{garg_2026_gqir} with a PPP of $3.25$. 
We adopt the identical zero-order-hold interpolation resulting in $7$ SPAD frames for every $1$ high speed CMOS frame ($7$:$1$). For fairness, we only use $77$ $1$-bit measurements in a reconstruction. As shown in \autoref{tab:res-gqir}, UnSPAD has the highest quality while minimizing distortion.
Since this setting is implausible in practice, we also adopt a more realistic setting where each ground truth frame is only sampled once ($1$:$1$), adding blur to the mini-bursts.

\begin{table}[t]
    \centering

    \begin{tabular}{lcccc}
        \toprule
        Method & PSNR$\uparrow$ & SSIM$\uparrow$ & LPIPS$\downarrow$ & E$^*$$\downarrow$ \\
        \hline
        QBP~\cite{Ma_SIGGRAPH20}        & {13.380} & {0.448} & {0.496} & {$\underline{1.0}$} \\
        QUIVER~\cite{chennuri2024quanta}     & {22.429} & {0.814} & {0.429} & {1.7} \\
        Burst-gQIR~\cite{garg_2026_gqir} & {$\underline{29.832}$} & {$\underline{0.856}$} & {$\underline{0.330}$} & {2.1} \\
        \midrule
        UnSPAD ($7$:$1$)       & {$\mathbf{33.007}$} & {$\mathbf{0.914}$} & {$\mathbf{0.193}$} & {$\mathbf{0.5}$} \\
        UnSPAD ($1$:$1$)       & {${31.618}$} & {${0.902}$} & {${0.251}$} & {${0.7}$} \\
        \bottomrule %
    \end{tabular}
    \caption{
    \textbf{Reconstruction on the Burst-gQIR dataset} from 77 $1$-bit simulated SPAD frames.
    UnSPAD outperforms existing methods across the board. We also demonstrate robustness to the setting where each ground truth frame is sampled once ($1$:$1$) rather than $7\times$. This is a more accurate model but not adopted in prior work.}
    \label{tab:res-gqir}
    \vspace{-18px}
\end{table}

\paragraph{QIS.} A QIS is another type of photon counting sensor that leverages tiny ``jots''~\cite{FossumQIS}.
Unlike a SPAD, a QIS works by integrating charge and reading out at high speeds to determine photon counts, rather than relying on avalanche breakdown multiplication.
Our technique can be adapted to work with QIS measurements even though it was not trained for this data.
Rather than estimating a photon arrival probability by averaging aligned $1$-bit frames, we instead estimate the photon count directly.
We adopt the $3$-bit QIS setting proposed in QUIVER~\cite{chennuri2024quanta}. Multi-bit QIS measurements are formed via direct quantization of integrated photon counts under Poisson arrival statistics with explicit dark current, read noise, and quantum efficiency modeling.
\mbox{\autoref{tab:res-i22k}} reports results on the I2-2k benchmark with $11$ $3$-bit QIS frames per reconstruction.
Even though UnSPAD is not trained on a QIS noise model, it still performs quite well.

Note gQIR~\cite{garg_2026_gqir} adopts an alternative QIS simulation strategy. Each measurement is constructed by aggregating $N=7$ binary SPAD-like frames, where each frame is modeled as a Bernoulli trial, resulting in a binomial observation. As a result, the binomial formulation deviates from Poisson statistics except in the extreme low-occupancy regime, where $p\xrightarrow{}0$, and introduces different variance behavior.
A further difference is that gQIR normalizes pixel values using a fixed maximum photon count, whereas QUIVER uses a data-dependent normalization based on the empirical mean intensity, leading to different saturation behavior and dynamic range scaling across scenes. We report an extra row of our method operating on gQIR assumptions (binomial).

\begin{table}[h]
    \centering

    \begin{tabular}{lccc}
        \toprule
        Method & PSNR$\uparrow$ & SSIM$\uparrow$ \\
        \hline
        Student-Teacher~\cite{Chi2020TeacherStudent} & {18.720} & {0.401} \\
        RVRT~\cite{liang2022rvrt} & {19.412} & {0.354} \\
        MemDeblur~\cite{Ji2022MemDeblur} & {19.488} & {0.387} \\
        EMVD~\cite{Maggioni_2021_CVPR}          & {20.019}   & {0.587} \\
        Spk2ImgNet~\cite{zhao2021spike} & {20.395} & {0.564} \\
        FloRNN~\cite{li2022unidirectional} & {21.034} & {0.679}  \\
        Transform Denoise~\cite{Chan2016TransformDenoise}   & {21.317} & {0.718} \\
        QBP~\cite{Ma_SIGGRAPH20}  & {21.548} & { 0.703}  \\
        QUIVER~\cite{chennuri2024quanta}   & {26.214} & {0.790} \\
        QuDi~\cite{Chennuri2025QuDI}     & {28.641} & {0.811} \\
        Burst-gQIR~\cite{garg_2026_gqir} & {$\mathbf{30.811}$} & {$\underline{0.868}$} \\
        \midrule
        UnSPAD (Poisson) & $\underline{28.654}$ & $\mathbf{0.884}$ \\
        UnSPAD (Binomial) & {36.130} & {0.934} \\
        \bottomrule
    \end{tabular}
    \caption{
    \textbf{Reconstruction on I2-2k benchmark} from $11$ simulated $3$-bit QIS frames. UnSPAD outperforms existing methods in SSIM, and is comparable in PSNR. If switched to the binomial simulator proposed in gQIR~\cite{garg_2026_gqir}, performance improves greatly, but is not directly comparable to the alternative methods.
    }
    \label{tab:res-i22k}
\end{table}

\subsection{Ablations}

We evaluate our method on the held-out samples. An additional ablation of wide burst sizes is in the supplement.

\subsubsection{Method Components}

\begin{figure*}[t]
    \centering

    \includegraphics[width=\linewidth]{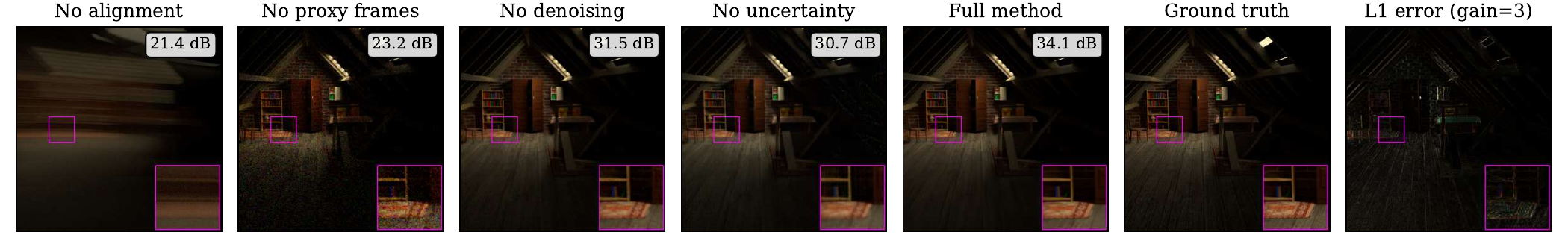}
    \caption{\textbf{Method component ablation.} Each component of UnSPAD improves image quality. Alignment with proxy frames resolves a sharp but noisy image. The uncertainty map helps guide the spatial denoiser to restore some regions more than others.}
    \label{fig:method-abla}
\end{figure*}

We remove each major component of the method in~\autoref{tab:results}. Motion compensation is the most important step, but is difficult to estimate directly with binary frames. Our proxy-frame formulation with DRUNet as a pre-denoiser is highly effective.
The uncertainty map also greatly improves the final denoiser's ability to selectively increase strength in regions that are under-observed or have error. \autoref{fig:method-abla} visualizes samples for each enhancement. Intermediate components on real data is displayed in \autoref{fig:method-intermediates}, observe how uncertainty is highest for moving objects.

\begin{table}[h]
    \centering

    \begin{tabular}{lccc}
        \toprule
        Method & PSNR$\uparrow$ & SSIM$\uparrow$ & LPIPS$\downarrow$ \\
        \hline
        Full method          & \textbf{34.95}   & \textbf{0.9636} & \textbf{0.2062} \\
        No uncertainty       & {31.49} & {0.9049} & {0.4175} \\
        No spatial denoising & {30.91} & {0.9196} & {0.4022} \\
        No proxy frames      & {18.04} & {0.4528} & {0.7066} \\
        No alignment   & {20.30} & {0.5879} & {0.6413} \\
        \bottomrule
    \end{tabular}
    \caption{\textbf{Ablation study.} Proper motion compensation is extremely important. Without the uncertainty estimate (second row), or without a spatial denoiser (third row), or if optical flow is computed directly on noisy frames (fourth row), or if no motion model is used at all (last row), image quality significantly degrades.}
    \label{tab:results}
\end{table}

\subsubsection{Denoiser Choice}

Since the proposed method is not end-to-end trained, arbitrary denoisers can be slotted into either the temporal or spatial denoising stages.
\autoref{tab:denoiser} ablates the two denoisers in our method, DRUNet~\cite{zhang2021plug} and Restormer~\cite{zamir2022restormer}, for use in either stage. DRUNet for alignment and Restormer for spatial denoising yields the highest quality result. 
We posit this is because over smoothing proxy images increases flow estimation success.
Since the spatial denoiser cost is fixed regardless of burst size, heavier weight architectures that pay more attention to details can be used.

\begin{table}[t]
\centering
\small
\begin{tabular}{lcccc}
\toprule
\multirow{2}{*}{\diagbox[width=8em]{Align}{Spatial}}
& \multicolumn{2}{c}{DRUNet}
& \multicolumn{2}{c}{Restormer} \\
\cmidrule(lr){2-3}
\cmidrule(lr){4-5}
& PSNR$\uparrow$ & Time$\downarrow$
& PSNR$\uparrow$ & Time$\downarrow$ \\
\midrule
DRUNet
& 34.96 & \textbf{8N+3}
& \textbf{35.18} & \underline{8N+24} \\
Restormer
& 34.89 & 29N+3
& \underline{35.13} & 29N+23 \\
\bottomrule
\end{tabular}
\caption{\textbf{Denoiser ablation.} PSNR is reported in dB and runtime in milliseconds for $N$ binary frames. DRUNet in the alignment stage slightly outperforms Restormer in terms of quality, while being substantially faster. For the final spatial denoising, Restormer has a larger quality increase, and since the runtime is dominated by the burst size $N$, this trade off is worthwhile.}\label{tab:denoiser}
\end{table}

\section{Conclusion}

This work achieves high-fidelity image reconstruction from a SPAD camera. 
By pre-denoising low-bit depth frames with a lightweight denoiser, we achieve robust optical-flow estimation over long intervals, allowing precise alignment of an unlimited number of frames with an estimate of uncertainty. Aligned frames can be merged and denoised into a final high-quality image.
On real color SPAD data, our method is visually significantly better than existing state of the art.
There are several avenues for future work. RAFT was used without fine-tuning, suggesting that complete re-training, or even architectural changes, would improve performance.
Alignment of binary data assumes linear motion between low bit-depth proxy frames, suggesting that a more general continuous time methodology exists.
The multiple disparate networks could be combined to further decrease inference time.
Finally, our intermediate representations of per-pixel confidence, flow, and intensity are likely to be directly beneficial to downstream applications outside of reconstruction, including robotics, tracking, and recognition.

\section*{Acknowledgments}
S.S.~was supported in part by an NDSEG Fellowship through the Air Force Office of Scientific Research under award number FA955025CB010 in the amount of \$153,984. L.B. was supported in part by the TRX Systems Postdoctoral Fellowship. This work was supported in part by ARO award no. W911NF2420113, ONR award no.~N000142312752, and NSF CAREER award no.~2339616. We gratefully acknowledge support from the Maryland Robotics Center.

{
    \small
    \bibliographystyle{ieeenat_fullname}
    \bibliography{main}

@String(CVPR= {IEEE Conf. Comput. Vis. Pattern Recog.})

@String(ICCV= {Int. Conf. Comput. Vis.})

@String(ECCV= {Eur. Conf. Comput. Vis.})

@String(TOG= {ACM Trans. Graph.})

@String(ICIP = {IEEE Int. Conf. Image Process.})

@String(CVPR  = {CVPR})

@String(ICCV  = {ICCV})

@String(ECCV  = {ECCV})

@String(TOG   = {ACM TOG})

@String(ICIP  = {ICIP})

@inproceedings{zamir2022restormer,
  title={Restormer: Efficient transformer for high-resolution image restoration},
  author={Zamir, Syed Waqas and Arora, Aditya and Khan, Salman and Hayat, Munawar and Khan, Fahad Shahbaz and Yang, Ming-Hsuan},
  booktitle={Proceedings of the IEEE/CVF conference on computer vision and pattern recognition},
  pages={5728--5739},
  year={2022}
}

@InProceedings{garg_2026_gqir,
author = {Garg, Aryan and Ma, Sizhuo and Gupta, Mohit},
title = {gQIR: Generative Quanta Image Reconstruction},
booktitle = {Proceedings of the IEEE/CVF Conference on Computer Vision and Pattern Recognition (CVPR)},
month = {June},
year = {2026}
}

@article{Ma_SIGGRAPH20,
title={Quanta Burst Photography},
author = {Ma, Sizhuo and Gupta, Shantanu and Ulku, Arin C. and Brushini, Claudio and Charbon, Edoardo and Gupta, Mohit},
journal ={ACM Transactions on Graphics (TOG)},
doi ={10.1145/3386569.3392470},
volume = {39},
number = {4},
year = {2020},
month = {7},
publisher = {ACM}
}

@inproceedings{chennuri2024quanta,
  title={Quanta Video Restoration},
  author={Chennuri, Prateek and Chi, Yiheng and Jiang, Enze and Godaliyadda, GM Dilshan and Gnanasambandam, Abhiram and Sheikh, Hamid R and Gyongy, Istvan and Chan, Stanley H},
  booktitle={European Conference on Computer Vision},
  pages={152--171},
  year={2024},
  organization={Springer}
}

@article{zhang2021plug,
  title={Plug-and-Play Image Restoration with Deep Denoiser Prior},
  author={Zhang, Kai and Li, Yawei and Zuo, Wangmeng and Zhang, Lei and Van Gool, Luc and Timofte, Radu},
  journal={IEEE Transactions on Pattern Analysis and Machine Intelligence},
  volume={44},
  number={10},
  pages={6360-6376},
  year={2021}
}

@article{zhang2018ffdnet,
  title={FFDNet: Toward a Fast and Flexible Solution for {CNN} based Image Denoising},
  author={Zhang, Kai and Zuo, Wangmeng and Zhang, Lei},
  journal={IEEE Transactions on Image Processing},
  year={2018},
}

@INPROCEEDINGS{Chennuri2025QuDI,
  author={Chennuri, Prateek and Fu, Dongdong and Chan, Stanley H.},
  booktitle={2025 IEEE International Conference on Image Processing (ICIP)}, 
  title={Quanta Diffusion}, 
  year={2025},
  volume={},
  number={},
  pages={229-234},
  doi={10.1109/ICIP55913.2025.11084414}}

@inproceedings{spynet2017,
    author = {Ranjan, Anurag and Black, Michael J.},
    title = {Optical Flow Estimation using a Spatial Pyramid Network},
    booktitle={Proceedings of the IEEE Conference on Computer Vision and Pattern Recognition},
    year = {2017}}

@software{visionsim,
    author = {Jungerman, Sacha and Gupta, Shantanu and Sadekar, Kaustubh and Leblang, Max and Gupta, Mohit},
    license = {MIT},
    month = may,
    title = {{visionsim}},
    url = {https://github.com/WISION-Lab/visionsim},
    year = {2025}
}

@inproceedings{teed2020raft,
  title     = {RAFT: Recurrent All-Pairs Field Transforms for Optical Flow},
  author    = {Teed, Zachary and Deng, Jia},
  booktitle = {Proceedings of the European Conference on Computer Vision (ECCV)},
  year      = {2020}
}

@article{singh2019recognizing,
  title={Recognizing disguised faces in the wild},
  author={Singh, Maneet and Singh, Richa and Vatsa, Mayank and Ratha, Nalini K and Chellappa, Rama},
  journal={IEEE Transactions on Biometrics, Behavior, and Identity Science},
  volume={1},
  number={2},
  pages={97--108},
  year={2019},
  publisher={IEEE}
}

@article{hdrplus,
title= {Burst photography for high dynamic range and low-light imaging on mobile cameras},author	= {Sam Hasinoff and Dillon Sharlet and Ryan Geiss and Andrew Adams and Jonathan T. Barron and Florian Kainz and Jiawen Chen and Marc Levoy},year	= {2016},URL	= {http://www.hdrplusdata.org/hdrplus.pdf},journal	= {SIGGRAPH Asia}}

@article{lafenetre23handheld,
  title={Implementing Handheld Burst Super-Resolution},
  author={Lafenetre, Jamy and Facciolo, Gabriele and Eboli, Thomas},
  journal={Image Processing On Line},
  year={2023},
}

@inproceedings{chugunov2023shakes,
title={Shakes on a Plane: Unsupervised Depth Estimation from Unstabilized Photography},
author={Chugunov, Ilya and Zhang, Yuxuan and Heide, Felix},
booktitle = {Proceedings of the IEEE/CVF Conference on Computer Vision and Pattern Recognition (CVPR)},
pages={13240--13251},
year={2023}
}

@inproceedings{chugunov2024neural,
title={Neural spline fields for burst image fusion and layer separation},
author={Chugunov, Ilya and Shustin, David and Yan, Ruyu and Lei, Chenyang and Heide, Felix},
booktitle={Proceedings of the IEEE/CVF Conference on Computer Vision and Pattern Recognition},
pages={25763--25773},
year={2024}
}

@article{Libalowlight,
author = {Liba, Orly and Murthy, Kiran and Tsai, Yun-Ta and Brooks, Tim and Xue, Tianfan and Karnad, Nikhil and He, Qiurui and Barron, Jonathan T. and Sharlet, Dillon and Geiss, Ryan and Hasinoff, Samuel W. and Pritch, Yael and Levoy, Marc},
title = {Handheld mobile photography in very low light},
year = {2019},
issue_date = {December 2019},
publisher = {Association for Computing Machinery},
address = {New York, NY, USA},
volume = {38},
number = {6},
issn = {0730-0301},
url = {https://doi.org/10.1145/3355089.3356508},
doi = {10.1145/3355089.3356508},
journal = {ACM Trans. Graph.},
month = nov,
articleno = {164},
numpages = {16}
}

@inproceedings{yang2022maniqa,
  title={MANIQA: Multi-dimension Attention Network for No-Reference Image Quality Assessment},
  author={Yang, Sidi and Wu, Tianhe and Shi, Shuwei and Lao, Shanshan and Gong, Yuan and Cao, Mingdeng and Wang, Jiahao and Yang, Yujiu},
  booktitle={Proceedings of the IEEE/CVF Conference on Computer Vision and Pattern Recognition},
  pages={1191--1200},
  year={2022}
}

@inproceedings{Lai-ECCV-2018-ewarp,
    author    = {Lai, Wei-Sheng and Huang, Jia-Bin and Wang, Oliver and Shechtman, Eli and Yumer, Ersin and Yang, Ming-Hsuan}, 
    title     = {Learning Blind Video Temporal Consistency}, 
    booktitle = {European Conference on Computer Vision},
    year      = {2018}
}

@article{itf,
	author = {Aguilar, Wilbert G. and Angulo, Cecilio},
	date = {2014/09/22},
	doi = {10.1186/1687-5281-2014-46},
	id = {Aguilar2014},
	isbn = {1687-5281},
	journal = {EURASIP Journal on Image and Video Processing},
	number = {1},
	pages = {46},
	title = {Real-time video stabilization without phantom movements for micro aerial vehicles},
	url = {https://doi.org/10.1186/1687-5281-2014-46},
	volume = {2014},
	year = {2014}}

@InProceedings{Chi2020TeacherStudent,
author="Chi, Yiheng
and Gnanasambandam, Abhiram
and Koltun, Vladlen
and Chan, Stanley H.",
editor="Vedaldi, Andrea
and Bischof, Horst
and Brox, Thomas
and Frahm, Jan-Michael",
title="Dynamic Low-Light Imaging with Quanta Image Sensors",
booktitle="Computer Vision -- ECCV 2020",
year="2020",
publisher="Springer International Publishing",
address="Cham",
pages="122--138",
isbn="978-3-030-58589-1"
}

@Article{Chan2016TransformDenoise,
AUTHOR = {Chan, Stanley H. and Elgendy, Omar A. and Wang, Xiran},
TITLE = {Images from Bits: Non-Iterative Image Reconstruction for Quanta Image Sensors},
JOURNAL = {Sensors},
VOLUME = {16},
YEAR = {2016},
NUMBER = {11},
ARTICLE-NUMBER = {1961},
URL = {https://www.mdpi.com/1424-8220/16/11/1961},
PubMedID = {27879687},
ISSN = {1424-8220},
DOI = {10.3390/s16111961}
}

@article{li2022unidirectional,
  title={Unidirectional Video Denoising by Mimicking Backward Recurrent Modules with Look-ahead Forward Ones},
  author={Li, Junyi and Wu, Xiaohe and Niu, Zhenxing and Zuo, Wangmeng},
  booktitle={ECCV},
  year={2022}
}

@inproceedings{zhao2021spike,
  title={Spk2ImgNet: Learning to Reconstruct Dynamic Scene from Continuous Spike Stream},
  author={Zhao, Jing and Xiong, Ruiqin and Liu, Hangfan and Zhang, Jian and Huang, Tiejun},
  booktitle={2021 IEEE/CVF Conference on Computer Vision and Pattern Recognition (CVPR)},
  year={2021}
}

@InProceedings{Maggioni_2021_CVPR,
  author    = {Maggioni, Matteo and Huang, Yibin and Li, Cheng and Xiao, Shuai and Fu, Zhongqian and Song, Fenglong},
  title     = {Efficient Multi-Stage Video Denoising With Recurrent Spatio-Temporal Fusion},
  booktitle = {Proceedings of the IEEE/CVF Conference on Computer Vision and Pattern Recognition (CVPR)},
  month     = {June},
  year      = {2021},
  pages     = {3466--3475},
  doi       = {10.1109/CVPR46437.2021.00347}
}

@INPROCEEDINGS{Ji2022MemDeblur,
  author={Ji, Bo and Yao, Angela},
  booktitle={2022 IEEE/CVF Conference on Computer Vision and Pattern Recognition (CVPR)}, 
  title={Multi-Scale Memory-Based Video Deblurring}, 
  year={2022},
  volume={},
  number={},
  pages={1909-1918},
  doi={10.1109/CVPR52688.2022.00196}}

@article{liang2022rvrt,
    title={Recurrent Video Restoration Transformer with Guided Deformable Attention},
    author={Liang, Jingyun and Fan, Yuchen and Xiang, Xiaoyu and Ranjan, Rakesh and Ilg, Eddy  and Green, Simon and Cao, Jiezhang and Zhang, Kai and Timofte, Radu and Van Gool, Luc},
    journal={arXiv preprint arXiv:2206.02146},
    year={2022}
}

@inproceedings{sim2021xvfi,
  title={XVFI: eXtreme Video Frame Interpolation},
  author={Sim, Hyeonjun and Oh, Jihyong and Kim, Munchurl},
  booktitle={Proceedings of the IEEE International Conference on Computer Vision (ICCV)},
  year={2021}
}

@Article{FossumQIS,
AUTHOR = {Fossum, Eric R. and Ma, Jiaju and Masoodian, Saleh and Anzagira, Leo and Zizza, Rachel},
TITLE = {The Quanta Image Sensor: Every Photon Counts},
JOURNAL = {Sensors},
VOLUME = {16},
YEAR = {2016},
NUMBER = {8},
ARTICLE-NUMBER = {1260},
URL = {https://www.mdpi.com/1424-8220/16/8/1260},
PubMedID = {27517926},
ISSN = {1424-8220},
DOI = {10.3390/s16081260}
}

@ARTICLE{bm3d,
  author={Dabov, Kostadin and Foi, Alessandro and Katkovnik, Vladimir and Egiazarian, Karen},
  journal={IEEE Transactions on Image Processing}, 
  title={Image Denoising by Sparse 3-D Transform-Domain Collaborative Filtering}, 
  year={2007},
  volume={16},
  number={8},
  pages={2080-2095},
  doi={10.1109/TIP.2007.901238}}

@article{Lebrun2013,
author = {Lebrun, M. and Buades, A. and Morel, J. M.},
title = {A Nonlocal Bayesian Image Denoising Algorithm},
journal = {SIAM Journal on Imaging Sciences},
volume = {6},
number = {3},
pages = {1665-1688},
year = {2013},
doi = {10.1137/120874989},

URL = { 
        https://doi.org/10.1137/120874989
    
},
eprint = { 
    
        https://doi.org/10.1137/120874989
    
    

}
}

@article{Tomasi1998BilateralFF,
  title={Bilateral filtering for gray and color images},
  author={Carlo Tomasi and Roberto Manduchi},
  journal={Sixth International Conference on Computer Vision (IEEE Cat. No.98CH36271)},
  year={1998},
  pages={839-846},
  url={https://api.semanticscholar.org/CorpusID:14308539}
}

@article{Condat2013tv,
	author = {Condat, Laurent},
	date = {2013/08/01},
	doi = {10.1007/s10957-012-0245-9},
	id = {Condat2013},
	isbn = {1573-2878},
	journal = {Journal of Optimization Theory and Applications},
	number = {2},
	pages = {460--479},
	title = {A Primal--Dual Splitting Method for Convex Optimization Involving Lipschitzian, Proximable and Linear Composite Terms},
	url = {https://doi.org/10.1007/s10957-012-0245-9},
	volume = {158},
	year = {2013}}

@inproceedings{
Mohan2020Robust,
title={Robust And Interpretable Blind Image Denoising Via Bias-Free Convolutional Neural Networks},
author={Sreyas Mohan and Zahra Kadkhodaie and Eero P. Simoncelli and Carlos Fernandez-Granda},
booktitle={International Conference on Learning Representations},
year={2020},
url={https://openreview.net/forum?id=HJlSmC4FPS}
}

@Inbook{zuo2018convolutional,
author={Zuo, Wangmeng and Zhang, Kai and Zhang, Lei},
editor={Bertalm{\'i}o, Marcelo},
title={Convolutional Neural Networks for Image Denoising and Restoration},
bookTitle={Denoising of Photographic Images and Video: Fundamentals, Open Challenges and New Trends},
year={2018},
publisher={Springer International Publishing},
address={Cham},
pages={93--123},
isbn={978-3-319-96029-6},
doi={10.1007/978-3-319-96029-6_4},
url={https://doi.org/10.1007/978-3-319-96029-6_4}
}

@article{zhang2017beyond,
  title={Beyond a {Gaussian} denoiser: Residual learning of deep {CNN} for image denoising},
  author={Zhang, Kai and Zuo, Wangmeng and Chen, Yunjin and Meng, Deyu and Zhang, Lei},
  journal={IEEE Transactions on Image Processing},
  year={2017},
  volume={26}, 
  number={7}, 
  pages={3142-3155}, 
}

@article{zhang2023practical,
   author = {Zhang, Kai and Li, Yawei and Liang, Jingyun and Cao, Jiezhang and Zhang, Yulun and Tang, Hao and Fan, Deng-Ping and Timofte, Radu and Gool, Luc Van},
   title = {Practical Blind Image Denoising via Swin-Conv-UNet and Data Synthesis},
   journal = {Machine Intelligence Research},
   DOI = {10.1007/s11633-023-1466-0},
   url = {https://doi.org/10.1007/s11633-023-1466-0},
   volume={20},
   number={6},
   pages={822--836},
   year={2023},
   publisher={Springer}
}

@inproceedings{
hurault2022gradient,
title={Gradient Step Denoiser for convergent Plug-and-Play},
author={Samuel Hurault and Arthur Leclaire and Nicolas Papadakis},
booktitle={International Conference on Learning Representations},
year={2022},
url={https://openreview.net/forum?id=fPhKeld3Okz}
}

@article{Bruschini2019bio,
	author = {Bruschini, Claudio and Homulle, Harald and Antolovic, Ivan Michel and Burri, Samuel and Charbon, Edoardo},
	date = {2019/09/18},
	doi = {10.1038/s41377-019-0191-5},
	id = {Bruschini2019},
	isbn = {2047-7538},
	journal = {Light: Science \& Applications},
	number = {1},
	pages = {87},
	title = {Single-photon avalanche diode imagers in biophotonics: review and outlook},
	url = {https://doi.org/10.1038/s41377-019-0191-5},
	volume = {8},
	year = {2019}}

@InProceedings{Goyal_2021_ICCV,
author = {Goyal, Bhavya and Gupta, Mohit},
title = {Photon-Starved Scene Inference Using Single Photon Cameras},
booktitle = {Proceedings of the IEEE/CVF International Conference on Computer Vision (ICCV)},
month = {October},
year = {2021},
pages = {2512-2521}
}

@inproceedings{Fabrizio2010high,
author = {Fabrizio Guerrieri and Simone Tisa and Alberto Tosi and Franco Zappa},
title = {{Single-photon camera for high-sensitivity high-speed applications}},
volume = {7536},
booktitle = {Sensors, Cameras, and Systems for Industrial/Scientific Applications XI},
editor = {Erik Bodegom and Val{\'e}rie Nguyen},
organization = {International Society for Optics and Photonics},
publisher = {SPIE},
pages = {753605},
year = {2010},
doi = {10.1117/12.838958},
URL = {https://doi.org/10.1117/12.838958}
}

@ARTICLE{Yang2012bits,
  author={Yang, Feng and Lu, Yue M. and Sbaiz, Luciano and Vetterli, Martin},
  journal={IEEE Transactions on Image Processing}, 
  title={Bits From Photons: Oversampled Image Acquisition Using Binary Poisson Statistics}, 
  year={2012},
  volume={21},
  number={4},
  pages={1421-1436},
  doi={10.1109/TIP.2011.2179306}}

@INPROCEEDINGS{Ma2023Burst,
  author={Ma, Sizhuo and Mos, Paul and Charbon, Edoardo and Gupta, Mohit},
  booktitle={2023 IEEE/CVF Winter Conference on Applications of Computer Vision (WACV)}, 
  title={Burst Vision Using Single-Photon Cameras}, 
  year={2023},
  volume={},
  number={},
  pages={5364-5374},
  doi={10.1109/WACV56688.2023.00534}}

@article{Ma_SIGGRAPH23,
title = "Seeing Photons in Color",
author = "Ma, Sizhuo and Sundar, Varun and Mos, Paul and Brushini, Claudio and Charbon, Edoardo and Gupta, Mohit",
journal = "ACM Transactions on Graphics (TOG)",
year = "2023",
publisher = "ACM"
}

@inproceedings{bit2bit,
 author = {Liu, Yehe and Krull, Alexander and Basevi, Hector and Leonardis, Ale\v{s} and Jenkins, Michael},
 booktitle = {Advances in Neural Information Processing Systems},
 doi = {10.52202/079017-2806},
 editor = {A. Globerson and L. Mackey and D. Belgrave and A. Fan and U. Paquet and J. Tomczak and C. Zhang},
 pages = {88443--88485},
 publisher = {Curran Associates, Inc.},
 title = {bit2bit: 1-bit quanta video reconstruction via self-supervised photon prediction},
 url = {https://proceedings.neurips.cc/paper_files/paper/2024/file/a128e43ed4708862aec51549b0050fc6-Paper-Conference.pdf},
 volume = {37},
 year = {2024}
}

@InProceedings{Blau_2018_CVPR,
author = {Blau, Yochai and Michaeli, Tomer},
title = {The Perception-Distortion Tradeoff},
booktitle = {Proceedings of the IEEE Conference on Computer Vision and Pattern Recognition (CVPR)},
month = {June},
year = {2018}
}
}

\clearpage
\setcounter{page}{1}

\setcounter{section}{0}
\setcounter{figure}{0}
\setcounter{table}{0}
\setcounter{equation}{0}

\maketitlesupplementary

In this supplementary material, we show the remaining real scene comparison (\S\ref{sec:supp:cmp}), 
preservation of photography effects (\S\ref{sec:supp:defocus}), 
show more qualitative ablations (\S\ref{sec:supp:abla}), 
show UnSPAD over extreme motions (\S\ref{sec:supp:motion}), 
evaluate UnSPAD's pixel-level accuracy (\S\ref{sec:supp:pix-acc}), 
discuss changes from QBP (\S\ref{sec:supp:qbp}), 
provide a comparison with bit2bit (\S\ref{sec:supp:bit2bit}), 
and describe evaluation metrics (\S\ref{sec:supp:metrics}).

\section{Real Color SPAD Comparison}\label{sec:supp:cmp}
\begin{figure*}
     \centering

    \makebox[0.23\linewidth][c]{$1$-bit}
    \makebox[0.23\linewidth][c]{Average of $77$}
    \makebox[0.23\linewidth][c]{Burst-gQIR~\cite{garg_2026_gqir}}
    \makebox[0.23\linewidth][c]{UnSPAD (Ours)}
    \includegraphics[width=0.23\textwidth]{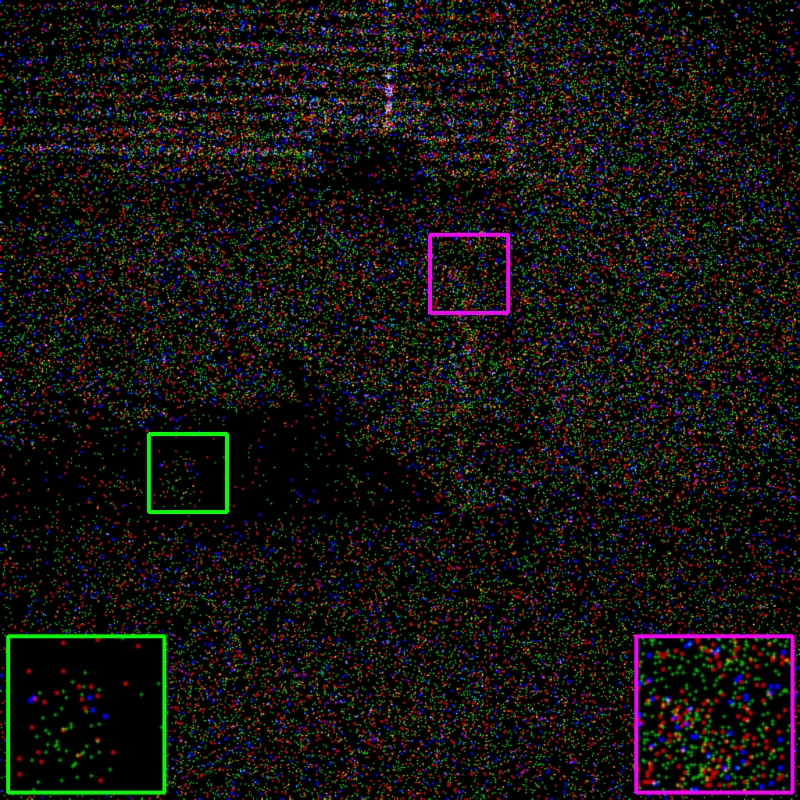}
    \includegraphics[width=0.23\textwidth]{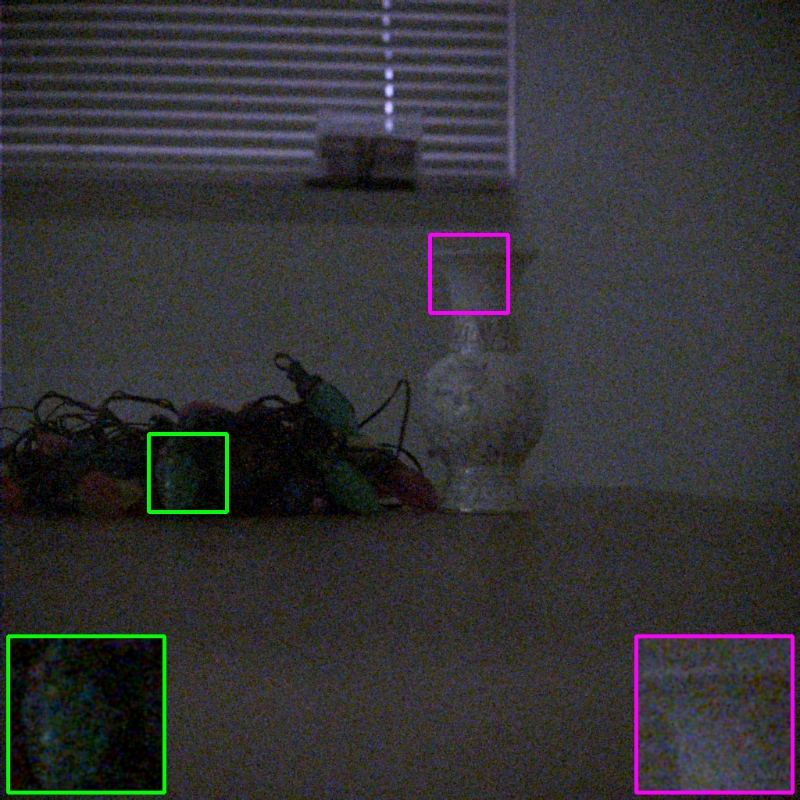}
    \includegraphics[width=0.23\textwidth]{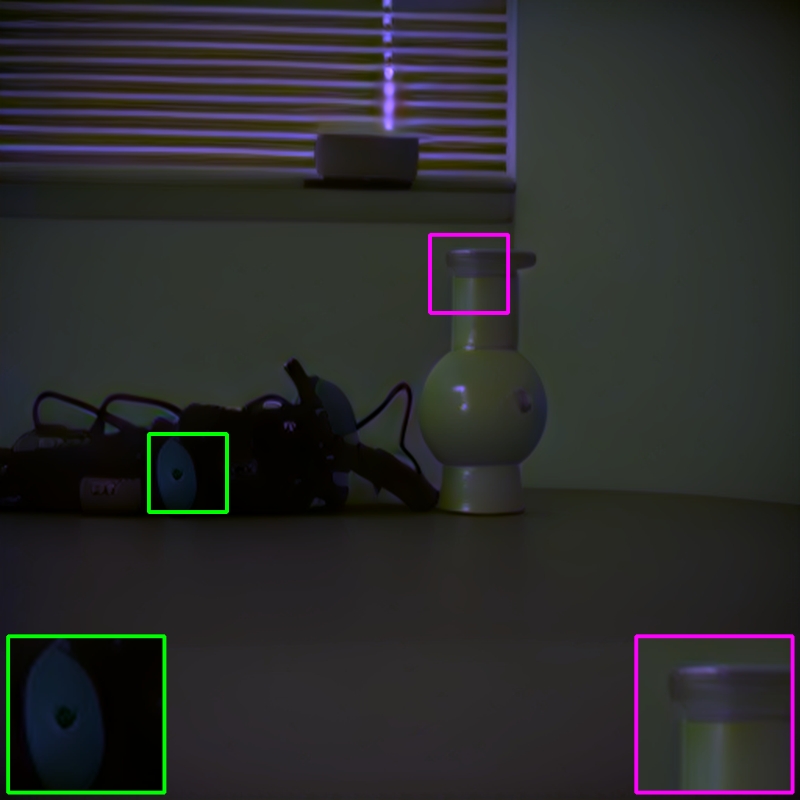}
    \includegraphics[width=0.23\textwidth]{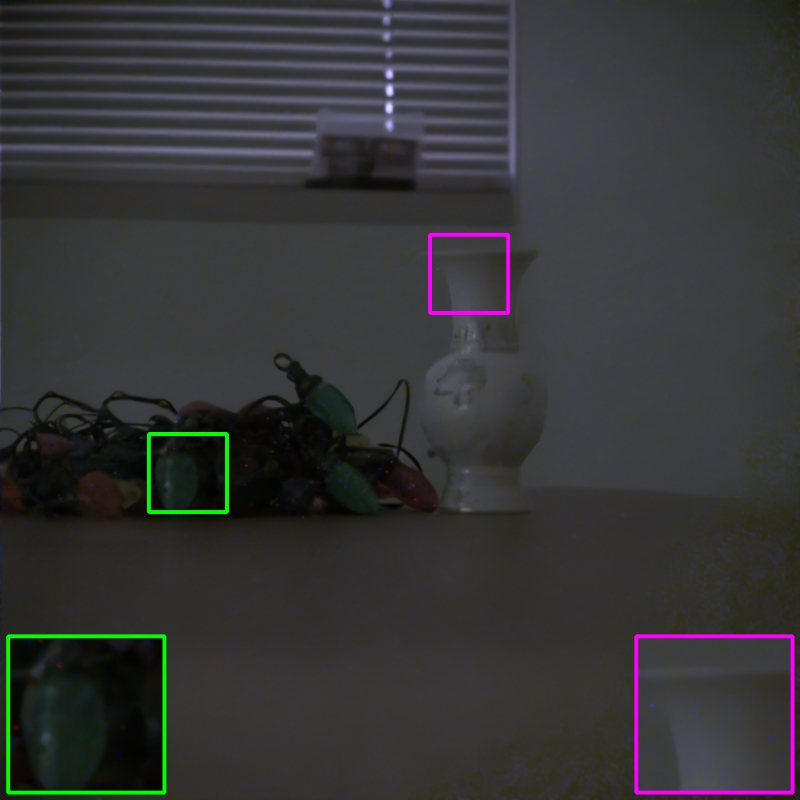}
    \caption{\textbf{Additional comparison.} On the Christmas lights scene, when the lights are off, generative methods hallucinate different color bulbs (green box), or change the vase shape (pink box), UnSPAD on the other had preserves these details. }
\label{fig:remaining}
\end{figure*}

We include the sixth real color SPAD scene (Christmas lights) omitted from the main paper for space in \autoref{fig:remaining}.

\section{Photography Effect Preservation}\label{sec:supp:defocus}
\begin{figure*}
     \centering

    \makebox[0.23\linewidth][c]{$1$-bit}
    \makebox[0.23\linewidth][c]{Average of $77$}
    \makebox[0.23\linewidth][c]{Burst-gQIR~\cite{garg_2026_gqir}}
    \makebox[0.23\linewidth][c]{UnSPAD (Ours)}
    \includegraphics[width=0.23\textwidth]{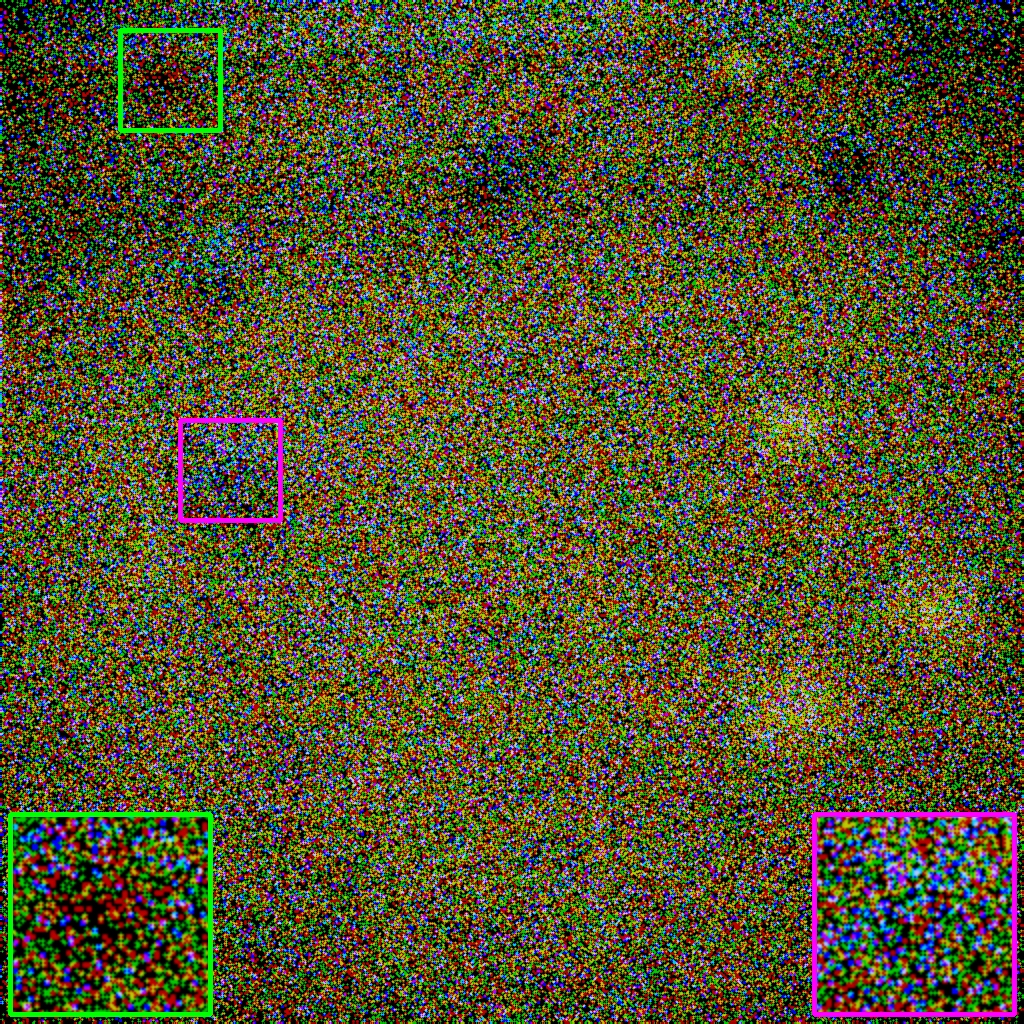}
    \includegraphics[width=0.23\textwidth]{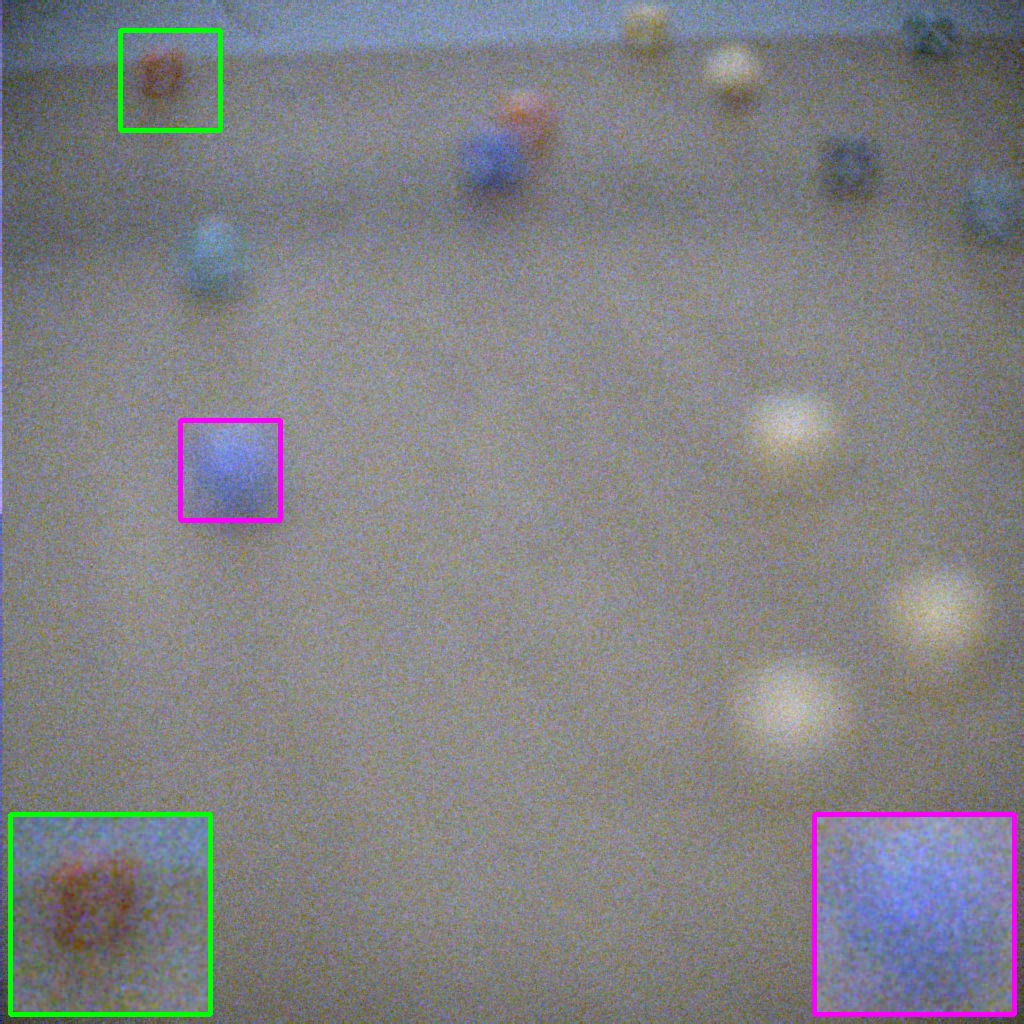}
    \includegraphics[width=0.23\textwidth]{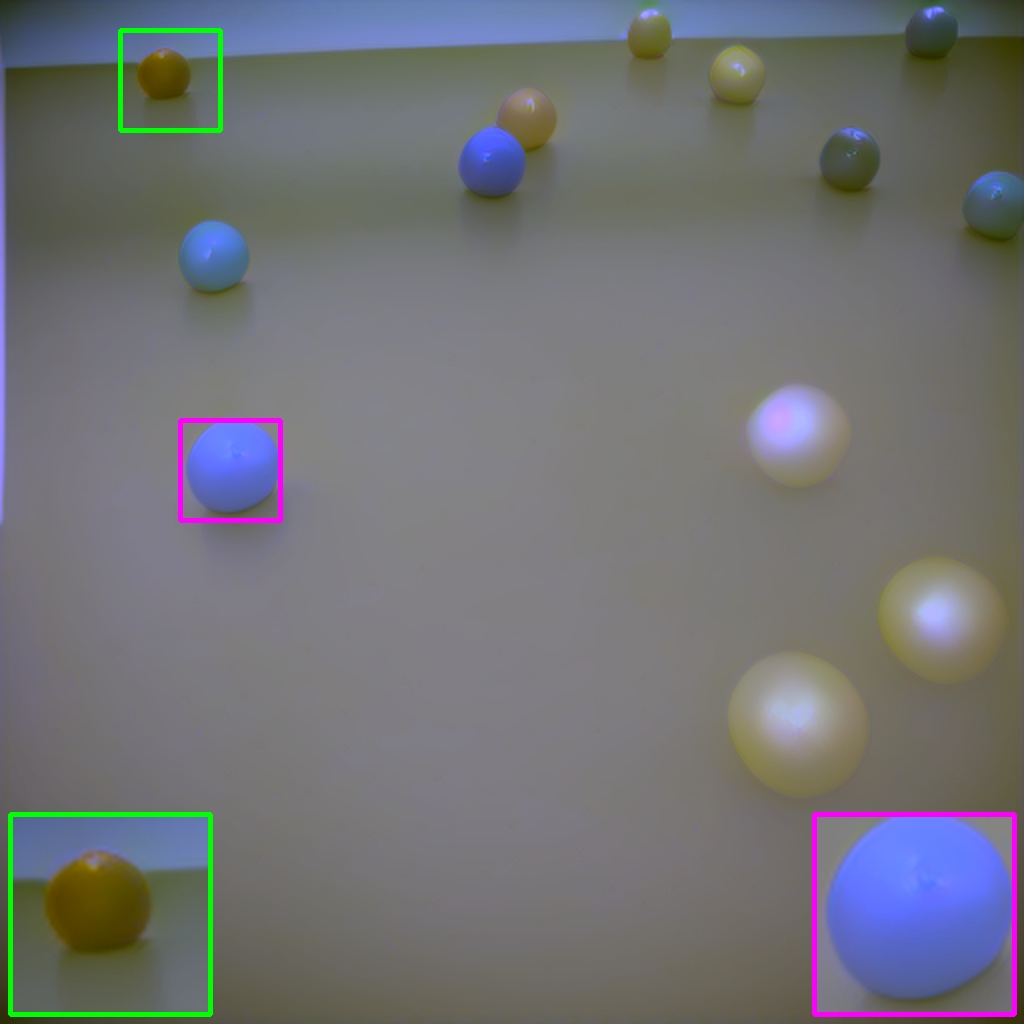}
    \includegraphics[width=0.23\textwidth]{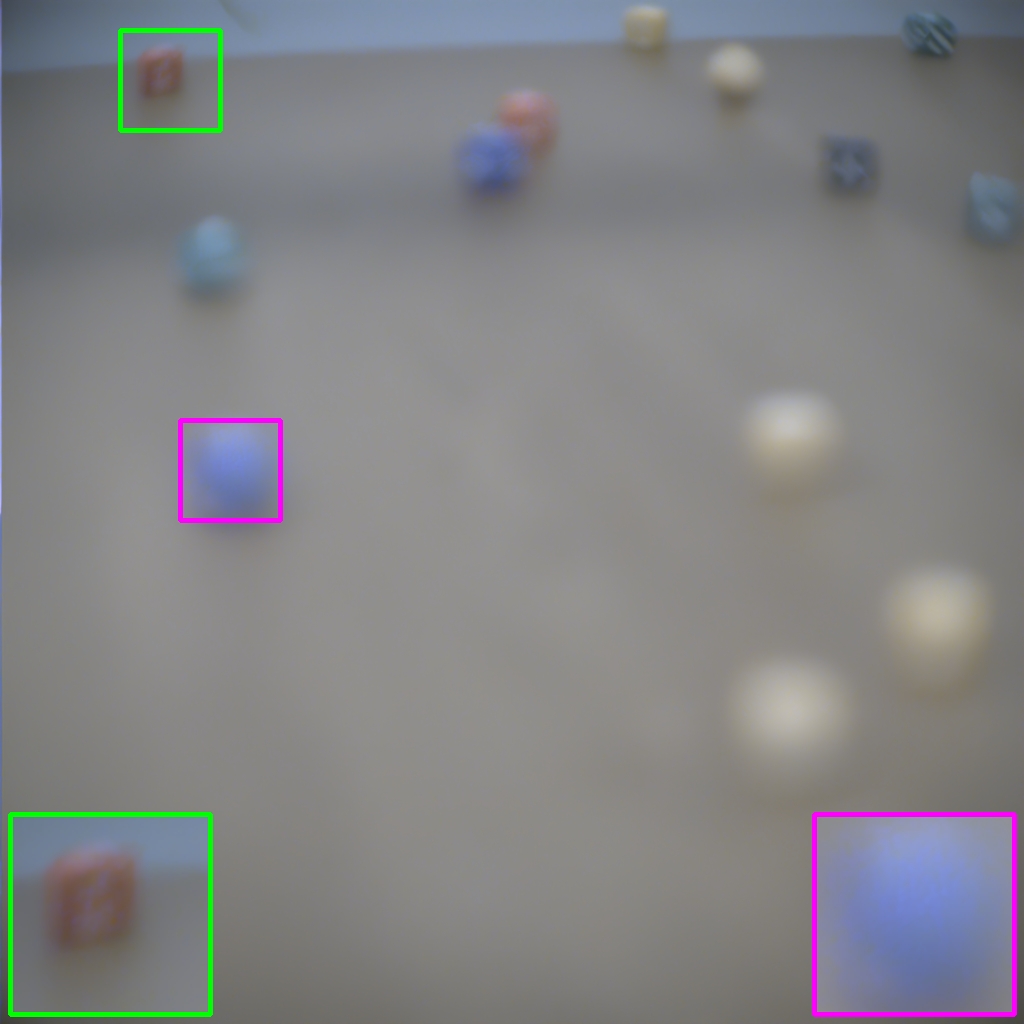}
    \caption{\textbf{Photography effects.} UnSPAD preserves natural photographic effects such as bokeh and defocus blur, recovering images consistent with the camera's optics and acquisition settings. In contrast, generative methods tend to hallucinate sharp details, producing visually appealing but potentially semantically inaccurate reconstructions. Burst-gQIR recovers blurred dice as marbles.}
\label{fig:defocus}
\end{figure*}

In \autoref{fig:defocus}, we show UnSPAD produces reconstructions that are faithful to the imaging process, preserving optical effects such as defocus blur. In contrast, generative methods often replace naturally blurred regions with artificially sharpened, semantically plausible content.

\section{Qualitative Method Ablation}\label{sec:supp:abla}

In \autoref{fig:grid-supp}, we show more sample reconstructions on the Single Photon Challenge dataset. UnSPAD can successfully align frames over long sequences.

\section{Widening Burst Size}\label{sec:supp:wide}

\begin{figure}
    \centering
    \includegraphics[width=\linewidth]{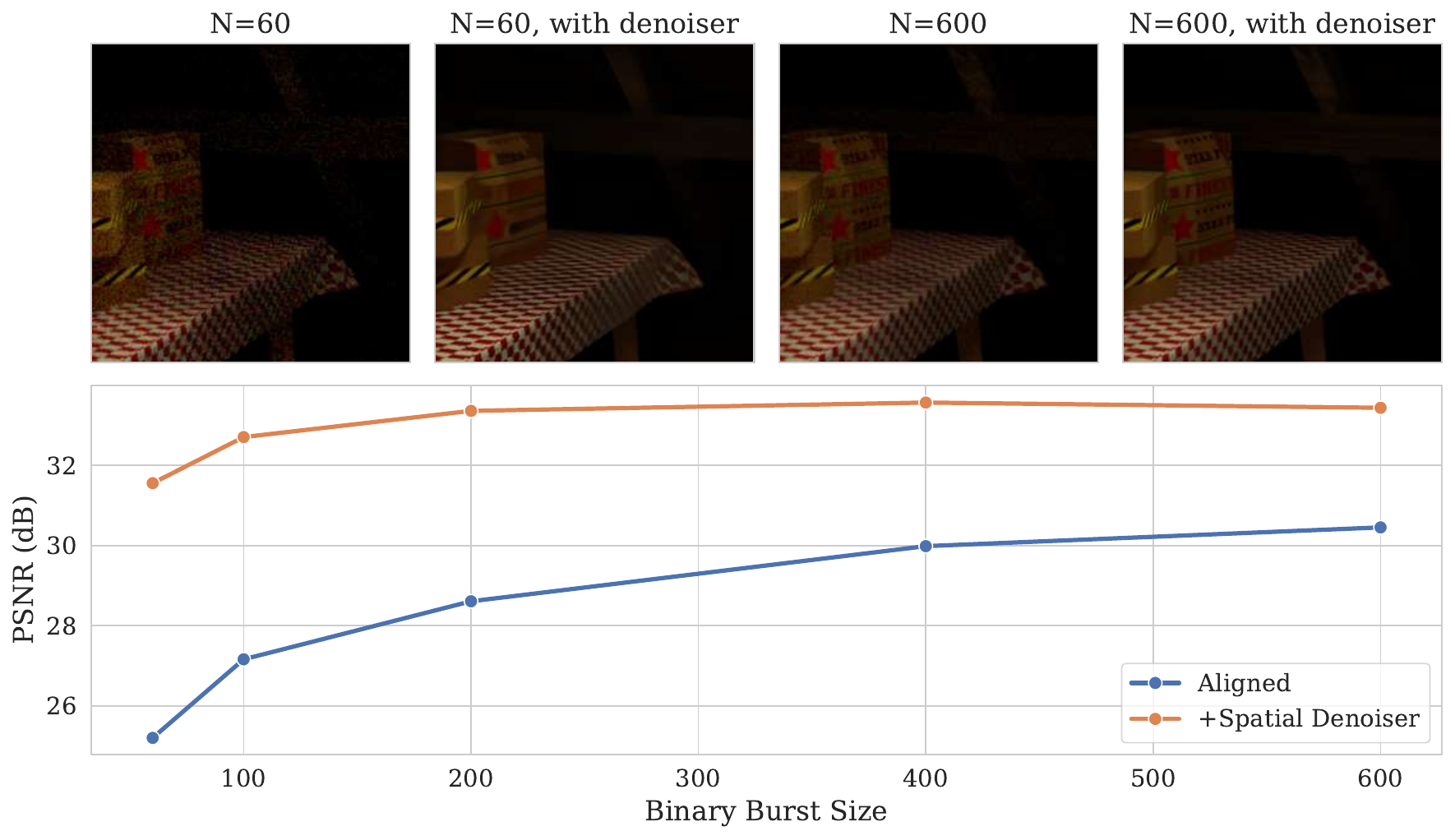}
    \caption{\textbf{Larger bursts improve quality}, up to field of view change when quality plateaus. At short burst sizes, the spatial denoiser needs to work harder to recover quality, while at longer bursts the aligned and merged image already has high quality. 
    }
    \label{fig:more-frames}
\end{figure}

\begin{figure*}
\centering
\includegraphics[width=\linewidth]{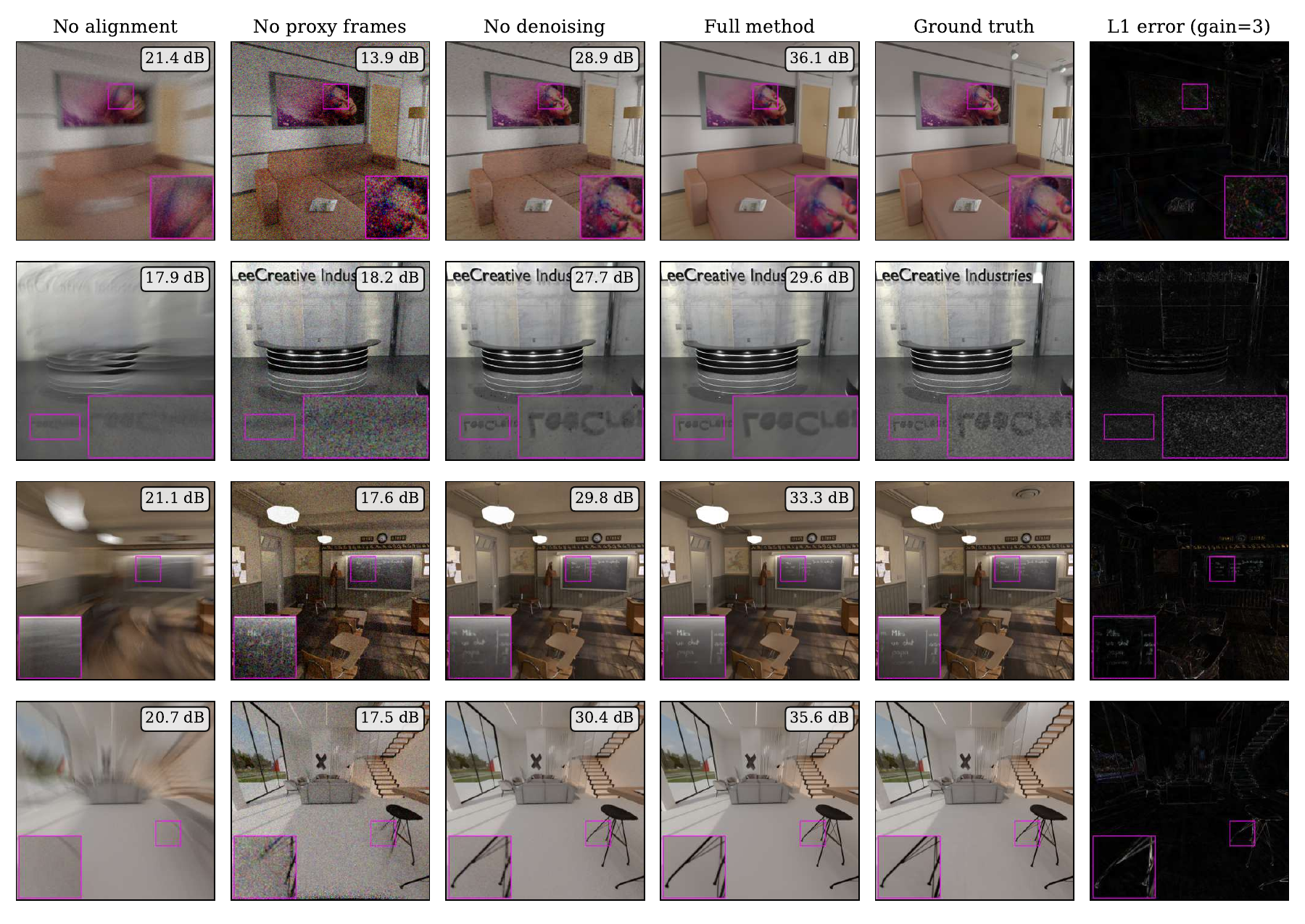}
\caption{\textbf{Additional qualitative method ablation.} Without motion compensation, fine details are quickly lost, while inaccurate flow estimation leaves a noisy result. Our full solution, combining robust temporal aggregation with spatial denoising, recovers low-blur and low-noise images. One limitation is high frequency details (such as the floor in row 2) may get oversmoothed.}
\label{fig:grid-supp}
\end{figure*}

Unlike prior learned methods, UnSPAD can effectively accumulate an infinite number of SPAD frames (as long as the scene continues to include relevant information).
In \autoref{fig:more-frames}, we show the effect of increasing burst size. Larger windows increase the effective bit depth of the reconstruction by aggregating more photon arrivals~\cite{Yang2012bits}. This reduces shot noise variance and enables finer intensity discrimination.
Longer bursts reduce the need for a spatial denoiser.

\section{Large Motion}\label{sec:supp:motion}

\begin{figure}
    \centering
    \includegraphics[width=\linewidth]{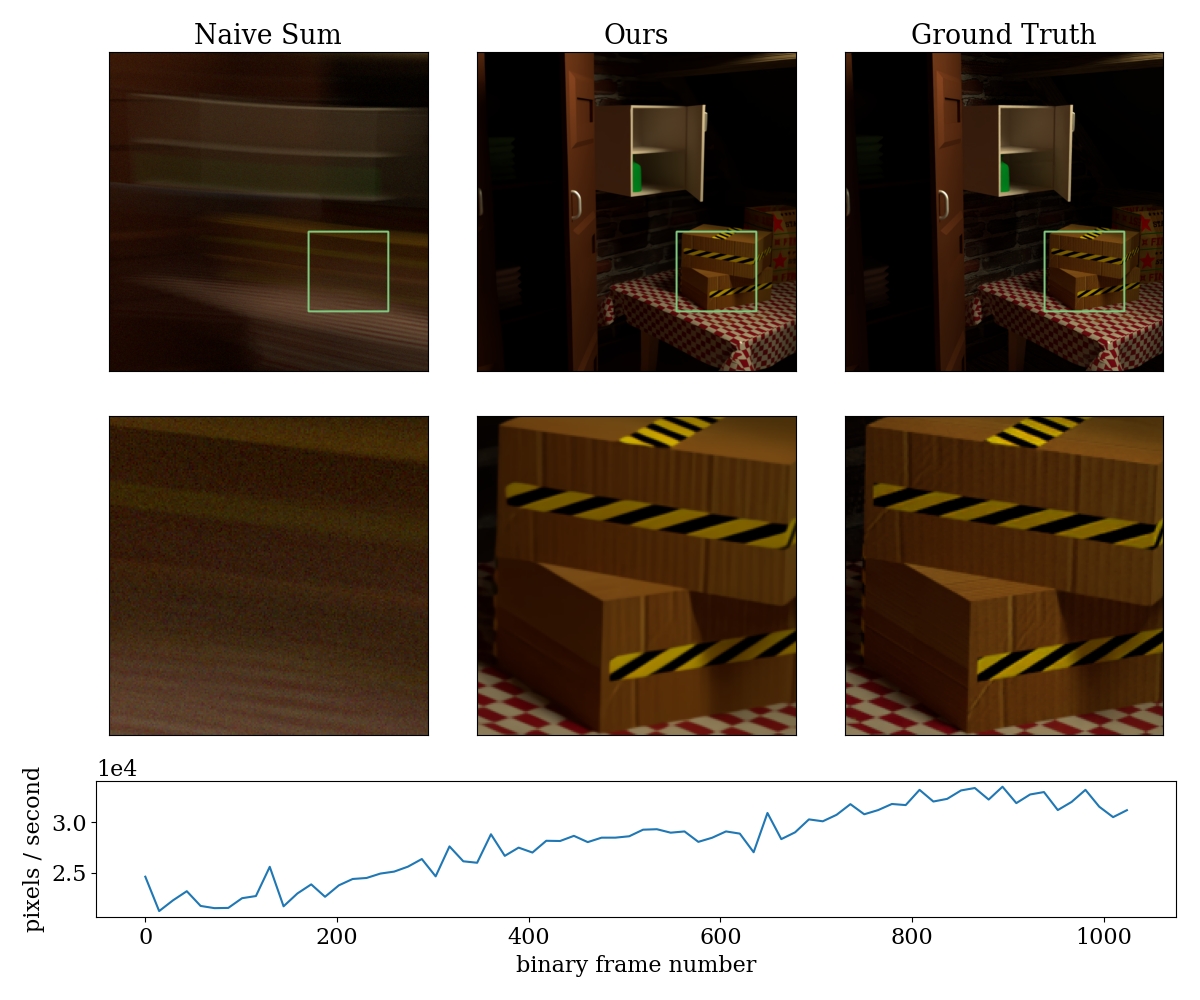}
    \caption{\textbf{Robust alignment across extreme motions.} Our method can successfully align $100$~kHz binary SPAD frames with virtually no motion blur, even when the motion is in excess of $25$k pixels per second (estimated with our optical flow algorithm).}
    \label{supp:fig:motion}
\end{figure}

We estimated the flow between high-speed CMOS frames in gQIR's evaluation with the Farneback method, revealing typical optical flow speeds from hundreds to millions of $\text{px}/\text{s}$. In \autoref{supp:fig:motion}, we show an example on the Single Photon Challenge dataset with up to $25$k $\text{px}/\text{s}$ ($300\times$ the field of view) due to camera rotation.

\section{Pixel Accuracy}\label{sec:supp:pix-acc}

\begin{figure}
    \centering
    \includegraphics[width=\linewidth]{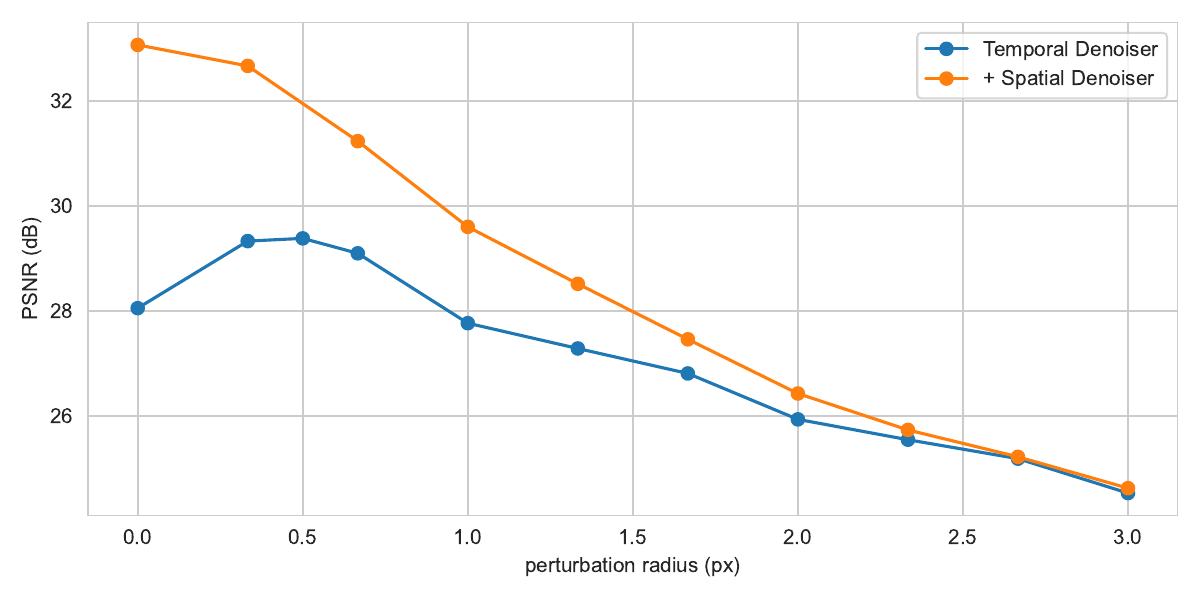}
    \caption{\textbf{Single-pixel flow accuracy.}
    As the reconstruction gets perturbed by increasingly strong shifts, image quality decreases. 
    The temporal denoised image does see a small bump at $<1$ pixel shifts due to bilinear interpolation acting as a spatial denoiser.}
    \label{supp:fig:shifts}
\end{figure}

To evaluate how well aligned our bursts are, we compute how perturbations change reconstruction quality.
For each image, we shift the image in a uniformly random direction by a radius using bilinear interpolation to handle sub-pixel shifts.
As the strength of perturbation increases, PSNR decreases (\autoref{supp:fig:shifts}).
Final reconstruction quality monotonically decreases as misalignment increases. A single pixel of misalignment causes $-3.5$ dB.
In the pre-spatially denoised reconstruction (Temporal Denoiser), a single pixel of misalignment causes $-0.3$ dB. Notice that quality peaks at $0.5$ pixels rather than $0$. This is due to bilinear interpolation acting as spatial denoiser, where the quality increase due to additional denoising is higher than the quality decrease due to misalignment.

\section{Changes to QBP}\label{sec:supp:qbp}

QBP~\cite{Ma_SIGGRAPH20} is effectively a temporal denoiser similar to the first stage of UnSPAD. There are 2 primary differences. (1) QBP uses block matching instead of a network based flow estimator. (2) QBP uses a robust Wiener merge to handle misalignment instead of masked sums.

\begin{figure}
    \centering
    \includegraphics[width=\linewidth]{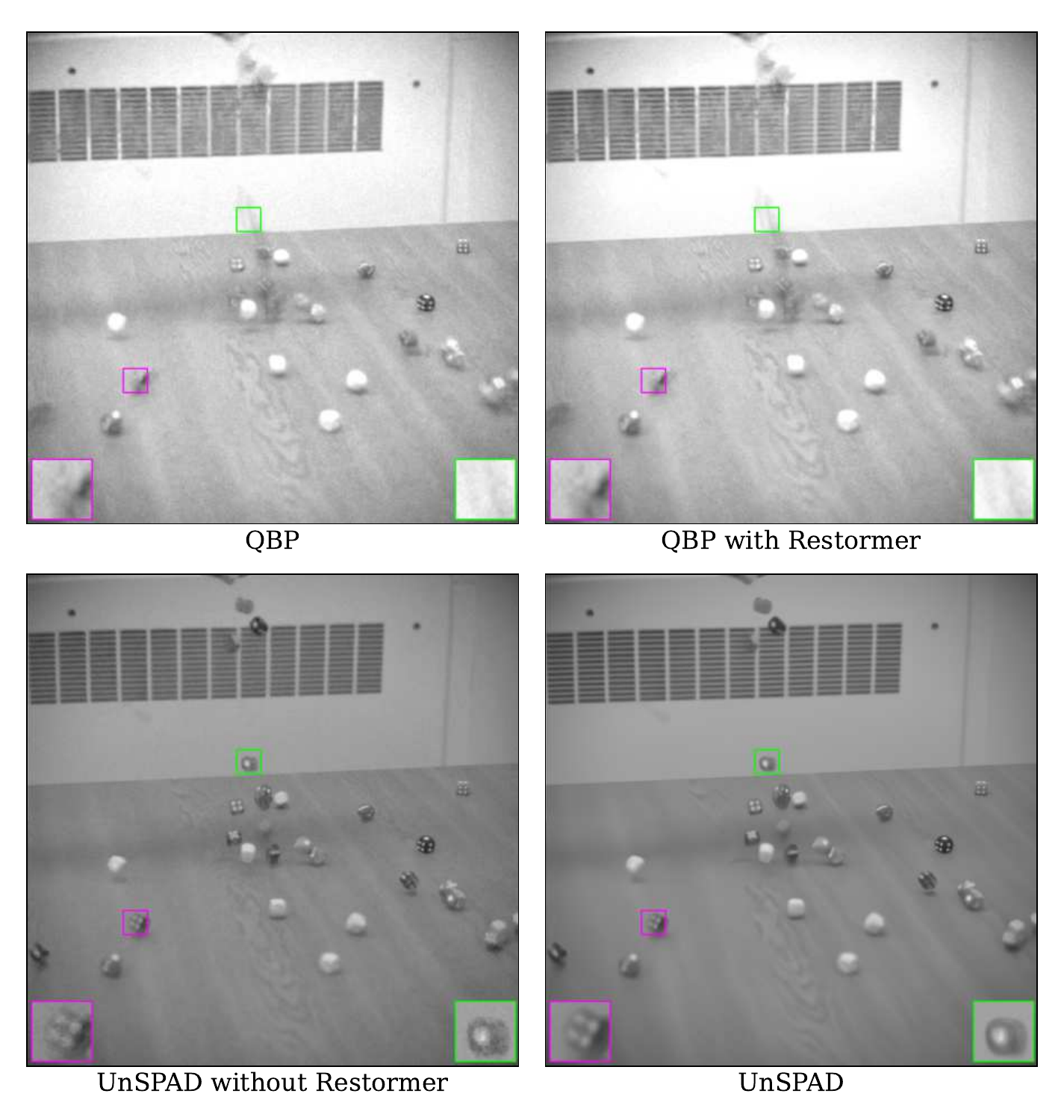}
    \caption{\textbf{QBP comparison on experimental SPAD data.} Notice QBP fails to preserve small moving objects, preventing spatial denoiser to improve accuracy.}
    \label{supp:fig:qbp-cmp-real}
\end{figure}

\begin{figure}
    \centering
    \includegraphics[width=\linewidth]{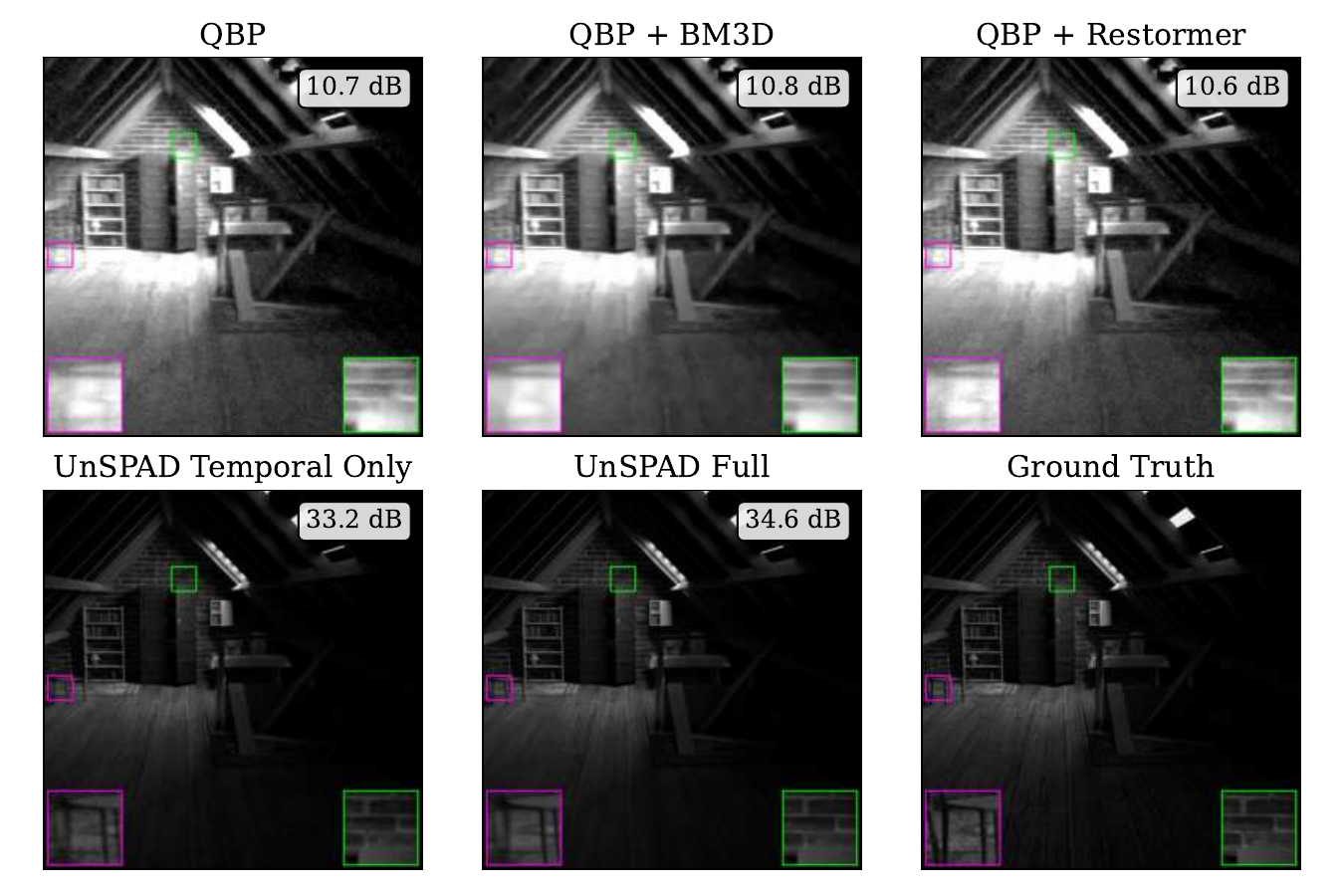}
    \caption{\textbf{QBP comparison on simulated data.} QBP overestimates brightness and has greater misalignment error. Two spatial denosiers, BM3D and Restormer, barely affect QBP.}
    \label{supp:fig:qbp-cmp-sim}
\end{figure}

We evaluate whether UnSPAD's improvement over QBP is due to the inclusion of a spatial denoiser or due to temporal denoiser algorithm differences. In \autoref{supp:tab:temp-res-gqir}, we show UnSPAD without a spatial denoiser (Restormer) is significantly better than QBP ($+14$ dB PSNR). Qualitatively on experimental SPAD data, Restormer has limited effect on QBP's output (\autoref{supp:fig:qbp-cmp-real}).
On simulated data with ground truth, spatial denoisers do not significantly improve QBP's quality (\autoref{supp:fig:qbp-cmp-sim}).
QBP has alignment errors on small objects and fast/complex movements the preclude spatial denoisers' ability to improve quality. 
We conclude that UnSPAD’s temporal denoising choices are the real advantage over QBP.

As QBP discusses, block matching's 2D motion model does not hold for non-rigid objects and 6-DOF trajectories. Especially for lower resolution sensors. Original burst photography considers much larger images (12 megapixels), which results in substantially smaller blocks that better follow 2D motion.
Our neural network based flow estimator (RAFT) can handle such motion.

Because shot noise dominates, the classical robust Wiener merge is ill-suited for $1$-bit SPAD frames (see Figure 5 of QBP). Instead, QBP pre-sums mini-bursts, then merges; precluding misalignment correction at SPAD frame-rates. By using flow-consistency for masked-sums, UnSPAD can.

\begin{table}[t]
    \centering

    \begin{tabular}{lccc}
        \toprule
        Method & PSNR$\uparrow$ & SSIM$\uparrow$ & LPIPS$\downarrow$ \\
        \hline
        QBP~\cite{Ma_SIGGRAPH20}        & {13.380} & {0.448} & {0.496} \\
        QUIVER~\cite{chennuri2024quanta}     & {22.429} & {0.814} & {0.429}  \\
        Burst-gQIR~\cite{garg_2026_gqir} & {$\underline{29.832}$} & {$\underline{0.856}$} & {$\underline{0.330}$} \\
        \midrule
        UnSPAD Temporal & $27.894$ & {$0.595$} & $0.3757$ \\
        UnSPAD Full       & {$\mathbf{33.007}$} & {$\mathbf{0.914}$} & {$\mathbf{0.193}$} \\
        \bottomrule %
    \end{tabular}
    \caption{
    \textbf{UnSPAD's temporal denoiser is competitive.} Evaluated on the Burst-gQIR dataset with $77$ simulated SPAD frames per reconstruction.}
    \label{supp:tab:temp-res-gqir}
\end{table}

\section{Bit2bit Comparison}\label{sec:supp:bit2bit}

\begin{figure}
    \centering
    \includegraphics[width=\linewidth]{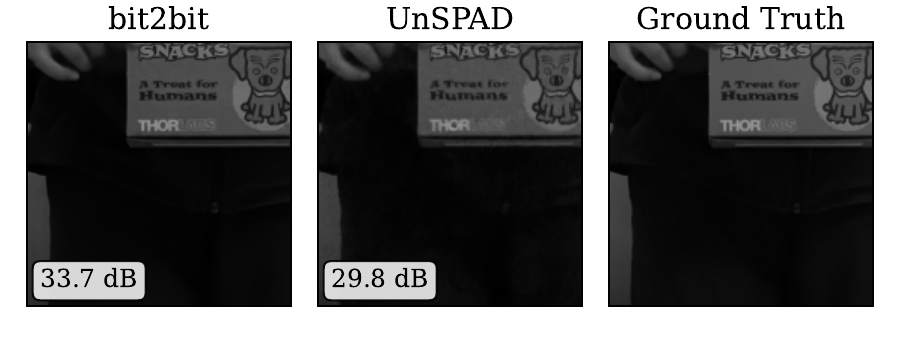}
    \caption{\textbf{Bit2bit comparison.} UnSPAD's result contains residual noise on bit2bit's simulated sequence. }
    \label{supp:fig:b2b}
\end{figure}

We were unable to get bit2bit's code to run on our data. On bit2bit's test sequence with ground truth, we achieve $29$ dB, while bit2bit reports $34$ dB. Bit2bit acknowledges that this performance is specific to the chosen parameters and dataset. We include a representative sample in \autoref{supp:fig:b2b}.
UnSPAD is faster than bit2bit ($0.5$ vs $6$-$10$ hours) as no network optimization is required at runtime.

\section{Evaluation Metrics}\label{sec:supp:metrics}

\noindent \textbf{Peak Signal-to-Noise Ratio (PSNR).} The standard PSNR definition is used,
\begin{align}
    \operatorname{PSNR} = 10 \log_{10} \frac{1^2}{MSE}
\end{align}
where $MSE$ is the error between the ground truth frame ${I}$  and the reconstruction $\hat{{I}}$.
\begin{align}
    MSE = \frac{1}{N} \sum_i \pare{ {I}_i - \hat{{I}}_i }^2
\end{align}

\noindent\textbf{Structural Similarity Index Measure (SSIM).} SSIM is made up of three components computed in $11\times 11$ patches:
\begin{align}
\operatorname{SSIM}(x, y)
= [l(x,y)]^\alpha
[c(x,y)]^\beta
[s(x,y)]^\gamma,
\end{align}
where
\begin{align}
    l(x,y) &=
\frac{2\mu_x\mu_y + C_1}
{\mu_x^2 + \mu_y^2 + C_1},\\
c(x,y) &=
\frac{2\sigma_x\sigma_y + C_2}
{\sigma_x^2 + \sigma_y^2 + C_2},\\
s(x,y) &=
\frac{\sigma_{xy} + C_3}
{\sigma_x\sigma_y + C_3}.
\end{align}
\( \mu_x, \mu_y \) are the means, \( \sigma_x^2, \sigma_y^2 \) are variances, and \( \sigma_{xy} \) is the covariance of images \( x \) and \( y \). Constants \( C_1, C_2, C_3 \) stabilize the division.

\noindent\textbf{Learned Perceptual Image Patch Similarity (LPIPS).}
Given a pretrained network \( \phi_l(\cdot) \) (VGG16 in this case) at layer \( l \), LPIPS is defined as:
\begin{align}
\operatorname{LPIPS}(x, y) = \sum_{l} \frac{1}{N_l} \sum_{i}
\left\| w_l \odot \left( \phi_l(x)_{i} - \phi_l(y)_{i} \right) \right\|_2^2,
\end{align}
where \( w_l \) are learned channel-wise weights, and \( \odot \) denotes element-wise multiplication.

\noindent\textbf{Inter-Frame Similarity (ITF).} At high frames, consecutive frames should be similar.
\begin{align}
\operatorname{ITF} = \frac{1}{T-1} \sum_{t=1}^{T-1} \operatorname{PSNR}(\hat{I}_t, \hat{I}_{t+1}).
\end{align}

\noindent\textbf{Warping error (${\mathbf E}_{\text{warp}}$).} Using optical flow $\mathbf{F}$, warping a frame to its neighbor should have low error.
\begin{align}
{E}_{\text{warp}}(I_t, I_{t+1}) = \left\| I_t - \mathcal{W}(I_{t+1}, \mathbf{F}_{t \to t+1}) \right\|_2.
\end{align}

\noindent\textbf{Multi-dimension Attention Network for Image Quality Assessment (MANIQA).} A neural network computes a quality score, approximately given as,
\begin{align}
\operatorname{MANIQA} = f_{\mathrm{reg}}
\left(
f_{\mathrm{MDA}}
\left(
f_{\mathrm{ViT}}(\hat{I})
\right)
\right).
\end{align}

\end{document}